\documentclass[journal]{IEEEtran}

\usepackage{amsmath,amsfonts,amssymb}
\usepackage{algorithmic}
\usepackage{algorithm}
\usepackage{array}
\usepackage{textcomp}
\usepackage{stfloats}
\usepackage{url}
\usepackage{verbatim}
\usepackage{graphicx}
\usepackage{subcaption}
\usepackage{cite}

\usepackage{xcolor}
\usepackage{siunitx}
\usepackage{booktabs}

\usepackage{xurl}
\usepackage{tabularx}
\usepackage{diagbox}
\usepackage{makecell}

\usepackage{tikz}

\usepackage{bbding}
\usepackage{pifont}% http://ctan.org/pkg/pifont
\usepackage{multirow}

\usepackage{scalerel}
\usepackage{tikz}
\usetikzlibrary{svg.path}

\definecolor{orcidlogocol}{HTML}{A6CE39}
\tikzset{
	orcidlogo/.pic={
		\fill[orcidlogocol] svg{M256,128c0,70.7-57.3,128-128,128C57.3,256,0,198.7,0,128C0,57.3,57.3,0,128,0C198.7,0,256,57.3,256,128z};
		\fill[white] svg{M86.3,186.2H70.9V79.1h15.4v48.4V186.2z}
		svg{M108.9,79.1h41.6c39.6,0,57,28.3,57,53.6c0,27.5-21.5,53.6-56.8,53.6h-41.8V79.1z M124.3,172.4h24.5c34.9,0,42.9-26.5,42.9-39.7c0-21.5-13.7-39.7-43.7-39.7h-23.7V172.4z}
		svg{M88.7,56.8c0,5.5-4.5,10.1-10.1,10.1c-5.6,0-10.1-4.6-10.1-10.1c0-5.6,4.5-10.1,10.1-10.1C84.2,46.7,88.7,51.3,88.7,56.8z};
	}
}

\newcommand\orcidicon[1]{\href{https://orcid.org/#1}{\mbox{\scalerel*{
				\begin{tikzpicture}[yscale=-1,transform shape]
					\pic{orcidlogo};
				\end{tikzpicture}
			}{|}}}}

\usepackage{hyperref} 

\usepackage{hyperref} 
\hypersetup{
    colorlinks=true,
    breaklinks=true,
    linkcolor=blue,
    filecolor=magenta,      
    hypertexnames=true,
    urlcolor=cyan,
    pdfpagemode=FullScreen,
    citecolor=green
    }
\title{A New Perspective On AI Safety \\ Through Control Theory Methodologies}

\author{Lars Ullrich\IEEEauthorrefmark{1} $^{\orcidicon{0009-0001-8166-3118}}$,
~Walter Zimmer\IEEEauthorrefmark{2} $^{\orcidicon{0000-0003-4565-1272}}$,~\IEEEmembership{Member,~IEEE},
~Ross Greer\IEEEauthorrefmark{3}\IEEEauthorrefmark{4} $^{\orcidicon{0000-0001-8595-0379}}$,~\IEEEmembership{Member,~IEEE},
~Knut Graichen\IEEEauthorrefmark{1} $^{\orcidicon{0000-0003-2865-8093}}$,\\~\IEEEmembership{Senior Member,~IEEE},
~Alois C. Knoll\IEEEauthorrefmark{2} $^{\orcidicon{0000-0003-4840-076X}}$ ,~\IEEEmembership{Fellow,~IEEE} 
and Mohan Trivedi\IEEEauthorrefmark{3} $^{\orcidicon{0000-0002-0937-6771}}$ ,~\IEEEmembership{Life Fellow,~IEEE}

	\thanks{This research is accomplished within the project ”AUTOtech.agil” (FKZ 01IS22088Y, FKZ 01IS22088U). We acknowledge the financial support for the project by the Federal Ministry of Education and Research of Germany (BMBF).\\Corresponding author: {\tt\footnotesize lars.ullrich@fau.de}}
	\thanks{\IEEEauthorrefmark{1} L. Ullrich and K. Graichen are with the Chair of Automatic Control at Friedrich-Alexander-Universität Erlangen-Nürnberg (FAU), Cauerstraße 7, 91058 Erlangen, Germany (E-mail: lars.ullrich@fau.de, knut.graichen@fau.de)}
    \thanks{\IEEEauthorrefmark{2} W. Zimmer and A. Knoll are with the Chair of Robotics and Artificial Intelligence at the Technical University of Munich (TUM), 85748 Garching bei M\"unchen, Germany (E-mail: walter.zimmer@tum.de, knoll@in.tum.de)}
    \thanks{\IEEEauthorrefmark{3} R. Greer, and M. Trivedi are with the Laboratory for Intelligent and Safe Automobiles at the University of California San Diego (UCSD) (E-mail: regreer@ucsd.edu, mtrivedi@ucsd.edu)}
    \thanks{\IEEEauthorrefmark{4} R. Greer is with the University of California Merced (UCM) (E-mail: rossgreer@ucmerced.edu)}}

\usepackage{hologo}
\usepackage{listings}
\begin{document}

% for arXiv publication with appropriate copyright notice
\twocolumn[
\begin{@twocolumnfalse}
	\Huge {IEEE copyright notice} \\ \\
	\large {\copyright\ 2025 IEEE. Personal use of this material is permitted. Permission from IEEE must be obtained for all other uses, in any current or future media, including reprinting/republishing this material for advertising or promotional purposes, creating new collective works, for resale or redistribution to servers or lists, or reuse of any copyrighted component of this work in other works.} \\ \\
	
	{\Large Accepted to be published in IEEE Open Journal of Intelligent Transportation Systems.} \\ \\
	
	Cite as:
	
	\vspace{0.1cm}
	\noindent\fbox{%
		\parbox{\textwidth}{%
			L.~Ullrich, W.~Zimmer, R. Greer, K.~Graichen, A.~Knoll, and M.~Trivedi, "A New Perspective On AI Safety Through Control Theory Methodologies" in \emph{2025 IEEE Open J. Intell. Transp. Syst. (OJ-ITS).}, to be published.
		}%
	}
	\vspace{2cm}
	
\end{@twocolumnfalse}
]

\noindent\begin{minipage}{\textwidth}
	
\hologo{BibTeX}:
\footnotesize
\begin{lstlisting}[frame=single]
@article{ullrich2025datacontrol,
    title={{A New Perspective On AI Safety Through Control Theory Methodologies}},
    author={Ullrich, Lars and Walter, Zimmer and Greer, Ross and Graichen, Knut and Knoll, Alois C. and Trivedi, Mohan},
    journal={IEEE Open Journal of Intelligent Transportation Systems (OJ-ITS)},
    year={2025}, 
    publisher={IEEE. to be published}
}


\end{lstlisting}
\end{minipage}

\maketitle
\thispagestyle{empty}
\pagestyle{empty}

\bstctlcite{IEEEexample:BSTcontrol}

\begin{abstract}\label{00_Abstract}
	While artificial intelligence (AI) is advancing rapidly and mastering increasingly complex problems with astonishing performance, the safety assurance of such systems is a major concern. Particularly in the context of safety-critical, real-world cyber-physical systems, AI promises to achieve a new level of autonomy but is hampered by a lack of safety assurance. While data-driven control takes up recent developments in AI to improve control systems, control theory in general could be leveraged to improve AI safety. Therefore, this article outlines a new perspective on AI safety based on an interdisciplinary interpretation of the underlying data-generation process and the respective abstraction by AI systems in a system theory-inspired and system analysis-driven manner. In this context, the new perspective, also referred to as data control, aims to stimulate AI engineering to take advantage of existing safety analysis and assurance in an interdisciplinary way to drive the paradigm of data control. Following a top-down approach, a generic foundation for safety analysis and assurance is outlined at an abstract level that can be refined for specific AI systems and applications and is prepared for future innovation.
\end{abstract}

\begin{IEEEkeywords}
	AI Safety, Control Theory, New Perspective, Interdisciplinary Approach, New Paradigm, Data Control
\end{IEEEkeywords}

\section{INTRODUCTION}\label{sec:01_Introduction}

Artificial intelligence (AI) methods provide a toolbox that enables a wide range of data-based tasks to be accomplished. These include tasks such as classification \cite{gao2018object}, regression \cite{letzgus2022toward}, clustering \cite{ezugwu2022comprehensive}, time series forecasting \cite{lim2021time}, image processing, \cite{zhang2019application}, natural language processing \cite{lauriola2022introduction} and more \cite{bareinboim2016causal, zhuang2020comprehensive}. Alongside the wide range of tasks, there exists a huge diversity in terms of applications and corresponding data properties. Accordingly, AI encompasses a variety of approaches and methods that reflect the respective diversity of needs. For instance, data-based methodologies like machine learning can be divided into main categories such as supervised \cite{cunningham2008supervised}, unsupervised \cite{hastie2009unsupervised}, or reinforcement learning \cite{sutton2018reinforcement}. Thereby, the categories address different characteristics of the applications and associated properties of available data. Furthermore, AI systems can also be distinguished into discriminative or generative AI \cite{jebara2012machine, wang2025generative}, and narrow or general AI \cite{jiang2022quo}. These distinctions illustrate the broad applicability and high diversity of the methods as well as the variety of requirements and perspectives. As a result, the challenge of a generic safety analysis and assurance guideline becomes apparent while underlining the need for a suitable level of abstraction. 

Apart from all the differentiations, current data-driven learning systems are united by the fact that they are data-based systems with defined input and output data. In other words, these systems represent a data-based counterpart to traditional signals and systems that could serve as a basis for a general safety methodology at an abstract level. This is especially the case as extensive system analyses and safety assessments based on system theory have already been carried out for many years. However, systems theory \cite{oppenheim1997signals, girod2013einfuhrung} was originally introduced as a mathematical abstraction of physical, chemical or technical systems. The technological progress with the advent of digital computing \cite{dally1998digital} once transferred these concepts from continuous time to discrete time \cite{phillips2007digital}. In comparison, the current upsurge in AI \cite{krittanawong2018rise, jha2019comprehensive, bohr2020rise, cioffi2020artificial, alhayani2021effectiveness, ahmad2021artificial} represents a transition from mathematically explicit signals and systems, e.g., ordinary differential equations (ODEs), to data-based implicit signals and systems such as neural networks. Thereby, the transition from classical system theory to data-based signals and systems is notably more difficult than the transition from mathematically continuous to mathematically discrete system descriptions. The significantly changed character of the representation and especially the implicit nature represent challenges \cite{kurd2007developing, forsberg2020challenges} and require an adapted methodology of system analysis.

\begin{figure*}[t]
	\centering	
    \includegraphics[width=1.0\textwidth]{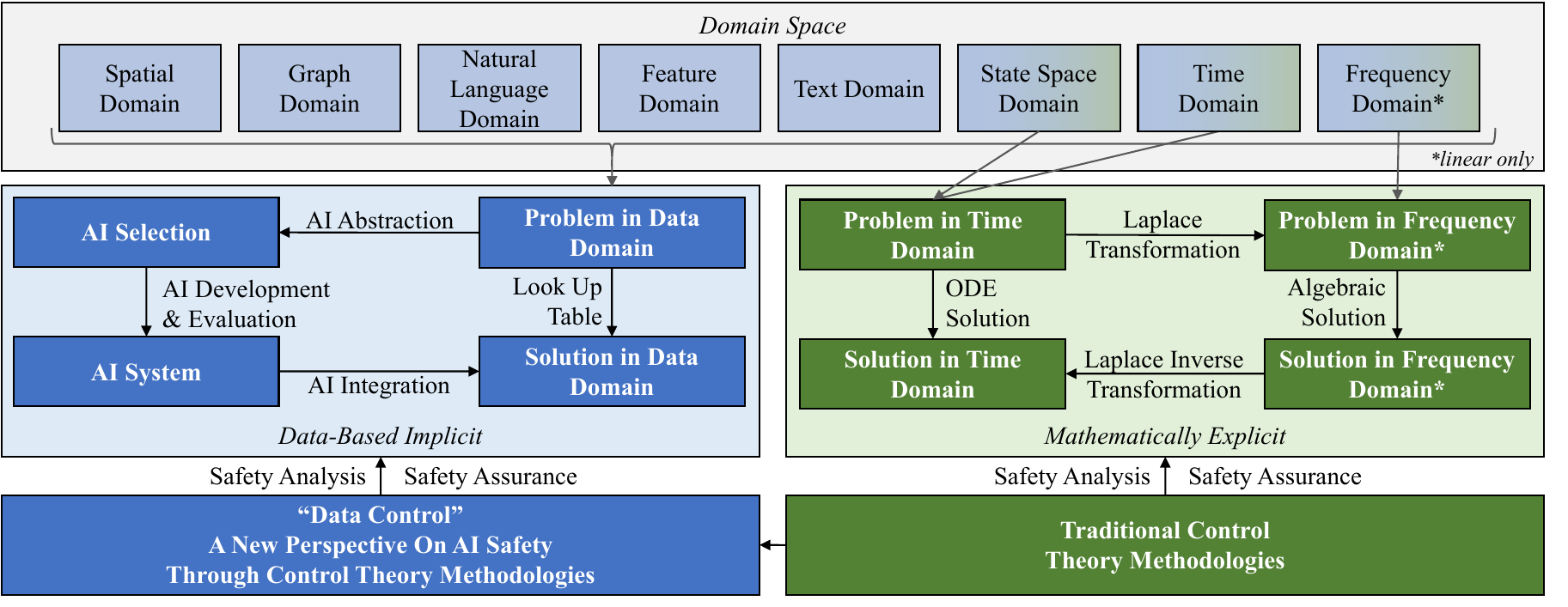}
	\caption{Comparison of mathematically explicit (green) and data-based implicit (blue) signal and system methods as well as visualization of commonalities in the processes, differences in the spheres of impact, and depiction of the field of action of this work.}
	\label{fig:general}
\end{figure*}

Moreover, the transition from mathematically explicit to data-based implicit systems can also be considered as an extension or expansion to a variety of domains. As depicted in Figure \ref{fig:general}, the domain space of AI systems encompasses a multitude of domains in addition to the time, frequency, and state space domains widely used in systems theory. This reflects the broad AI application \cite{zawacki2019systematic, sezer2020financial, chen2020artificial, abdallah2020artificial, khanagar2021developments} far beyond physical, chemical, and technical systems.

While a variety of AI approaches  \cite{hastie2009unsupervised, sutton2018reinforcement, jebara2012machine, jiang2022quo, pmlr-v37-romera-paredes15, vinyals2016matching, kadam2020review, vanschoren2019meta} have evolved and a generic AI synthesis, as shown in Figure \ref{fig:general}, has become established, current methods for system analysis of AI systems are insufficient \cite{neto2022safety}. In particular, dedicated approaches \cite{gillula2012guaranteed, corso2021survey, pei2017deepxplore} are developed for specific AI systems \cite{chen2020tensorfi, alemany2021jespipe, trusted2022github} and applications \cite{machin2016smof, schirmer2018considerations, jia2022role, liu2021dloam}. In contrast, a generic methodology that is transferable to different systems and applications is desirable. In other words, a system-theoretical methodology, as used in control engineering \cite{goodwin2001control, lunze2010regelungstechnik, follinger2011laplace, fadali2012digital, nise2020control}, which first defines terms such as stability and robustness generically and subsequently concretized them in different domains to enable targeted system analysis and validation. In order to address this problem and form a far-reaching, generic foundation, a new interdisciplinary perspective for AI safety is introduced in a system theory-inspired and system analysis-driven manner within this paper, that ultimately justifies the establishment of a new paradigm called data control. The corresponding placement of the new perspective and the data control paradigm is illustrated in Figure \ref{fig:general}. As outlined, data control is intended to represent a pendant to traditional methods, although inspired by them, with the claim of a generic basis for AI system analysis and subsequent validation. While existing approaches \cite{zhang2018overview, chang2019neural, berberich2020data, pauli2021training, wabersich2021predictive} have examined individual concepts of control engineering concerning AI system analysis and assurance, the new perspective goes further and is more disruptive. In particular, the fusion with existing system analysis approaches driven by AI research and the consideration of diverse data domains requires a reinterpretation of concepts such as stability. This partially radical reinterpretation, essential for fusion, further enables the integration of diverse backgrounds, thereby making AI safety accessible and advancing it in an interdisciplinary manner. Accordingly, the main contribution of this paper is the introduction of the data control paradigm, which includes the following sub-contributions:

\begin{itemize}
    \item The fusion of different perspectives towards an interdisciplinary perspective on the underlying data-generating process and data-based AI signals and systems.
    \item The subsequent definition of system classes and properties to provide a generic basis for the safety analysis and assurance of AI systems.
    \item The formalization of diverse mechanisms to enhance AI systems, aiming to increase safety by preventing harm, mitigating risks, and averting undesirable behaviors throughout their lifecycle.
\end{itemize}

\begin{figure}[]
	\centering	
	\includegraphics[scale=0.48]{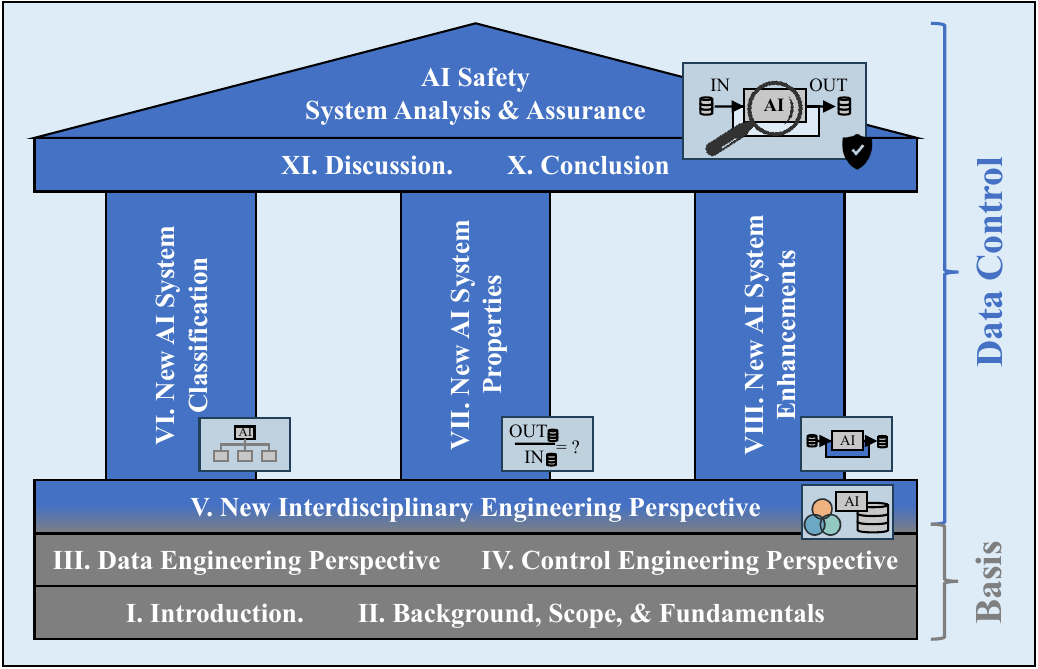}
	\caption{The illustration presents the overall structure of the proposed data control paradigm, complemented by a visual representation of the paper's organization. Grey elements represent existing baseline approaches, while blue elements highlight the novel contributions introduced in this work. The paradigm is built on an interdisciplinary perspective, grounded in systemic thinking and system analysis, integrating foundational concepts from control theory with perspectives from AI. By defining system classes, properties, and formalizing mechanisms, AI system safety is enhanced through the generic data control methodology, which aims to prevent harm, mitigate risks, and avoid undesirable behavior throughout the lifecycle.}
	\label{fig:DataControl}
\end{figure}

Overall, the various sub-contributions collectively justify the establishment of the new paradigm termed data control, which is summarized in the structured representation shown in Figure \ref{fig:DataControl}. The paper is structured as follows: Initially, Section \ref{sec:02_Background_Fundamentals} briefly outlines the background, the scope and fundamental definitions. Section \ref{sec:03_Data_Engineering_Perspective} is dedicated to the data engineering perspective from a theoretical and practical point of view. Section \ref{sec:04_Control_Engineering_Perspective} presents the system-oriented perspective of control engineering on AI systems. Building on Section \ref{sec:03_Data_Engineering_Perspective} and \ref{sec:04_Control_Engineering_Perspective}, Section \ref{sec:05_Interdisciplinary_Engineering_Perspective} introduces an interdisciplinary engineering perspective. In Section \ref{sec:06_AI_System_Classification}, the interdisciplinary perspective is applied more specifically to AI systems by defining system classes. Section \ref{sec:07_AI_System_Properties} continues with a description of system properties. Section \ref{sec:08_AI_Enhancements} illustrates an enhanced conceptualization of the new perspective, combining existing concepts with new ideas while taking regulatory requirements into account. Subsequently, Section \ref{sec:09_Discussion} discusses the new perspective and justifies the establishment of a new paradigm called data control. Section \ref{sec:10_Concluison} identifies future research directions and concludes this paper.

\section{BACKGROUND, SCOPE, \& FUNDAMENTALS}\label{sec:02_Background_Fundamentals}

This section first provides an overview of the related work. Building on this, a more context-specific problem description and delimitation are outlined. Furthermore, a generic definition of AI is provided that can be considered a common definition of AI systems and signals and serves as a foundation for the following sections.

\subsection{Related Work}
System-theoretic and control engineering methods \cite{goodwin2001control, lunze2010regelungstechnik, follinger2011laplace, fadali2012digital, nise2020control} focus on mathematically explicit system descriptions, making a direct and generic transfer of safety analysis and safety assurance challenging. Even though the AI subfield of online learning \cite{sutton2018reinforcement, nagabandi2018deep} and especially online reinforcement learning \cite{wei2017online, dogru2021online}, which can be seen as an AI-based counterpart to feedback control, addresses safety concerns \cite{berkenkamp2017safe, mao2019towards, valiente2022robustness, zheng2023learning}, the methods cannot be transferred to the generality of AI. Nevertheless, approaches within this field, like predictive safety filters \cite{wabersich2021predictive, wabersich2023data, leeman2023predictive} which rely on model predictive control (MPC) \cite{kouvaritakis2016model}, show that control theory methods are already being used to ensure the safety of a selection of AI systems. Furthermore, the intertwining of AI and control engineering is also demonstrated by the paradigm of data-driven control \cite{hou2013model, hou2016overview, maupong2017data, de2019formulas, torrente2021data, markovsky2021behavioral}. Although the focus is primarily on data-based modeling and controller design, research is likewise being conducted on the associated safety aspects \cite{rosolia2017learning, berberich2020data, dorfler2022bridging}. Thereby, for instance, Lyapunov stability is under consideration \cite{xu2013adaptive, chang2019neural, van2023behavioral}. Despite this, these methods are not universally applicable. In contrast, approaches that limit AI system performance or replace AI's with a traditional system based on safety-related thresholds \cite{shukla2020flight} are more widely transferable but limit AI capabilities.

In addition to the increasing integration of control engineering, other disciplines are also being considered for AI safety analysis. For instance, many formalized approaches \cite{scholkopf2012causal, sugiyama2007covariate, zhang2013domain, schulam2017reliable} stem from causality \cite{pearl2009causality}, which examines relationships, influencing factors, and effects of observable variables of the data-generating process \cite{pearl2009causality}.

 While formal, these approaches face practical challenges, such as unknown influences like latent variables or confounders \cite{pearl2009causality}. Conversely, practically driven methods focus on counteracting challenges without system analysis, giving rise to techniques like out-of-distribution detection \cite{shriram2025towards, greer2024perception}. New methods, such as few-shot learning \cite{wang2020generalizing, kadam2020review, greer2025language, greer2024towards} and meta-learning \cite{hochreiter2001learning, finn2017model, vanschoren2019meta}, address unseen data or changing relationships. Additionally, distributionally robust optimization \cite{esfahani2015data, sinha2017certifying, rahimian2019distributionally} has emerged to make systems more robust to predefined changes, with some initial analysis approaches defining shift stability \cite{subbaswamy2020development, subbaswamy2021evaluating}.

To address method-specific approaches and integrate causality-driven formalizations with system-theoretical safety assessments, the underlying data-generation process as well as the abstraction by AI models is reconsidered in an interdisciplinary, system-analytic way. Based on this, AI system classes and properties are defined, which enable a subsequent analysis of the multitude of AI systems in a corresponding system-theoretical manner, while respecting the AI system specific characteristics. Further details and delimitations are outlined below.

\subsection{Problem Statement \& Delimitation}
While AI system analysis and safety assurance are critical \cite{amodei2016concrete}, method-specific approaches dominate \cite{neto2022safety, afxentiou2025evaluation}. A formalized causality-based methodology has recently provided a broader definition of stability \cite{subbaswamy2021evaluating, subbaswamy2023causal}, primarily focusing on dataset shift-invariance. However, this contrasts with the control engineering view of stability \cite{goodwin2001control, lunze2010regelungstechnik, follinger2011laplace, fadali2012digital, nise2020control}, which is centered on dynamic systems. While most AI systems have been static learned mappings, dynamic AI systems, like Recurrent Neural Networks (RNNs) \cite{rumelhart1986learning, hochreiter1997long}, also exist. Recent efforts to achieve higher autonomy and intelligence in systems increasingly rely on dynamic systems, such as objective-driven recurrent world models \cite{lecun2022path}, which integrate model predictive control into AI. This highlights that focusing solely on dataset shift invariance is insufficient for stability and AI system analysis. Instead, a system-theoretical classification is necessary to differentiate AI systems by their behavior, defining system properties similar to control engineering as a general basis of system analysis. This contrasts with the performance-oriented approach in AI development, aiming to address challenges. However, direct application of systems theory is limited by the different system descriptions and the broader scope of AI systems. Finally, the difference in the system integration between AI systems and control systems is also significant. While control systems can exert influence, AI systems are mostly exposed to their environment. These challenges must be addressed and an interdisciplinary perspective needs to be adopted. 

To this end, static, non-stationary, and dynamic systems and system properties of stability, robustness, and sensitivity need to be defined in general terms from a system analysis-driven perspective. Similar to control engineering, these can be concretized and specified in individual AI system classes in the respective future. However, a superordinate general understanding is elementary, which is intended to be formalized in the course of this paper. An inherent and implicit consideration of such aspects is further illustrated by a subsequent conceptualization that considers regulatory requirements as well. This establishes a reference to existing approaches while providing an outlook. Furthermore, this opens up a formalized perspective along with the establishment of a new paradigm called data control while raising a multitude of future research questions.

\subsection{Fundamental Definitions: AI Signals, Systems, and Safety}

\begin{figure*}
	\centering	
	\includegraphics[width=1.0\textwidth]{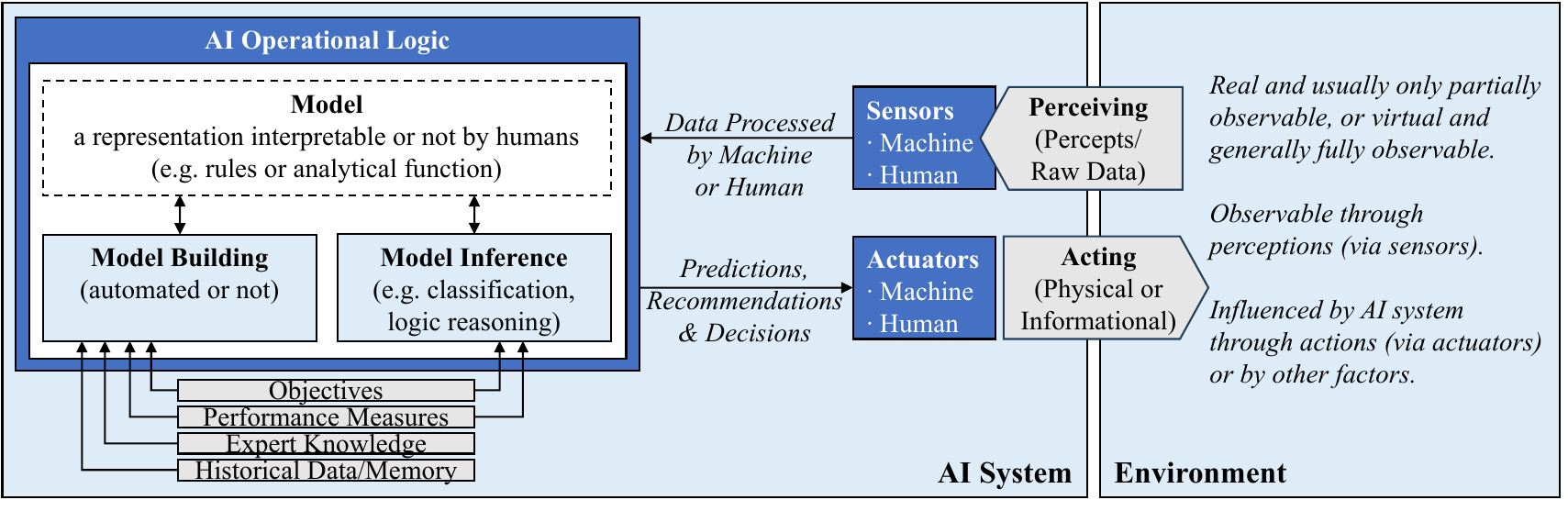}
	\caption{AI system according to OCED \cite{OECDpub}.}
	\label{fig:OECD}
\end{figure*}

AI has evolved significantly over the years, and its definition has been continuously refined. A widely recognized reference for defining AI systems, signals, and safety is provided by the OECD. As illustrated in Figure \ref{fig:OECD}, the OECD defines AI systems as follows:

\textbf{Definition 2.1 (Artificial intelligence)} \textit{An AI system is a machine-based system that is capable of influencing the environment by producing an output (predictions, recommendations, or decisions) for a given set of objectives. It uses machine and/or human-based data and inputs $X$ to (i) perceive real and/or virtual environments $\mathcal{E}$; (ii) abstract these perceptions into models $\mathcal{M}$ through analysis in an automated manner (e.g., with machine learning), or manually; and (iii) use model inference $\mathcal{M}(X)$ to formulate options for outcomes  $Y$. AI systems are designed to operate with varying levels of autonomy.}\cite{OECDpub}

While this definition is broadly formulated and encompasses all types of AI systems, the application scope of this paper is specified. Symbolic AI systems from the first wave of AI are, on the one hand, more manageable and do not pose the challenges of modern AI systems. On the other hand, they play a subordinate role in today's context. Accordingly, this paper addresses modern data-based AI systems of the second and third waves of AI, which are data-driven learning methodologies known, for instance, as statistical learning and contextual adaption. Definition 2.1, which can be interpreted as a combined AI signal and system definition in general form, serves as a general foundation.

\textbf{Definition 2.2 (AI Safety)} \textit{Ensures AI systems function as intended throughout their entire lifecycle, while mitigating risks, preventing harm, and averting undesirable behavior. Thus, AI safety enables effective risk management and ensures the reliability and trustworthiness of AI systems.}

This definition aligns with existing definitions, such as the explicit definition stated by the OECD AI Principles (Principle 1.4 on Robustness, Security, and Safety) \cite{OECD_AI_safe}, the implicit definition outlined in the International AI Safety Report \cite{bengio2024international}, as well as definitions from various research facilities\cite{ruess2022safe, NIST_AI_safe, herrera2025responsible, hendrycks2025introduction}. Overall, it becomes evident that specifying and analyzing AI systems beyond performance measures, focusing more on systematic behavior specification and analysis, as well as incorporating mechanisms to detect and address potential issues, is essential.
\section{DATA ENGINEERING PERSPECTIVE}\label{sec:03_Data_Engineering_Perspective}

This section examines data engineering's perspective, laying the groundwork for an interdisciplinary approach and formalization in later sections. It outlines the influencing factors, vulnerabilities, and countermeasures of data-based AI systems to support safety analysis.

\subsection{General Challenge}
Data is central to AI systems, encompassing various modalities such as low-dimensional, high-dimensional, structured, unstructured, raw, processed, and aggregated meta-data. The combination of large data volumes, high-performance computing, and advanced machine learning methods has driven significant success in large-scale pattern recognition. These statistical techniques excel at working with diverse observational data \cite{pearl2009causality, li2023probabilities}.

In the context of data-based AI systems, data is of central importance and has a wide variety of modalities. There exist low-dimensional and high-dimensional data, structured and unstructured data, raw,  processed, and highly aggregated meta-data. In particular, the vast amounts of data combined with high-performance computer systems and high-capacity machine and deep learning methodologies have led to great successes in large-scale pattern recognition. These statistical learning techniques offer the advantage of being able to work with arbitrary observational data \cite{pearl2009causality, li2023probabilities}. 

At the same time, these approaches lack formalism and explainability, with performance rapidly degrading outside experimental conditions, such as violations of the \textit{independent and identically distributed (i.i.d.) assumption}. Even minor changes in data distributions can have a significant impact \cite{bissoto2024even, holzinger2021next}, creating vulnerabilities to adversarial attacks. Additionally, slight deviations between training and deployment are expected, especially in cyber-physical systems due to factors like signal noise, production tolerances, or recalibrations.

The following subsections explore this challenge from both theoretical and practical data engineering perspectives. The theoretical perspective focuses on causal and statistical relationships, their linkages, and classifications of shifts. The practical perspective highlights the limitations of the theoretical approach and presents practice-based methods to address current challenges.

\subsection{Theoretical Perspective on AI Vulnerability}\label{sec:dataeng_theo}

Classical statistical learning methods create \textit{static} AI systems, where statistical relationships can be either static or dynamic. If \textit{dynamic statistical relationships} are present, reliable use of the classical statistical AI is only admissible under training-related static conditions (i.e., when the i.i.d. assumption holds). In contrast, with \textit{static statistical relationships}, the generalizing usability beyond the distribution of the training dataset is given. Thus, in this case, making the learned data-based AI system inherently less vulnerable and only subject to the risk of local over-simplification of the statistical relationship due to finite training data \cite{storkey2008training}.  

The observable data applied in statistical learning stem from an underlying data-generating process (DGP) \cite{pearl2009causality}. For instance, statistical dependencies, such as causal conditional probabilities generated by an physical mechanisms, represent static or invariant statistical relationships. Understanding the task's causality provides insights into the vulnerability and transferability of data-based AI systems. This connection between causality and statistical dependence was noted by Reichenbach in 1956 \cite{reichenbach1991direction} and is compactly described by \cite{scholkopf2021toward} as follows:

\begin{quote}"\textbf{Reichenbach's Common Cause Principle (RCCP)} \textit{If two observables $X$ and $Y$ are statistically dependent, then there exists a variable $Z$ that causally influences both and explains all the dependence in the sense of making them independent when conditioned on $Z$."}\cite{scholkopf2021toward}\end{quote}

In particular, causality represents the structural knowledge of the DGP \cite{pearl2009causality}, offering a framework to formally account for the variability and dynamic behavior of statistical relationships \cite{scholkopf2012causal}. This is illustrated in more detail in the following.

\begin{figure}[h]%
	\centering
	\subfloat[\centering Causal AI; predicting effect $Y$ from cause $X$.\label{fig:Causal} ]{{\includegraphics[scale=0.43]{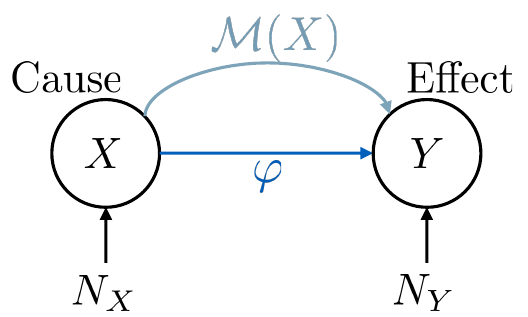} }}
	\qquad
	\subfloat[\centering Anticausal AI; predicting cause $Y$ from effect $X$.	\label{fig:AntiCausal}]{{\includegraphics[scale=0.43]{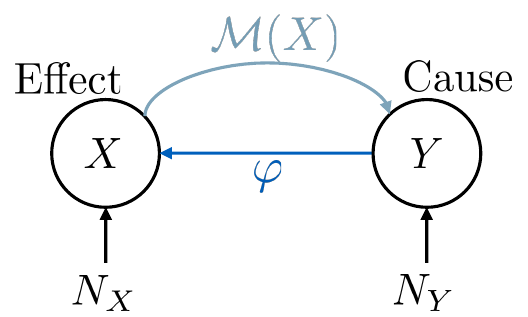} }}
	\caption{Comparison of causal and anti-causal AI systems $\mathcal{M}(X)$ based on \cite{scholkopf2012causal}. Illustrates that the input variable $X$ and the output variable $Y$, which are subject to the respective independent noise variables $N_X, N_Y$, can be both cause and effect depending on $\varphi$, the underlying deterministic mechanism.}	\label{fig:CausalAntiCausal}%
\end{figure}

The observable consequences of the DGP, the data distributions, indicate a statistical change through a change in the data distribution, also called a \textit{distribution shift} or \textit{dataset shift} \cite{pearl2009causality}. If $X$ causes $Y$ ($X \rightarrow Y$), then $P(Y|X)$ represents a causal conditional probability \cite{scholkopf2012causal}, like Newton's second law. Due to the causal relationship (Figure \ref{fig:Causal}), the mechanism is independent of the input, allowing the learned statistical relationship to generalize across input distribution shifts. This \textit{\textbf{covariate shift}} \cite{sugiyama2007covariate} is thus manageable.

However, numerous AI applications, such as object recognition, follow an anticausal effect-cause direction (Figure \ref{fig:AntiCausal}), where the effect is the input $X$ and the desired output $Y$ is the cause ($X \leftarrow Y$) \cite{scholkopf2012causal}. Thus, the statistical relationship of the causal mechanism $P(X|Y)$ is contrary to the prediction pattern $P(Y|X)$. Despite this, the causal mechanism remains independent of the causal input, hence the output of the application. This is referred to as the \textit{\textbf{target shift}} \cite{zhang2013domain}, where the statistical relationship $P(Y|X)$ to be learned is dependent on the marginal $P(X)$ according to Bayes. To summarize, Figure \ref{fig:CausalAntiCausal} visually contrasts the different causal patterns.

From an application perspective, \textit{\textbf{domain shift}} \cite{storkey2008training, takahashi2020review, zhang2013domain} is of great importance, referring to changes in data distributions from training \textit{(source)} to deployment \textit{(target)}, typically modeled as a change in $P(X, Y)$. However, a shift in the marginal distribution is usually assumed \cite{zhou2022domain}, whereby the domain shift can be reduced to a covariate shift \cite{sugiyama2007covariate, shimodaira2000improving, sugiyama2007direct}, if the statistical relationship follows a causal cause-effect structure. Nevertheless, while a marginal distribution shift is typically assumed, target shifts, conditional shifts \cite{zhang2013domain}, and their combinations \cite{schulam2017reliable} can occur, resulting in a general domain shift. Additionally, there are also distributional shifts over time, especially in relationships and semantics, which are referred to as \textit{\textbf{concept shift}} \cite{raza2013dataset, takahashi2020review}. An example is a conditional shift over time, which affects the joint distribution. Furthermore, beyond the existing challenges, the structure and causal relationship of the DGP can only be inferred from observables, with latent variables, confounders, and Simpson's paradox \cite{otte1985probabilistic, pearl2009causality} pose further challenges. Thus, theoretical knowledge has limited applicability to real tasks.

\subsection{Practical Perspective on Countermeasures}\label{sec:dataeng_prac}

One way to address these challenges is through \textit{expert knowledge}. Another, data-driven option is \textit{\textbf{causal learning}} \cite{gopnik2004theory, gopnik2007causal, holyoak2011causal, scholkopf2021toward}, where causal models are learned from observational data \cite{pearl2009causality}. Thereby, \textit{interventional data}, resulting from deliberate actions or manipulations of statistical variables or the relationship between variables, are particularly important \cite{pearl2009causality, scholkopf2021toward}.

Thus, interventions affect the DGP, causing a change in the joint distribution. These intervention data empower causal learning to uncover causal relationships. Causal relationships allow us to consider interventions that take us beyond the i.i.d. assumption, where statistical relationships no longer apply. Therefore, causal relationships are more stable than statistical ones \cite{pearl2009causality} in terms of transferability and vulnerability to dataset shifts. In addition, causal models can explicitly model interventions \cite{scholkopf2021toward}. For example, in the context of predicting the trajectories of other road users, a traffic light can be considered as an intervention. While statistical models represent a dedicated relationship, the causal model provides a set of distributions for different interventions (e.g., red or green).

Another data-based alternative to overcome the challenges is \textit{\textbf{statistical learning}} \cite{locatello2019challenging, tran2017disentangled, cheng2023disentangled, higgins2017beta}, inspired by structural causal models (SCM) \cite{pearl2009causality}. These models use a directed acyclic graph (DAG) to enable disentangled factorization \cite{kim2018disentangling}, leveraging the independent mechanism principle \cite{scholkopf2021toward}.

Another data-based, but \textit{\textbf{statistical learning}} \cite{vapnik1998statistical} alternative to overcome the challenges is \textit{disentangled feature representation learning} \cite{locatello2019challenging, tran2017disentangled, cheng2023disentangled, higgins2017beta}, which is inspired by structural causal models (SCM) \cite{pearl2009causality}. These models use a directed acyclic graph (DAG) to enable disentangled factorization \cite{kim2018disentangling}, leveraging the principle of independent mechanism stated by \cite{scholkopf2021toward}:

\begin{quote}"\textbf{Independent Causal Mechanism (ICM) Principle} \textit{The causal generative process of a system’s variables is composed of autonomous modules that do not inform or influence each other. In the probabilistic case, this means that the conditional distribution of each variable given its causes (i.e., its mechanism) does not inform or influence the other mechanisms."}\cite{scholkopf2021toward}\end{quote}

In accordance with the ICM principle the decomposition of the joint causal distribution into individual causal mechanisms is possible. As a result, only dedicated and not all relationships are needed to be adapted in the event of changes. In addition, specific mechanisms can be reused for other applications. Following this approach, learning disentangled feature representations \cite{kim2018disentangling, cheng2023disentangled} is a data-based learning method allowing for more expressive and meaningful modeling capabilities.

Nevertheless, also several advanced statistical learning methods address the i.i.d. limitations of classical statistical systems. For example, \textit{multi-task learning} \cite{caruana1997multitask, zhang2018overview} accounts for multiple i.i.d. assumptions, while \textit{zero-shot learning} \cite{pmlr-v37-romera-paredes15, 10.1145/3293318} handles unnoticed shifts shifts due to labeling changes. \textit{Domain adaptation} \cite{ganin2015unsupervised, saito2018maximum, saenko2010adapting} and \textit{transfer learning} \cite{zhuang2020comprehensive, pan2009survey} address source-target shifts using additional target data, while \textit{domain generalization} \cite{blanchard2011generalizing, zhou2022domain} methods like meta-learning \cite{hochreiter2001learning, finn2017model, vanschoren2019meta} handle versatile domain shifts with task-clustered datasets, which can be seen as interventional data. A comparison of AI categories and methods is given in Table \ref{tab:comparison_AI_method}, with further details in \cite{zhou2022domain}.

\begin{table}[h]
	\centering
	\caption{Comparison of AI categories with regard to source/target ($\mathcal{S/T}$) shifts on joint distribution ($P_{XY}$) and label space ($Y$) as well as general settings like the number of source domains/tasks ($K$) and the availability of target marginal ($P_{X}^{\mathcal{T}}$) based on \cite{zhou2022domain}.}
	\resizebox{\linewidth}{!}{
		\label{tab:comparison_AI_method}
		\begin{tabular}{p{2.9cm}  c c c  c c c}
			\toprule
			\textbf{AI Categories}  & \multicolumn{3}{c}{\textbf{Settings}} & \multicolumn{3}{c}{\textbf{Shifts}} \\
			&& K & $P_{X}^{\mathcal{T}}$ &&  $Y_{S/T}$  & $P_{XY}^{\mathcal{S/T}}$   \\
			\midrule
			Supervised Learning && $=1$ & N/A  && $=$ & $=$  \\
			Multi-Task Learning && $> 1$ & N/A && $=$& $=$   \\
			Transfer Learning && $\geq 1$ & Avail. && $\neq$ & $\neq$  \\
			Zero-Shot Learning && $= 1$ & N/A && $\neq$ & $\neq$ \\
			Domain Adaptation && $\geq 1$ & Avail.&& $=, \neq$ &  $\neq$ \\
			Test-Time Training && $\geq 1$ & Partial &&  $=$  & $\neq$ \\
			Domain Generalization && $\geq 1$ & N/A && $=, \neq$ & $\neq$  \\
			\bottomrule \\[-6pt]
	\end{tabular}}
	\newline
	\textit{Note: Avail. stands for Available, N/A stands for Not Available, and Partial indicates partial availability.}
\end{table}

Noteworthy, advanced domain generalization methods like meta-learning \cite{hochreiter2001learning, finn2017model, vanschoren2019meta} use observational data from known and unknown interventions, implicitly incorporating structural knowledge about the DGP. Thus, considering interventionals and counterfactuals \cite{pearl2009causality} is not exclusive to causal inference.Moreover, counterfactual thinking \cite{roese1994functional}, is also addressed in reinforcement learning \cite{lu2020sample, buesing2018woulda}. Nonetheless, is native to causal systems, which offer more explicit and explainable structural assumptions. However, causal systems struggle with unstructured high-dimensional data like images \cite{scholkopf2021toward} but also in low-dimensional data with limited observability of desired variables, such as friction in vehicle dynamics. In these cases, statistical methods like meta-learning \cite{hochreiter2001learning, finn2017model, vanschoren2019meta} provide alternatives. Additionally, causal systems also face challenges in tasks involving humans in the DGP, where no purely physical mechanisms exist, such as predicting road users.

\section{CONTROL ENGINEERING PERSPECTIVE}\label{sec:04_Control_Engineering_Perspective}

This section explores AI systems and their analysis and safety assurance from a control engineering perspective. First, a classification aligned with control engineering is introduced, followed by a discussion of AI system analysis in the context of control system analysis. This perspective complements the data engineering view and supports interdisciplinary efforts.

\subsection{System Classification Perspective}\label{sec:control_eng_sys_class}

Data-based signals and AI systems can be characterized by specific properties. Due to nonlinear activation functions \cite{sharma2017activation}, statistical deep learning models are inherently nonlinear, unlike the broader scope in system theory. In control engineering, classifying systems (e.g., time-varying vs. time-invariant) is crucial for system analysis and safety assurance \cite{lunze2010regelungstechnik, follinger2011laplace}. Similarly, AI systems can be classified as static or dynamic, and time-invariant or time-variant \cite{zhang2018new, an2020novel, hua2023dynamic}, though such distinctions are less common than in control engineering due to less formal definitions and overlapping of methods for desired properties and system classes. On one hand, MLP \cite{minsky1969introduction}, CNN \cite{fukushima1980neocognitron, lecun1998gradient}, GNN \cite{scarselli2008graph}, and (Variational) Autoencoders \cite{baldi2012autoencoders, kingma2013auto} can be labeled as static AI systems, while RNN \cite{rumelhart1986learning}, ESN \cite{jaeger2004harnessing}, GRU \cite{cho2014learning, jaeger2007echo} and LSTM \cite{hochreiter1997long} may be considered as dynamic AI systems. On the other hand, time invariance can be interpreted as an AI system that remains unchanged, and online or continuous learning can be viewed as time-varying AI systems \cite{qiao2014online, liu2019adaptive}. 

In practice, temporal changes are often addressed not through online learning, but by using dynamic, adaptive, or generalizing AI systems \cite{guo2012novel, pmlr-v37-romera-paredes15, vinyals2016matching, wang2020generalizing, ullrich2023cnp}. Unlike control engineering, where temporal behavior and parameters are separable \cite{foellinger1983statespace}, neural networks integrate both. AI systems can directly apply properties like memoryless or memory-containing, deterministic or stochastic, and single input single output (SISO), single input multiple output (SIMO), multiple input single output (MISO), multiple input multiple output (MIMO) structures. Overall, AI systems are nonlinear, multivariable, and often tackle complex tasks beyond classical modeling.

\subsection{System Analysis Perspective}\label{sec:control_eng_sys_pers}

On the basis of system classification, control theory offers a variety of methods for proving robustness and stability. These ideas were early on transferred to dedicated neural network architectures \cite{liao2002lmi, yang2005stability, wang2006stability, wan2010exponential, zhang2014comprehensive}, and remain a part of ongoing research \cite{kim2018standard, zhang2018overview, fazlyab2020safety, shi2020artificial, jin2020stability, hu2020reach, yin2021stability, pauli2021training}. Thereby, the methods are characterized by two main aspects: the use of selected neural network architectures and specific control engineering methods, e.g. Lyapunov stability and Lipschitz bounds.

Simultaneously, the analysis and verifiability of desired behavior in AI systems, particularly under the terms reliable, robust, trustworthy, and safe AI \cite{kaur2022trustworthy}, are gaining increasing attention in AI research. As analyzed by \cite{neto2022safety}, there is a significant increase in publications in this area. Here, as well, the majority focuses on basic and decoupled systems or specific architectures and applications \cite{neto2022safety}. At the same time, AI systems are evolving rapidly. New and increasingly complex architectures are appearing \cite{chao2022fusing, ullrich2023cnp, seff2023motionlm}. Moreover, technological initiatives such as explainable \cite{gunning2019xai, grushin2019decoding, confalonieri2021historical} or causal AI \cite{pearl2019seven, scholkopf2022causality} are emerging and offer advantages in terms of verifiability. However, the general challenges of AI safety \cite{amodei2016concrete} are extensive and apply to all AI approaches. In addition to reliability and robustness \cite{subbaswamy2018counterfactual, schulam2017reliable, subbaswamy2021evaluating}, which are close to concepts in control engineering, stability in the AI domain is defined differently from the stability known in control engineering. In the field of control engineering \cite{lunze2010regelungstechnik, follinger2011laplace, fadali2012digital}, stability is inseparable from dynamic systems, whereas in the field of AI, stability pertains so far to both static and dynamic systems, with a primary focus on dataset shift invariance \cite{subbaswamy2019preventing, subbaswamy2020development, subbaswamy2021evaluating}. 

While the previously mentioned approaches mainly consider dedicated systems, methods, or applications and thus do not offer a concrete system-theoretical counter-method for AI systems, there are the so-called safety monitors. These leverage the core idea of agnosticism of systems theory, although their methodology is less formal and more empirical. These AI system-agnostic safety monitors, depending on the concept, consider input data \cite{liu2020input, sabokrou2018adversarially} and/or AI-internal system states \cite{cheng2019runtime, henzinger2019outside} and/or output data \cite{hendrycks2016baseline, liang2017enhancing} to monitor safety at runtime without limiting the capabilities of the AI systems themselves. 
\section{INTERDISCIPLINARY ENGINEERING PERSPECTIVE}\label{sec:05_Interdisciplinary_Engineering_Perspective}

This section brings together different perspectives on the basic principle of modeling. While data-based AI systems usually consider statistical and causal learning to accomplish a task, systems theory and control engineering traditionally use concrete mathematical system descriptions. We build on Schölkopf \cite{mooij2013ordinary, scholkopf2021toward} and the idea of a strong underlying commonality between the different principles with respect to the general data-generating process (DGP) \cite{pearl2009causality}. In engineering, ordinary differential equations (ODE) are the most frequently used manifestation of the DGP. These very compact and precise descriptions of the DGP represent a counterpart to statistical models. Causal models can be interpreted as intermediate models. The linking of these modeling principles is formalized by means of a newly established corollary, postulate, and principle. Moreover, throughout the section, an abstract, general, technical example is used to illustrate the way of thinking. Furthermore, the population dynamics of the predator-prey relationship are taken into account to substantiate the corollary, the postulate, and the principle introduced. 

\subsection{Common Structural Knowledge}\label{sec:interdisciplinary_structur}
Considering, for example, the general phenomenon of wear and tear in a mechanical system. The structure of the DGP remains unaffected even if the data changes slightly. Similarly, the causal relationship between the observable variables of the DGP remains unchanged. However, the consequences of the DGP, the data distributions, and the statistical relationship may change to some extent. When considering an ODE in this example, the equations are still valid, only the parameterization may have certain errors due to wear and tear. Thus, one can postulate that the structural knowledge of the DGP is captured by the mathematically explicit systems, likewise to causal models.

This is also underlined by the counterexample. Imagine the wear gets out of control and leads to significant damage, e.g. deformation. In this case, the data-generating process is subject to change. Both the mathematical relationships in the ODE and in the causal models must be adjusted, as the models are no longer correct. 

Nevertheless, causal models have a lower level of information and can be regarded as a first level of abstraction. Hence, in terms of structural knowledge, causal learning can be viewed as a data-based counterpart to the ODE, albeit without direct physical interpretation. To formalize this, we introduce the following corollary

\textbf{Common Structural Knowledge Corollary (CSKC)} \textit{Let $\mathcal{X}$ denote a set of observable variables of an underlying data-generating process $D$ whose temporal evolution is described by an ordinary differential equation (ODE) as follows}
\begin{equation}
	\frac{dX}{dt} = F(X, t)
	\label{eq:ODE_Corollary}
\end{equation}
\textit{where $F(X, t)$ models $D$ and inherits the structural knowledge $\Sigma$ of $D$, denoted as $\Sigma(D)$. We postulate that a (learned) causal model $\mathcal{M}_{\mathrm{caus.}}$ of the observables $X$ also inherits $\Sigma(D)$, such that the corresponding structural knowledge representation of $\mathcal{M}_{\mathrm{caus.}}$, the causal graph $\mathcal{G}$, is subject to:}
\begin{equation}
	%G \text{ inherits } \Sigma
	\mathcal{G}(\mathcal{M}_{\mathrm{caus.}}(X)) \equiv \Sigma(\mathcal{M}_{\mathrm{caus.}}(X)) \subseteq \Sigma(D).
\end{equation}
\textit{In summary, the causal graph $G$ serves as an explicitly domain-independent representation of the structural knowledge $\Sigma(D)$, which is inherent to the explicit ODE describing the data-generating process, thereby facilitating broader interpretations and analyses.}

\begin{figure}[h]
	\centering	
	\includegraphics[scale=0.45]{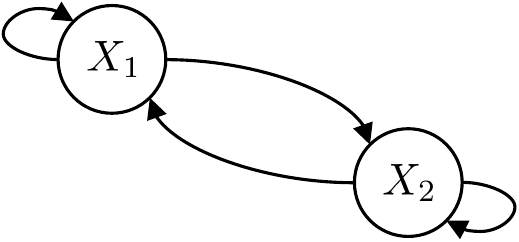}
    \vspace{3mm}
	\caption{Representation of a causal graph $\mathcal{G}$ based on the transformation of the ordinary differential equations into structural causal model according to \cite{mooij2013ordinary}.}
	\label{fig:LotkaVolterra}
\end{figure}

This corollary is further illustrated by the population dynamics of the predator-prey relationship, which is defined by the Lotka-Volterra \cite{murray2002models} equations
\begin{equation}
	\frac{dX_1}{dt} = X_1(\epsilon_{1} - \gamma_{1} X_{2}), \quad
	\frac{dX_2}{dt} = - X_2(\epsilon_{2} - \gamma_{2} X_{1}),
	\label{eq:ODE_LV}
\end{equation}
where $X_1, X_2$ denotes the number of prey and predators, respectively. In addition, $\epsilon_{1}$ represents the reproduction rate of prey and $\gamma_{1}$ the eating rate of predators. Moreover, $\epsilon_{2}$ designate the mortality rate of predators and $\gamma_{2}$ the reproduction rate of prey. A respective causal diagram of the predator-prey relationship is depicted in Figure \ref{fig:LotkaVolterra}. Although causal models are usually based on a directed acyclic graph, some approaches and extensions allow modeling of the entire dynamics (\ref{eq:ODE_LV}) besides modeling only the right-hand side of an ODE. Moreover, the three Lotka-Volterra rules \cite{murray2002models} can be thought of as structural knowledge of the data-generating process. The example demonstrates that both (\ref{eq:ODE_LV}) and the graph from Figure \ref{fig:LotkaVolterra} share common structural knowledge. Furthermore, it can be recognized that the graph only depicts the structural part, while the ODE reveals more information. In other words, a causal model can be seen as a generic abstraction with a special emphasis on the structure of the observables.

In contrast, statistical models either neglect the structural knowledge as in classical statistical models like MLPs or they incorporate it as in advanced methods that intuitively acquire the structural knowledge, as with meta-learning \cite{hochreiter2001learning, finn2017model, vanschoren2019meta}. Thus, structural knowledge can also be taken into account in selected statistical learning methodologies, but does not offer the explicit analysis as, for example, a causal graph.

\subsection{Common Parameter Considerations}\label{sec:interdisciplinary_parameter}

It is apparent that structural knowledge of a DGP can be taken into account by mathematically explicit models as well as by learned models and, in the case of causal models, can be explicitly represented by the graphs \cite{mooij2013ordinary}. At the same time, it is evident that a mathematical model such as an ODE contains more information, for example parameters. In causal models, these are neglected due to abstraction. Statistical models that map dedicated relationships integrate parameters along the overall task without explicitly determining them. The following subsection takes a closer look at the common parameter consideration. First, the general example is used. Second, a principle of commonality is introduced, and third, the principle is examined again using the Lotka-Voltera \cite{murray2002models} example.

In the case of the general technical system, e.g. a spring-mass-damper system, there are constants such as the spring stiffness or the damping constant. In the field of ODE-based modeling, system identification is used to determine the parameters. In the field of statistical learning, the parameters are part of the overall learned mapping. If the existing spring in the technical system is replaced by a harder spring, its stiffness constant must be identified. The analog in statistical learning is domain adaptation \cite{ganin2015unsupervised, saito2018maximum, saenko2010adapting} and transfer learning \cite{zhuang2020comprehensive, pan2009survey}, whereby the learned system is fine-tuned using data from the target application. In neural networks, for example, this results in modified weights and, therefore, adjusted statistical relationships. Consequently, the commonality of parameters seems to be apparent and is formalized below.

\textbf{Common Alignment Postulate (CAP)} \textit{Let $A$ denote the architecture of a statistical model $\mathcal{M}_{\text{stat}}$, and let $\mathcal{P}_{\text{stat}}$ represent the parameter set (i.e. activation function, layers, weights, etc.) that concretizes this architecture $A$. Additionally, consider $\mathcal{P}_{\text{ODE}}$ as the parameter set of an Ordinary Differential Equation (ODE). Recognizing that  $\mathcal{P}_{\text{stat}}$ and $\mathcal{P}_{\text{ODE}}$ exist in different spaces and representations, we introduce two distinct operators $\mathcal{S}_{\text{stat}}$ and $\mathcal{S}_{\text{ODE}}$, that map them in a common meaning space. Then, we postulate the common alignment:}
\begin{equation}
	\mathcal{S}_{\text{stat}}(\mathcal{P}_{\text{stat}}) \supseteq \mathcal{S}_{\text{ODE}}({\mathcal{P}}_{\text{ODE}})
\end{equation}
\textit{This common alignment postulate asserts that the architecture specification $\mathcal{P}_{\text{stat}}$ of a statistical model $M_{\text{stat}}$ includes the meaning of the parameter set $\mathcal{P}_{\text{ODE}}$ of an ODE and more. It implies that the structural configuration of the statistical model mirrors and captures the same fundamental information as the parameters of the ODE and more.}

The Lotka-Volterra example \cite{murray2002models} is taken for further illustration. When learning the relationship (\ref{eq:ODE_LV}) based on data to provide predictions, for example, a causal model is not sufficient. A statistical model such as an LSTM \cite{hochreiter1997long}, which can address the dynamics using internal memory states, could be a suitable choice. During the parameterization of the architecture and the learning process of the weights, the parameters $\gamma_1$ and $\gamma_2$ are learned implicitly. However, the parameterization $\mathcal{P}_{\mathrm{LSTM}}$ of the LSTM covers besides the individual parameters $\mathcal{P}_{\mathrm{ODE}}=\{\epsilon_{1}, \epsilon_{2}, \gamma_{1}, \gamma_{2} \}$ also structural knowledge otherwise captured by the ODE, which is why the corresponding meaning space $\mathcal{S}_{\text{ODE}}$ represents a subset of $\mathcal{S}_{\text{stat}}$. 

Furthermore, this example illustrates that the selection of the statistical model architecture reflects the requirements of the underlying structure of the data-generating process and determines the extent to which structural knowledge can be learned natively or is neglected. At the same time, it becomes clear that architecture cannot be regarded as the equivalent of structural knowledge, as this is ultimately trained into the system if the architecture permits it. 

\subsection{Common Dynamics Consideration}\label{sec:interdisciplinary_dynamics}

Previously, the commonalities of classical system modeling and learned models primarily focused on the consideration of structural knowledge and the integration and actualization of parameters. In contrast, this subsection is dedicated to the commonalities of the dynamics and, thus, the generation of an interdisciplinary view on the underlying data-generating process in general form.

In a technical system, for example, the spring-mass-damper system mentioned above, the dynamic state, which evolves over time, consists of the position and velocity and describes the current behavior and system conditions. While such a dynamical system can be modeled indirectly by a learned static statistical relationship that predicts the deviation between the current state and the subsequent state \cite{lu2018beyond, haber2017stable}, other alternatives such as recurrent neural networks (RNN) \cite{rumelhart1986learning, hochreiter1997long} consider and model the dynamics of the task at hand directly and inherent. 

Admittedly, simple static AI approaches are often preferred when implementing AI systems. However, as has been shown from the data engineering perspective, this has some drawbacks, e.g. the i.i.d. assumption is not always justified. Thus, similar to modeling with ODEs, the simplifications in AI systems also cause assumptions that cannot always be met in the real world. In particular, the desire for safe and reliable applicability requires a reduction of the basic assumptions in order to achieve better generalization. This necessitates a more detailed and accurate consideration of the underlying data-generating process.

The data-generating process and its observables appear in the form of distributions. Despite this, a large number of system descriptions such as ODEs, but also AI systems, represent deterministic systems, which is often a permissible simplification. Nevertheless, according to the underlying DGP and the data engineering perspective in Section \ref{sec:03_Data_Engineering_Perspective}, it is consistent to describe the system behavior generically in a statistical manner as a conditional input-output probability $P(Y|X)$.

A closer examination of the underlying process of data generation shows that the statistical input-output relationship itself changes over time. Consider, for example, the yaw rate of a vehicle under different environments and operating conditions. Especially in the highly dynamic boundary regime, other effects occur, causing the input-output relationships to differ from the normal driving dynamics range \cite{schramm2014vehicle}. This illustrates that the statistical input-output probability could change over time. As it turns out, statistical input-output relationships can be subject to a dynamic process. This assertion becomes clearer when considering domain generalization. Here, non-stationary environments and the different contexts and tasks result in different conditions. These different conditions may lead to different statistical relationships. Accordingly, the conditional input-output probability $P(Y|X)$ can be interpreted as an observable representation of the internal dynamic state, albeit in probabilistic terms. In other words, dynamics occur in the field of AI as well. This is formalized by the following principle. 

\textbf{Principle of Statistical Dynamics (PSD)} 
\textit{Let $P(Y|X)$ denote the conditional input-output relationship of a task to be learned. It can be stated that this statistical input-output relationship is subject to a dynamic process}  
\begin{equation}\label{PSD}
	\resizebox{\linewidth}{!}{$\frac{dP(Y|X)}{dt} = F(\mathcal{X} , \mathrm{CAU} , \mathrm{LV}, \mathrm{CF}, I, \mathrm{CTX}, \mathrm{TSK}, E, t)$.}
\end{equation}

\textit{where $\mathcal{X}$ denotes observable input variables, $\mathrm{CAU}$ the underlying (partially hidden) causal relationship, $\mathrm{LV}$ (unknown) latent variables, $\mathrm{CF}$ unknown and/or unobservable confounders, $I$ interventions, $\mathrm{CTX}$ (varying) context conditions, $\mathrm{TSK}$ task conditions, $E$ (non-stationary) environments and $t$ time in general.}

To illustrate this, the example of the predator-prey relationship (\ref{eq:ODE_LV}) is once again drawn upon. In this case, the statistical input-output relationship is the conditional prediction of the populations, i.e. $X_1$ and $X_2$, based on past and current population data. The DGP corresponding statistical relationship is subject to a variety of factors. For example, causality is described by assuming the three Lotka-Volterra rules \cite{murray2002models}. A change, such as the occurrence of a second predator population, would cause a significant change in the causal relationship as well as the statistical dynamics. Other influencing factors can, be latent variables and unknown or unobservable confounding factors. For example, factors that have an impact on reproduction rates but are neither observable nor known. Moreover, interventions could be interpreted as a human manipulation of a specific population. Furthermore, the general context, such as seasons, could also have an effect on the statistical relationship. In contrast, task-specific changes are not to be expected in the present example, since no concrete control variables exist. Although a change in the predator's food consumption behavior would be conceivable, albeit this could also be attributed to causality, illustrating that influencing factors cannot always be clearly separated. This is also the case with the influencing factors like environment and time. Generally, the principle introduced also serves to demonstrate possible influencing factors of the dynamic statistical relation in a generic, yet formalized way. This generic abstraction of the data-generating process enables clear documentation of the assumptions associated with the choice of AI architecture as well as the input and output variables of an AI system. Thus, gaps in the system development can be reduced.

\section{CONTROL THEORY INSPIRED AI SYSTEMS CLASSIFICATION}\label{sec:06_AI_System_Classification}

In contrast to the previous section, which focused on the similarities of modeling approaches and the underlying process, this section is dedicated to AI systems. Thereby taking a closer look at the classification of AI systems while drawing inspiration from control theory and the previously introduced formalization of the statistical input-output relationship using conditional probability. Furthermore, the system-oriented view is characterized by the discrete-time perspective, which is opposed to the continuous time of the underlying process. In the following, the dynamic, non-stationary, and static AI system classes are introduced, which represent different basic assumptions w.r.t. the variability of (\ref{PSD}), the underlying process.

\subsection{Dynamic AI System}
According to control theory, dynamical systems are present if the system incorporates memory states. On the one hand, memory states can exist within AI systems in the form of compact states or variables, as is the case of recurrent \cite{rumelhart1986learning, hochreiter1997long} or autoregressive models \cite{huang2018neural, kaiser2018fast}, for example. On the other hand, memory states in AI systems can also be available in the form of collected data, which can be used to adapt the system online, e.g., through online reinforcement learning \cite{wei2017online}. However, a data memory that only collects data but does not exploit it online for adjustments is not a memory state in terms of dynamics. We define a dynamic AI system in line with these ideas. 

\textbf{Definition 2.2 (Dynamic AI System)}\label{def:dynamicAI} \textit{Let $\mathcal{M}$ represent an AI model with the statistical input-output relationship $P(Y|X)$. The model is deemed a dynamic AI system if the statistical relationship undergoes temporal changes}
\begin{align}\label{eq:dynamicAI}
	\begin{split}
		{P(Y|X)}_{t+1} &= f_{\mathrm{dyn}}({P(Y|X)}_{t}, h_{t}), \\
	\end{split}
\end{align}
\textit{upon an internal memory state $h_{t}$}
\begin{align}\label{eq:dynamicAI_mem}
	\begin{split}
		{h}_{t+1} &= g(h_{t}, X_{t}), 
	\end{split}
\end{align}
\textit{which is continuously revisited and updated by the inputs $X_t$.} 

According to the definition, a dynamic AI system evolves depending on an internal memory state that is itself updated. This means that the system behavior, similar to dynamic systems in control engineering, depends on the current and past inputs. This requires the past and previous inputs of such an AI system to be considered part of the system analysis, which is why this classification is introduced here.

To exemplify the validity of the definition above, the RNN from \cite{choe2017probabilistic} is defined as 
\begin{equation}
	\begin{aligned}
		{h}_{s} &= f_{\alpha}(h_{s-a}, x_{s}) \quad \forall s = 1, \dots, t \\
		{x}_{t+1} &= g_{\beta}(h_{t}), 
	\end{aligned}
\end{equation}
where $h_s$ denotes hidden states, $x_s$ sequence input data and $f_{\alpha}, g_{\beta}$ nonlinear functions parametrized by $\alpha, \beta$ respectively, is considered. As shown by \cite{choe2017probabilistic}, a fully probabilistic graphical model representation can be derived from the generative RNN such that the conditional probability of the RNN $P(x_{t+1}| x_{\leq t})$ can be described and reformulated as
\begin{equation}
	\begin{aligned}
		P(x_{t+1}| x_{\leq t}) &= P(x_{t+1}| h_{t}) \\
		&=  P(x_{t+1}| f_{\alpha}(h_{t-1}, x_t))\\
		&=  P(x_{t+1}| f_{\alpha}(h_{t-1}, g_{\beta}(h_{t-1}))\\
		&=  P(x_{t+1}| f_{\alpha}(h_{t-1}, P(x_t| h_{t-1})),\\
	\end{aligned}
\end{equation}
whereby conditional independence can be exploited. Aligning the probabilistic interpretation with $P(Y|X)$, the general conditional input-output probability shows
\begin{equation}
	\begin{aligned}\label{eq:rnn_conversion}
		P(Y|X)_{t} &= P(x_{t+1}| x_{\leq t}) = P(x_{t+1}| h_{t}) \\
		&=  P(x_{t+1}| f_{\alpha}(h_{t-1}, P(x_t| h_{t-1}))\\
		&=  P(x_{t+1}| f_{\alpha}(h_{t-1}, P(Y|X)_{t-1}))\\
		&\approx   f_{\mathrm{dyn}}(P(Y|X)_{t-1}, h_{t-1}),\\
	\end{aligned}
\end{equation}
The validity of the dynamic AI system definition is determined using the RNN example. Thereby, particularly the last conversion in (\ref{eq:rnn_conversion}) enables a generic abstraction towards the general consideration of various dynamic systems. Besides \cite{choe2017probabilistic, bitzer2012recognizing} also shows the successful fusion of different perspectives to derive probabilistic interpretations of dynamic AI systems, albeit from a biological cybernetic background. 

Furthermore, the internal memory $h_t$ in (\ref{eq:dynamicAI_mem}) can be considered as internal dynamics based on \cite{bitzer2012recognizing}, while (\ref{eq:dynamicAI}) can be labeled as input-output dynamics. While \cite{choe2017probabilistic} and \cite{bitzer2012recognizing} span the analogies between RNNs and hidden Markov models \cite{eddy1996hidden}, state-space models \cite{baum1966statistical}, Kalman-Bucy filters \cite{jazwinski2007stochastic}, particle filters \cite{doucet2003parameter, djuric2003particle} and underpin the interdisciplinary engineering perspective in Section \ref{sec:05_Interdisciplinary_Engineering_Perspective}, RNNs have not been interpreted in the context of Byrnes-Isidori standard form. Nevertheless, both the internal dynamics and the input-output dynamics indicate similarities. For instance, both assume that the internal dynamics of the internal memory state of the dynamical system are unobservable. Furthermore, as explained in Section \ref{sec:04_Control_Engineering_Perspective}, AI systems and, thus also, dynamic AI systems are non-linear systems. Beyond that, the interpretation of the relative degree is also conceivable with respect to the recursive elements within an RNN, for example. Finally, it should be mentioned that the derivatives of the system output, which are an integral part of the Byrnes-Isidori standard form, represent pivotal elements for the analysis of black-box AI systems. As mentioned in Subsection \ref{sec:control_eng_sys_pers}, there are less formalized approaches that take a closer look at the output behavior and its change \cite{hendrycks2016baseline, liang2017enhancing}. These similarities support the efforts of the interdisciplinary perspective and open up a multitude of further research questions, such as the existence examination of zero dynamics in AI systems. However, this work aims to create an interdisciplinary and generic basis for the safety analysis and safeguarding of AI systems. This is why these further research questions are reserved for future work. An overview of possible dynamic AI systems is given below.

The internal memory can be different. It can be integrated into the system in several ways. Table \ref{tab:dynamicAI} lists a selection of AI approaches and their memory. Concerning safety reasoning and system analysis, dynamic AI systems can be classified as systems that cannot be analyzed apart from their previous exploitation history.

\begin{table}[ht]
	\centering
	\caption{Illustration of a selection of dynamic AI systems}
	\resizebox{\linewidth}{!}{
		\begin{tabular}{p{1.6cm}p{5cm}}
			\toprule
			\textbf{AI Approach} & \textbf{Memory State $h_t$} \\
			\midrule
			Recurrent Neural Networks (RNN) & RNNs have hidden states that act as memory $h_t$. The hidden state is updated at each time step based on the current input and the previous hidden state. In this way, information about the past is preserved. In addition, some approaches, such as the LSTM, have additional explicit memory cells that serve as long-term memory.  \cite{rumelhart1986learning, hochreiter1997long} \\ \midrule
			Temporal Difference (TD) Learning & While no special memory is used in TD learning, the corresponding value function or Q-value function can be referred to as a memory, as the values are successively updated. In this way, previous experience is considered in the current estimates. Therefore, the value function can be interpreted as a memory $h_t$. \cite{sutton2018reinforcement} \\  \midrule
			Neural Turing Machines (NTM) & NTMs essentially consist of a neural network, a network controller, and a memory bank. The memory bank is readable and writable via respective heads by the network controller. The network controller can be either a feedforward model or a recurrent neural network. The memory $h_t$ can, therefore, be either the memory bank or an internal hidden state in the network controller.  \cite{graves2014neural} \\  \midrule
			Differential Neural Computers (DNC) &  DNCs are extensions of NTMs with more intelligent use of memory through attention mechanisms. These mechanisms are used to search for content through an associative data structure, temporal links for sequential processing, and the allocation of unused memory for writing. \cite{graves2016hybrid} \\ \midrule
			Neural Attention Memory (NAM) & NAMs are attention mechanisms as readable and writable memory networks and build dynamic Long Short-term Attention Memory Networks or NAM Turing Machines. Networks that employ NAMs as $h_t$ and whose conditional probability is time-varying can be considered dynamic AI systems. \cite{nam2023neural} \\ 
			\bottomrule
	\end{tabular}}
	\label{tab:dynamicAI}
\end{table}

Table \ref{tab:dynamicAI} provides an insight into the various characteristics of memories. Some of these approaches such as NTM \cite{graves2014neural}, DNC \cite{graves2016hybrid} or memory networks \cite{weston2014memory} are also referred to as Memory Augmented Neural Networks (MANN). On the one hand, the approaches listed in Table \ref{tab:dynamicAI} also include other AI approaches such as TD Learning \cite{sutton2018reinforcement} or RNNs \cite{rumelhart1986learning, hochreiter1997long}. On the other hand, approaches such as Hopfield networks \cite{hopfield1984neurons, krotov2016dense} or attention-based \cite{vaswani2017attention} transformers, which could be classified as MANNs, do not fall under our classification. 

Hopfield networks \cite{hopfield1982neural, hopfield1984neurons, krotov2016dense} are known as auto-associative memory for pattern retrieval. Although the neuron wiring incorporates feedback loops and the recall process represents an iterative energy minimization pattern retrieval, the network does not fit into the classification of dynamic AI systems. On the one hand, the memory is pre-trained and static during deployment. On the other hand, the underlying dynamic process that exists during recall is only present during an individual retrieval but does not represent a dynamic process of the statistical dynamics as required for a dynamic AI system. Also, from the analysis point of view, the input-output behavior does not change depending on the history. 

Moreover, one of the central building blocks of transformers is the attention mechanism \cite{vaswani2017attention}, often referred to as memory. This is based on various aspects. On the one hand, attention mechanisms can be interpreted as successors to RNN \cite{rumelhart1986learning}, LSTM \cite{hochreiter1997long}, and GRU \cite{cho2014learning} in terms of their field of application. On the other hand, human attention requires memory to be effective. However, the usual attention mechanisms have neither a readable and writable memory nor an internal hidden state that is updated over time.
In contrast, the attention mechanism depends on the given inputs. Consequently, the system behavior is not dependent on the previous usage. Thus, even if the statistical input-output relationship evolves, the required memory according to Definition 2.2 (\ref{def:dynamicAI}) is not given. Neural attention memory \cite{nam2023neural}, on the other hand, fulfills the conditions of Definition 2.2  (\ref{def:dynamicAI}), which illustrates that a straightforward classification based on existing method classes such as transformers is not sufficient. Instead, the specific architecture needs to be analyzed for the behavior-oriented classification of AI systems. 

To bridge the gap towards an application-oriented perspective, the area of intelligent transportation systems (ITS) and automated driving (AD) is covered. Therefore, Table \ref{tab:dynamicAI_AD} provides an overview of dynamic AI systems used for different tasks in AD.  Here, the increasing complexity of dynamic AI systems particularly accounts for realistic temporal relationships.

% Within this field, dynamic AI systems sind in vielen Bereichen gewünscht, wie in Tabelle X aufgeführt

% Letztlich sind die aktuellen Aspekte mehr non-stationary, da kein richtigen buffer! bei der aktuellen Entwicklung wird der übergang fließend, Die Ansätze addressieren temporal information /context consideration, aber sind mehr im klassischen Sinne transformers, sprich weniger richtige state like memory, more retrival based data memory, wo also eingruppieren? Oder auf einfacherer Aspekte/Ansätze gehen RNN based scene modelling prediction, planning?

% Our requirement, internal memory, that si revised and updated?

% Safe: 
% \cite{song2024motion} ist richtig gut vereieint bereits den aspekte von scene modelling and prediction ! - Is to complex and confusing!

%https://openaccess.thecvf.com/content/ICCV2021/papers/Chen_Learning_To_Drive_From_a_World_on_Rails_ICCV_2021_paper.pdf 
% Aber findet hier ein update start?

% how should we think about temporal self-attention? No!

% Scene as Occupancy \cite{tong2023scene} (is  bit edgy)
% cascaded voxel decoder
% voxel-based temporal self-attention
% 3D deformable attention

% 3D deformable attention applied to voxel-based temoral self-attention for cascaded voxel decoder

% BEVFormer: \cite{li2022bevformer} Temporal information is crucial in human visual perception 

% \cite{pang2023streaming, song2024motion}

\begin{table}[ht]
	\centering
	\caption{Application illustration of dynamic AI systems in AD.}
	\resizebox{\linewidth}{!}{
		\begin{tabular}{p{1.2cm}p{1cm}p{4.5cm}}
			\toprule
			\textbf{Task} & \textbf{Method} & \textbf{Description} \\
			\midrule
            Motion Prediction & Social LSTM \cite{alahi2016social}& Multiple LSTMs, combined with a social pooling layer for information sharing, predict human motion dynamics in crowded scenes. \\ \midrule 
            Trajectory Planning & Imitative Models \cite{Rhinehart2020Deep}& Trajectory planning using a deep imitative model with an RNN-based structure that captures forward dynamics in a probabilistic latent state representation  \\ \midrule 
            Scene Understanding & MILE \cite{hu2022model}& Scene evolution is imagined using a latent dynamics model implemented via gated recurrent cells, which is integrated into an inference model for state estimation and a generative model for future prediction \\  
			\bottomrule
	\end{tabular}}
	\label{tab:dynamicAI_AD}
\end{table}

\subsection{Non-stationary AI System}
Dynamic AI systems represent a system class that considers the past using internal memory and thus dynamically responds to changes in the data-generating process. At the same time, this requires a more complex system analysis, as the system cannot be decoupled from its previous use. This is in contrast to non-stationary AI systems, which, unlike dynamic systems, do not have an internal memory. Instead, relevant information required to adjust the statistical input-output relationship is transferred via the current input. Non-stationary AI systems are defined accordingly in the following.
\newpage

\textbf{Definition 2.3 (Non-stationary AI System)} \textit{Let $\mathcal{M}$ represent an AI model with the statistical input-output relationship $P(Y|X)$. The model is considered a non-stationary AI system if i) the statistical relationship is subject to temporal changes as a function of the input and ii) the model has no internal memory states $h_{t}$. Accordingly, a non-stationary AI system can be described by}
\begin{equation}
	{P(Y|X)}_{t} = f_{\mathrm{ns}}(X_{t}), 
\end{equation}
\textit{the conditional probability at time $t$, which depends according to $f_{\mathrm{ns}}$ on the current input $X_t$.}

The definition states that a non-stationary AI system's statistical input-output relationship depends on the input. While the output of AI systems generally depends on the input, most traditional AI systems have a fixed mapping where the conditional probability remains unchanged over time. In contrast, advanced approaches such as transformers take input sequences and dynamically adjust the internal weights \cite{vaswani2017attention} and thus the input-output relationship. However, the behavior across input sequences is independent of previous usage. Another similar approach is meta-learning \cite{garnelo2018conditional} that, for example, uses previous input-output pairs as current contextual input and adjusts the input-output relationship accordingly. However, there are no internal dynamics or memory, so using a non-stationary AI system is independent of the previous or subsequent use and allows the adaptation of the statistical input-output probability. 

This demonstrates that systems in this class impose different requirements on system analysis than dynamic AI systems or traditional systems with a fixed input-output mapping. The importance of the distinction becomes particularly clear in the context of empirical safety monitors \cite{ferreira2021benchmarking}. While static systems exploit a change in the input-output relationship to detect undesirable behavior, non-stationary AI systems require an adjustment of the input-output relationship, e.g., depending on the context. Table \ref{tab:nonstationaryAI} provides an overview of non-stationary AI systems and how their input can be designed and influence the statistical input-output relationship.

\begin{table}
	\centering
	\caption{Illustration of non-stationary AI systems}
	\resizebox{\linewidth}{!}{
		\begin{tabular}{p{1.6cm}p{5cm}}
			\toprule
			\textbf{AI Approach} & \textbf{Input $X_t$} \\
			\midrule
			Conditional Neural Processes (CNP) & CNPs \cite{garnelo2018conditional} represent a meta-leraning approach that allows additional information to be provided via a context dataset as input $X_t$, such that the statistical input-output relationship can be adapted to the given context. This results in a non-stationary statistical input-output relationship without internal memory and internal dynamics.  \\ \midrule
			Attention-based Transformers &  The attention mechanism \cite{vaswani2017attention} enables transformers to recognize dependencies and relationships within the input sequences $X_t$. This enables transformers to be context-dependent and adaptive. The statistical input-output relationship is therefore flexible in time, dependent on the input, and has no internal memory and consequently no internal dynamics.\\
			\bottomrule	\end{tabular}}
	\label{tab:nonstationaryAI}
\end{table}

Non-stationary AI systems can, therefore, also address dynamic processes and tasks. However, due to non-stationarity and the simultaneous absence of dynamics within the AI system, function evaluation and analysis are considerably simplified, as the system itself has no memory states, and the evaluation can, therefore, be carried out statically using dedicated inputs. Dynamic AI systems are more challenging to analyze due to their internal memory states. For this reason, AI systems have often been designed as static mapping of the right-hand side of the differential equation \cite{lu2018beyond, haber2017stable} to exploit invariance and increase generalizability.

To bridge the gap towards an application-oriented perspective, the area of intelligent transportation systems (ITS) and automated driving (AD) is covered. Therefore, Table \ref{tab:dynamicAI_AD} provides an overview of dynamic AI systems used for different tasks in AD.  Here, the increasing complexity of dynamic AI systems particularly accounts for realistic temporal relationships.

In order to provide an application-oriented context for the given system class, Table \ref{tab:nonstationaryAI_AD} presents a selection of non-stationary AI systems across different AD tasks. Thereby, it becomes evident that in practice, dynamic processes are often modeled in a simplified way as non-stationary processes. 

\begin{table}
	\centering
	\caption{Application illustration of non-stationary AI systems in AD.}
	\resizebox{\linewidth}{!}{
    \begin{tabular}{p{1.2cm}p{1cm}p{4.5cm}}
			\toprule
			\textbf{Task} & \textbf{Method} & \textbf{Description} \\
			\midrule
            Object Detection & DETR \cite{carion2020end}& Directly predicts a final set of bounding boxes and categories based on a common CNN and a transformer architecture. \\ \midrule
            Motion Prediction & MTR \cite{shi2022motion} & Trajectory predictions are provided via a motion transformer that optimizes global intention and local motion jointly. \\ \midrule 
            Trajectory Planning & Planning Transformer \cite{chen2024vadv2}&  Probabilistic trajectory planning is performed using a transformer that leverages a planning vocabulary and environment token embeddings to model a non-stationary stochastic process conditioned on the environment.\\  
			\bottomrule
	\end{tabular}}
	\label{tab:nonstationaryAI_AD}
\end{table}

\subsection{Static AI System}

Many of today's AI systems have a static mapping of the input-output relationship. Systems that can be assigned to this class are the easiest to analyze compared to the previous systems and are defined as follows.

\textbf{Definition 2.4 (Static AI System)} \textit{Let $\mathcal{M}$ represent an AI model with the statistical input-output relationship $P(Y|X)$. The model is referred to as a static AI system if the statistical relationship}
\begin{equation}
	\forall t: {P(Y|X)}_{t+1} = {P(Y|X)}_{t}
\end{equation}
\textit{is static for all times. The underlying dynamic process is therefore assumed to be stationary: $\frac{dP(Y|X)}{dt} = \text{constant}$.}

As already mentioned in Subsection \ref{sec:dataeng_theo}, static AI systems are suitable for static statistical relationships that are generated, for example, by physical generation mechanisms. Accordingly, the principle of statistical dynamics from Subsection \ref{sec:interdisciplinary_dynamics} has a small number of influences in this case. In addition to classical static AI methods such as multi-layer perceptrons (MLP) \cite{rosenblatt1958perceptron}, this class also includes more advanced AI methods such as few-shot learning \cite{wang2020generalizing, kadam2020review} or disentangled representation learning \cite{locatello2019challenging, tran2017disentangled, cheng2023disentangled, higgins2017beta}. Even if these approaches can be applied concerning domain generalization like meta-learning, they differ for the system class. This is because, compared to meta-learning, these approaches strive for a generalized input-output relationship that can be applied across previously unseen data. In other words, the approaches attempt to generate an AI system that is as robust as possible and independent of influencing factors through a sophisticated static AI system design. Here, robustness refers to unchanged functionality in the face of system or process disturbances. Static AI systems thus represent a counter-design to dynamic and non-stationary AI systems by making the influencing factors of the principle of statistical dynamics robust. A more detailed list of static AI systems, as provided within non-stationary and dynamic AI systems, is omitted due to its clarity.

From an application-oriented perspective, static AI systems are often used in AD. Examples include methods such as YOLO \cite{redmon2016you}, Fast R-CNN \cite{girshick2015fast}, and K-means clustering-based image segmentation, which are frequently used for tasks such as object recognition and scene analysis.

From an application-oriented perspective, static AI systems are commonly employed in automated driving. Methods like YOLO \cite{redmon2016you}, Fast R-CNN \cite{girshick2015fast}, and K-means clustering-based image segmentation \cite{dhanachandra2015image} are frequently used for tasks such as object recognition and scene analysis. This highlights that, in addition to a task-based perspective, a system-oriented classification provides a clearer understanding of the specific characteristics of the AI systems used, allowing for an enhanced and systematic analysis and evaluation of AI systems in accordance to their system classification.

\subsection{Differentiation to Control Systems}

Beyond the definition of static, non-stationary \& dynamic AI systems, the concept of disturbance variables $d$ and control variables $U$ of control engineering can be transferred. Control variables $U$ are intentional actions. Causal interventions $I$ are deliberate actions or manipulations. Just as control variables in control engineering represent deliberate interventions to modify the state, interventions are deliberate or known manipulations that modify the statistical relationship. Therefore, $U \equiv I$ can be stated from a conceptual point of view. In addition to this analogy, however, it is also important to point out the differences. Control variables can be chosen to achieve a desired state, while interventions usually occur rather than being freely selectable. From a control engineering perspective, this would suggest an analogy to the disturbances. However, systems should be robust against disturbances. Control variables and interventions, on the other hand, are concrete inputs that have an impact, which is why the analogy described above is more appropriate. At the same time, this again illustrates the general differences between control systems and AI systems. AI systems are integrated into the environment to fulfill a task in the best possible way. Controllers, on the other hand, have the explicit task of influencing the process. This demonstrates that the analogy is also preserving the respective characteristics. Furthermore, unknown interventions $(\mathrm{UI})$ in observational data can be reinterpreted as unknown disturbance variables $d$. Other disturbance variables $d$ are unobservables, latent variables $(\mathrm{LV})$ and confounders $(\mathrm{CF})$.

\subsection{Future Prospects Regarding System Developments}
From the perspective of cyber-physical systems, changes are to be expected, mainly due to the non-stationary nature of the real-world environment. AI systems in use have to deal with these changes. It can be seen that classic statistical learning methods, which are only reliable in the area of the i.i.d. assumption, may not lead to the desired autonomy. The classical statistical learning methods can be regarded as static systems that play a subordinate role from the perspective of cyber-physical systems. Beyond this, dynamic, non-stationary, or static generalizing AI systems are necessary for human-like intelligence. This development is undoubtedly desirable from a cognitive perspective. Namely in terms of increased demanded autonomy to work in imagined space in the sense of Konrad Lorenz \cite{lorenz1973ruckseite}. Or, from Yann LeCun's perspective \cite{lecun2022path}, to achieve human-like intelligence via world models.

\section{AI SYSTEM PROPERTIES}\label{sec:07_AI_System_Properties}

While until now an interdisciplinary basis has been created and AI system classes have been defined in terms of system analysis, this section is concerned with the definition of AI system properties that enable a description and evaluation of the AI system's behavior. The three basic properties of robustness, sensitivity, and stability are introduced below, taking into account the AI system definitions to enable targeted development, validation, approval, monitoring, and maintenance of AI systems.

\subsection{AI Circumstance Robustness}

Considering the various influencing factors of the principle of statistical dynamics (\ref{PSD}) in Subsection \ref{sec:interdisciplinary_dynamics}, it can be stated that AI systems should be robust, e.g., against latent variables and confounders. According to \cite{subbaswamy2018counterfactual, subbaswamy2023causal}, this does not mean that the AI system must be designed in such a way that no latent variables or confounders exist, but that no confounders or latent variables are active in the specified application, which according to \cite{subbaswamy2018counterfactual, subbaswamy2023causal} corresponds to the idea of non-active unstable edges. This idea should be taken up in the definition of the robustness property, which is why the robustness is defined with regard to specified circumstances.

\textbf{Definition 3.1 (AI Circumstance (AIC) Robustness)} \textit{Denote an AI system as a model $\mathcal{M}$ with inputs $X$, outputs $Y$ and parameters $\theta$ that can be used in the environment $\mathcal{E}$ and whose performance can be evaluated using a loss $\ell$. The AI system is said to be circumstance robust, if for any two circumstances, $\gamma_{1}, \gamma_{2}$, that are instantiations of a pre-specified set $\Gamma$ of required robustified influencing factors, where each individual required robustified influencing factor is within the pre-specified range $\forall \gamma_j \in \Gamma_j : \gamma_{j, \mathrm{low}} < \gamma_j < \gamma_{j, \mathrm{up}}$, the statistical input-output relationship
	\begin{equation}
		P_{\gamma_{1}}(Y|X) = P_{\gamma_{2}}(Y|X)
	\end{equation}	
	holds. Thus, the corresponding pre-specified set of influencing factors $\Gamma$, assumed to be within pre-specified bounds $\mathcal{R} = \{(\gamma_{1,\mathrm{low}}, \gamma_{1,\mathrm{up}}), \ldots, (\gamma_{n,\mathrm{low}}, \gamma_{n,\mathrm{up}})\}$, does not impact the statistical input-output relationship $P(Y|X)$. \newline $\Rightarrow$ It can be stated that the performance of the AI system $\mathbb{E}_{\Gamma} [\ell(\mathcal{M}(\theta; X))]$ remain as specified. Consequently, the AI system is robust w.r.t. $\Gamma$.}

This allows us to directly include the AI requirement specification in the synthesis and analysis via $\Gamma$ and $\mathcal{R}$. For example, the tolerances of the components used can be taken up directly. Any assumptions, e.g. about the signal noise, are thus explicitly defined and documented.  

This type of robustness is of general relevance and can be applied to static, non-stationary, or dynamic AI systems. Furthermore, neither a DAG nor low-dimensional data is required, compared to \cite{subbaswamy2023causal}. Thus, the required flexibility is accommodated. In addition, AIC robustness is also helpful for security and adversarial defense aspirations. Especially since the methods normally used to prevent attacks, e.g. out-of-distribution detection, are likely to fail in the event of minimal deviations. 

In general, it should be mentioned that this definition differs by explicitly specifying and investigating the system behavior under pre-specified circumstances. Compared to the prevailing, strictly performance-oriented definitions, this offers the key advantage of being able to verify and monitor the behavior alongside the performance analysis, thus enabling a more adequate system analysis.

To illustrate the connection to real world applications, Table \ref{tab:prop_example_Robustness} presents a systematic overview of various task-specific application domains in AD, highlighting benefits of AIC robustness across them.

\begin{table}[h!]
	\centering
	\caption{Overview of AIC robustness across task-specific domains in automated driving.}
	\resizebox{\linewidth}{!}{
    \begin{tabular}{p{1.2cm}p{1cm}p{4.5cm}}
			\toprule
			\textbf{Task} & \textbf{Method} & \textbf{Description} \\
			\midrule
            Perception & Object Detector & For a comprehensive and reliable real world use, the statistical input-output relationship of an object detecor, e.g., \cite{zhu2020deformable}, should be robust to sensor signal noise $\gamma_{\mathrm{noise}}$, sensor drift $\gamma_{\mathrm{drift}}$, and adversarial attacks $\gamma_{\mathrm{per.attack}}$, e.g., universal perturbations \cite{li2021universal}.\\ \midrule
            Scene Understanding & Motion Prediction  & The statistical input-output relationship of a motion prediction AI, e.g., \cite{shi2022motion}, should be robust to perception noise $\gamma_{\mathrm{noise}}$, e.g., slight LIDAR intensity variation due to fog, and bounded inaccuracies $\gamma_{\mathrm{inacc.}}$, e.g., minor errors in bounding boxes or small localization inaccuracies.
            \\ \midrule
            Planning & Decision-making  & 
            To ensure a consistent behavior, the statistical input-output relationships of an decision-making AI system, e.g., \cite{li2018humanlike}, should be robust across $\gamma_{\mathrm{scen.}}$, a set of structurally similar driving situations. \\
			\bottomrule
	\end{tabular}}
	\label{tab:prop_example_Robustness}
\end{table}

\subsection{AI Circumstance Sensitivity}

While AIC robustness makes it possible to specify the robustness of AI systems against certain influences, e.g. signal noise, and to analyze them on this basis, the principle of statistical dynamics also entails influences that an AI system may be required to consider. In particular, the AI system classes of dynamic and non-stationary AI systems inherit the property of a changing statistical input-output relationship in order to accommodate influences. Circumstances to which an AI system must be responsive can be divided into internal conditions $(\mathrm{TSK})$ and external conditions $(\mathrm{CTX})$ or interventions $(I)$. In order to specify and analyze this responsiveness, we introduce the AIC sensitivity.

\textbf{Definition 3.2 (AI Circumstance (AIC) Sensitivity)} \textit{Denote an AI system as a model $\mathcal{M}$ with inputs $X$, outputs $Y$ and parameters $\theta$, applicable in the environment $\mathcal{E}$. The performance of the AI system is evaluated using a loss function $\ell$. Given a pre-specified set $\Lambda$ of $j$ required sensitivity influencing factors $\lambda$, each with an associated pre-specified sensitivity threshold $\tau$, sensitivity ratio $\alpha$ and sensitivity ratio tolerance $\epsilon$, then the AI system is said to be circumstance sensitive if $\forall \lambda_{j} \in \Lambda, \forall \lambda_{j,m} \in \Lambda_{j}, \forall \lambda_{j,n} \in \Lambda_{j}, n \neq m: |\lambda_{j, m} - \lambda_{j, n}| = \delta_{j,m,n} > \tau_{j}$ there exists a change in the statistical input-output relationship}
\begin{equation}
	\Delta P_{\lambda_{j,m,n}}(Y|X) = D_{\mathrm{KL}}(P_{\lambda_{j, m}}(Y|X) || P_{\lambda_{j, n}}(Y|X))  
\end{equation}	
\textit{measured by the KL-divergence such that}
\begin{equation}
	(\alpha_{j} - \epsilon_{j})  \frac{\delta_{j,m,n}}{\tau_j} < \Delta P_{\lambda_{j,m,n}}(Y|X) < (\alpha_{j} + \epsilon_{j})\frac{\delta_{j,m,n}}{\tau_j}
\end{equation}	
\textit{holds. Thus, it is assumed that the AI system responds as pre-specified by $\mathcal{S} = \{(\lambda_{j,\tau}, \lambda_{j,\alpha}, \lambda_{j,\epsilon})\}$ for all pre-specified influencing factors $\lambda_j \in \Lambda$ within desired ranges  $(\alpha_j \pm \epsilon_j)\delta_{j} | \delta_{j} > \tau_{j} $. \newline $\Rightarrow$ Consequently, the AI system is sensitive w.r.t. $\Lambda$ as desired and the performance of the AI system $\mathbb{E}_{\Lambda} [\ell(\mathcal{M}(\theta; X))]$ can be assumed to remain as specified.}

Here as well, the desired behavior is explicitly defined and documented via $\Lambda$ and $\mathcal{S}$, allowing the required and intended functionality to be analyzed and verified. For instance, a situation can be easily modified and analyzed to determine if the required sensitivity is present without looking into the AI system. In automated driving, for example, the behavioral decision can be checked depending on a slight change in the behavior of a child on the sidewalk. In general, AIC sensitivity thus offers a generic methodological approach that can be used, e.g., to analyze scenarios in the context of automated driving development. While AIC robustness can be applied to any AI, AIC sensitivity focuses on non-stationary and dynamic AI systems that are able to react to changing circumstances. Depending on the system class, the sensitivity needs to be carried out taking the past into account. 

\begin{figure}[h!]
	\centering	
	\includegraphics[scale=0.5]{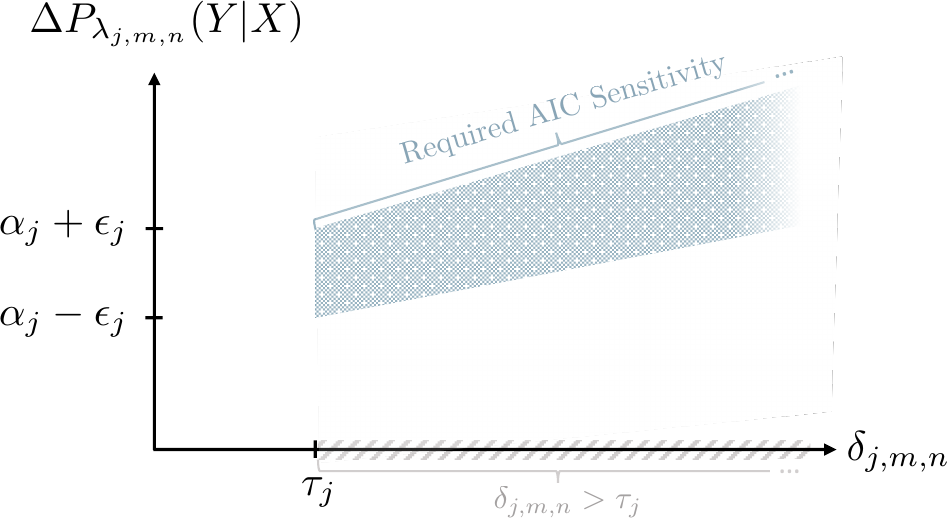}
	\caption{Illustration of AIC sensitivity.}
	\label{fig:AIC_sensi}
\end{figure}

The AIC sensitivity is illustrated visually in Figure \ref{fig:AIC_sensi}. Furthermore, to underscore its practical relevance, Table \ref{tab:prop_example_Sensi} highlights automated driving tasks that stand to benefit significantly from AIC sensitivity. A comprehensive analysis is beyond the scope of this work and is left for future research.

\begin{table}[h!]
	\centering
	\caption{Overview of AIC sensitivity across task-specific domains in automated driving.}
	\resizebox{\linewidth}{!}{
    \begin{tabular}{p{1.2cm}p{1cm}p{4.5cm}}
			\toprule
			\textbf{Task} & \textbf{Method} & \textbf{Description} \\
			\midrule
            Scene Understanding & Motion Prediction &
            An appropriate context understanding requires the statistical input-output relationship of an motion prediction AI, e.g., \cite{{narayanan2021divide}}, to be road structure and topology sensitive $\lambda_{\mathrm{struct.}}$. 
            % Highly sensitive
            % \begin{itemize}
            %     \item Traffic flow patterns (e.g., dense vs. free-flowing traffic
            %     \item Driving cultures (e.g., aggressive vs. defensive driving styles)
            %     \item Road structure and topology (e.g., intersections vs. highways)
            %     \item Agent type and kinematics (e.g., pedestrians vs. cars vs. trucks) 
            % \end{itemize}
            \\ \midrule
            Planning & Trajectory-planning  & 
            In order to ensure an appropriate and adaptive system behavior, the statistical input-output relationship of a trajectory planning AI, e.g., \cite{chen2020learning}, should be scenario-sensitive. For instance, predefined scenario variations $\lambda_{\mathrm{scen.}}$, e.g. in pedestrian position/orientation, should result in respective change in the underlying planning mechanism.
            % Urban vs. highway sould also not follow the same trajectory generation principle
            %For instance, sensitivity influencing factors are the traffic density, the 
            %Planning should be sensitive to traffic density changes, but also occlusions    
            \\ \midrule
            Control & Ego dynamics& 
            In order to bound model errors and real world miss-assumptions, the statistical input-output relationship of ego dynamics, e.g., \cite{ullrich2023cnp}, should be sensitive to significant internal changes, such as vehicle mass and center of gravity shifts $\lambda_{\mathrm{mass}}$, as well as external changes like environment-induced friction variations $\lambda_{\mathrm{\mu}}$.
            \\
			\bottomrule
	\end{tabular}}
	\label{tab:prop_example_Sensi}
\end{table}

\subsection{AI Dynamics Stability}

While static and non-stationary AI systems have no internal memory and can, therefore, be investigated under static conditions, dynamic systems have an internal memory, which is why temporal behavior needs to be investigated as well. For this purpose, we introduce AI dynamics stability.

\textbf{Definition 3.3 (AI Dynamics (AID) Stability)} \textit{Denote a dynamic AI system as a system with an internal memory state $h_{t}$ and a dynamic statistical input-output relationship $P(Y|X)$, such that ${P(Y|X)}_{t+1} = f_{\mathrm{dyn}}({P(Y|X)}_{t}, h_{t})$ hold. The dynamic AI system is said to be AI dynamics stable if for any change in any AI circumstances $\zeta$ that impact the statistical input-output relationship, the following hold:}
\begin{equation}
	\begin{aligned}
		\forall t \in \mathbb{R} \setminus \{ t_{i} | \zeta \text{ occurs at }  t_{i} \} : D_{\text{KL}}[t+1] - D_{\text{KL}}[t] \leq 0.
	\end{aligned}
\end{equation}		
\textit{with $D_{\text{KL}}[t] = D_{\text{KL}}(P(Y|X)_t || P(Y|X)_{t-1})$.}
\newline $\Rightarrow$ \textit{Thus, it can be concluded for $\lim_{{t \to \infty}}$ that the dynamic AI system is stable, which is referred to as AI dynamics stability.}

\begin{figure}[h!]
	\centering	
	\includegraphics[scale=0.5]{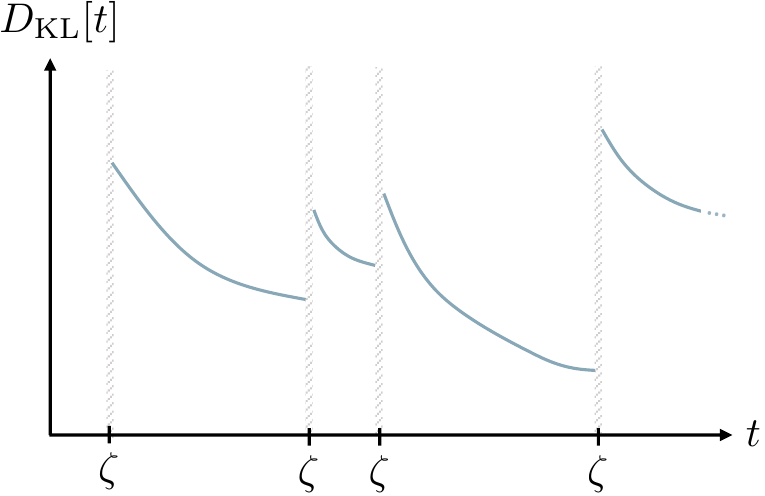}
	\caption{Visual illustration of AID stability.}
	\label{fig:AID_stabi}
\end{figure}

This notion of stability definition is inspired by the discrete-time Lyapunov stability \cite{shevitz1994lyapunov, sastry1999lyapunov} condition 
\begin{equation}
	V(\boldsymbol{x}[t+1]) - V(\boldsymbol{x}[t]) \leq 0
\end{equation}
according to \cite{drgovna2022dissipative}, which is defined by the Lyapunov function $V$ and the state $\boldsymbol{x}$. Here $V(\boldsymbol{x}) :  \mathbb{R}^{n_{x}} \rightarrow \mathbb{R}$, where $V(\textbf{0})=0$ and $\forall \boldsymbol{x}[t], t \in \mathbb{Z} \setminus{\{0\}}: V(\boldsymbol{x}[t]) \geq 0 $. The statistical input-output relationship $P(Y|X)$ can be regarded as a representation of the overall state of a dynamic AI system. Consequently, the KL-divergence \cite{kullback1951information} takes the entire state into account, which is necessary for an admissible Lyapunov function. Furthermore, the value range of the KL-divergence is $[0, \infty )$. Thus, the KL divergence can be treated as an admissible Lyapunov function. Beyond this, the definition of AID stability allows the criterion to be applied in operation, where circumstances $\zeta$ can occur. In doing so, $\zeta \supset \Lambda$ also takes into account sensitivity aspects that are not required but are present in the real world. A visual illustration of AID stability over a deployment horizon is depicted in Figure \ref{fig:AID_stabi}. Moreover, Table \ref{tab:prop_example_Stability} presents an application-oriented overview across various automated driving tasks.

\begin{table}[h!]
	\centering
	\caption{Overview of AID stability across task-specific domains in automated driving.}
	\resizebox{\linewidth}{!}{
    \begin{tabular}{p{1.2cm}p{1cm}p{4.5cm}}
			\toprule
			\textbf{Task} & \textbf{Method} & \textbf{Description} \\
            Perception & Object Tracking & Dynamic multi-object tracking (MOT) AI systems, e.g., \cite{milan2017online}, aggregate target motion dynamics $h_t$ for improved accuracy. Here, stability of the statistical input-output relationship should be ensured in the presence of object track births $\zeta_{\mathrm{birth}}$ and deaths $\zeta_{\mathrm{death}}$ of tracks. \\
			\midrule
            Scene Understanding & Motion Prediction &
            A dynamic motion prediction AI system, e.g., \cite{song2024motion}, aggregates contextual knowledge within $h_t$ to handle occlusions, maintaining stability when the change in the AI's statistical input-output relationship $D_{\text{KL}}[t]$ diminishes following a visibility disruption $\zeta_{\mathrm{occl.}}$.
            %\textbf{an sich kein wirklich richtige dynamische AI! Deshalb etwas schwierig}
            \\ \midrule
            Planning & Trajectory-planning  & 
            A dynamic planning AI system, such as \cite{renz2022plant}, aggregates tokenized scene representations into $h_t$ to model spatiotemporal interactions. Stability analysis of the statistical input-output relationship allows for the characterization of the dynamic behavior, thus enabling improved risk assessments. 
            %\textbf{an sich dynamisch AI, aber transformer begriff könnt für verwirrung sorgen}
            \\
			\bottomrule
	\end{tabular}}
	\label{tab:prop_example_Stability}
\end{table}

In addition to the AID stability as well as the AIC robustness and sensitivity properties, other properties such as generalization could also be specified in more detail in the future, e.g. with regard to the intended operational design domain on $P(Y|X)$. Although the transferability, reusability, and scalability of AI systems are important from an economic point of view and generalization is therefore important, this paper is initially limited to the definition of the three central properties: AIC robustness, AIC sensitivity, and AID stability. These properties can be used for targeted development, validation, approval, monitoring, and maintenance, as mentioned in the introduction to this section. However, in addition to properties and corresponding criteria, this requires general AI system enhancements. This is described in more detail in the following section along with initial conceptualizations according to the interdisciplinary perspective.

\section{AI SYSTEM ENHANCEMENTS TOWARDS INCREASED SAFETY}\label{sec:08_AI_Enhancements}

Data-based AI systems are subject to data-based iterative development, verification, and validation processes. Accordingly, online monitoring and evaluation of the system is of central importance and is increasingly required \cite{eu_parliament_2024corr}. While the interdisciplinary foundations and the definition of system classes and properties form the basis for a more formalized specification, development, and analysis process, this section provides insight into possible generic extensions of any AI system to improve AI safety during deployment and iteratively across the lifecycle. The corresponding perspective concepts are divided into enhanced assumption validation, enabled online system analysis, and responsible AI systems. Before these are explained in more detail below, the most recent regulatory requirements are discussed in compact form.

\subsection{Regulatory Requirements}

The regulatory landscape in AI is relatively recent, with only limited far-reaching regulations. The existing approaches in regulation are currently very heterogeneous and mostly more guidelines than proper regulations  \cite{metiaigovernanceguidelines, dsiaiframework, billc27, aigovernanceprinciples, ai2023artificial}. In contrast, the recently enacted EU AI Act \cite{eu_parliament_2024corr} represents the first far-reaching regulation that takes a risk-based approach. In particular, high-risk AI systems, e.g., cyber-physical systems that interact in the real world, are subjected to more comprehensive requirements. Some selected passages from \cite{eu_parliament_2024corr} are quoted directly below to give a closer insight. 

\begin{quote} \textit{"The risk-management system should consist of a continuous, iterative process that is planned and run throughout the entire lifecycle of a high-risk AI system. That process should be aimed at identifying and mitigating the relevant risks of AI systems ... This process should ensure that the provider identifies risks or adverse impacts and implements mitigation measures for the known and reasonably foreseeable risks of AI systems ... including the possible risks arising from the interaction between the AI system and the environment within which it operates."}\cite{eu_parliament_2024corr}  \end{quote}  % 65

This illustrates the need for a continuous risk management system and, at the same time, highlights the particular challenge posed by the environment. Furthermore, the requirement to take measures against known and foreseeable risks becomes apparent. This is specified in more detail as follows.

\begin{quote} \textit{" ... ensure that ... [AI systems] are used as intended and that their impacts are addressed over the system’s lifecycle. ...  It is also essential, as appropriate, to ensure that high-risk AI systems include mechanisms to guide and inform a natural person to whom human oversight has been assigned to make informed decisions if, when and how to intervene in order to avoid negative consequences or risks, or stop the system if it does not perform as intended."}\cite{eu_parliament_2024corr}  \end{quote} % 73

On the one hand, this excerpt emphasizes the intended purpose, which goes hand in hand with the requirement that the assumptions made for development are valid during execution. On the other hand, it illustrates that further mechanisms and system outputs should exist in addition to the actual AI system and the corresponding primary AI output in order to increase safety. While human oversight is particularly highlighted here, the following requirement is more generic and focuses on a technical solution. 

\begin{quote} \textit{"Technical robustness is a key requirement for high-risk AI systems. They should be resilient in relation to harmful or otherwise undesirable behaviour that may result from limitations within the systems or the environment in which the systems operate (e.g. errors, faults, inconsistencies, unexpected situations). Therefore, technical and organisational measures should be taken to ensure robustness of high-risk AI systems, for example by designing and developing appropriate technical solutions to prevent or	minimize harmful or otherwise undesirable behaviour. Those technical solution may include for instance mechanisms enabling the system to safely interrupt its operation (fail-safe plans) in the presence of certain anomalies or when operation takes place outside certain predetermined boundaries."} \cite{eu_parliament_2024corr} \end{quote} % 75

This quote argues for technical measures to address risks and, in particular, recognizes the correlation between the AI system and the environment in terms of risks and safety. In addition, it highlights the need to continuously review the assumptions during operation. By defining the properties of an AI system as in Section \ref{sec:07_AI_System_Properties}, taking into account the assumptions, circumstances of the respective environment, and the system behavior specification, the system properties open up promising possibilities about the required measures. In particular, the requirements for technical measures, extended information provision, and targeted intervention can be built on this. The resulting classification and conceptualization of the following enhanced assumption validation enabled online system analysis and responsible AI systems to combine existing concepts with new ideas while taking regulatory
requirements into account along the new perspective.

\subsection{Enhanced Assumption Validation}
AI systems are integrated subsystems exposed to an input that, unlike control technology, usually cannot be manipulated. For the safety of AI systems, checking for intentional use is, therefore, essential and elementary. For instance, it is necessary to consider the input specification \cite{burton2020mind}, \cite{burton2023closing}, and assumptions of the AI system. With data-based AI systems, the assumptions are encoded into the datasets used. Accordingly, the training, validation, and test data represent a basis for checking the assumptions during operation. 

A commonly used assumption validation technique is represented by approaches such as out-of-distribution (OOD) detection \cite{hendrycks2016baseline, liu2020energy}. These could consider various aspects based on AI input data as shown in Table \ref{tab:example_ODD}. 

\begin{table}[h]
	\centering
	\caption{OOD classification examples.}
	\begin{tabular}{lp{0.5\linewidth}}
		\toprule
		\textbf{OOD class} & \textbf{Description} \\
		\midrule
		Data characteristics & feature values, data format or source \\
		Noise \& errors & high noise, anomaly, unknown pattern \\
		Outlier & datapoints deviating from norm \\
		Rare events & unusual events that have not occurred \\
		Concept shift & change in data distribution over time \\
		Domain difference & unadapted model or changed domain \\
		Adversarial attacks & specific crafted model deceiving input \\
		Unknown class & unclassified OOD buffer \\
		\bottomrule
	\end{tabular}
	\label{tab:example_ODD}
\end{table}

Moreover, besides generic OOD detectors, input-centric safety monitors \cite{liu2020input}, \cite{sabokrou2018adversarially} are also commonly used. In both cases, mismatches between input and development data \cite{gangal2020likelihood}, \cite{xu2020deep} are usually detected. In this way, the AI accessing components could be informed that the AI cannot be trusted since it is used outside specification. An illustrative example of such an enhancement in the context of AD is provided by \cite{hacker2023insufficiency}. Generally, superordinate software management takes up the information about AI misuse and switches to redundant software components \cite{shukla2020flight}. However, the downstream task does not benefit any switch if the redundant software considers equally degraded data sources. In contrast, the developer of an AI system or even the AI system itself could imagine foreseeable risks and incorporate self-awareness to provide externals with task-specific support even in the event of assumption violations.

\begin{figure*}
	\centering	
	\includegraphics[scale=0.5]{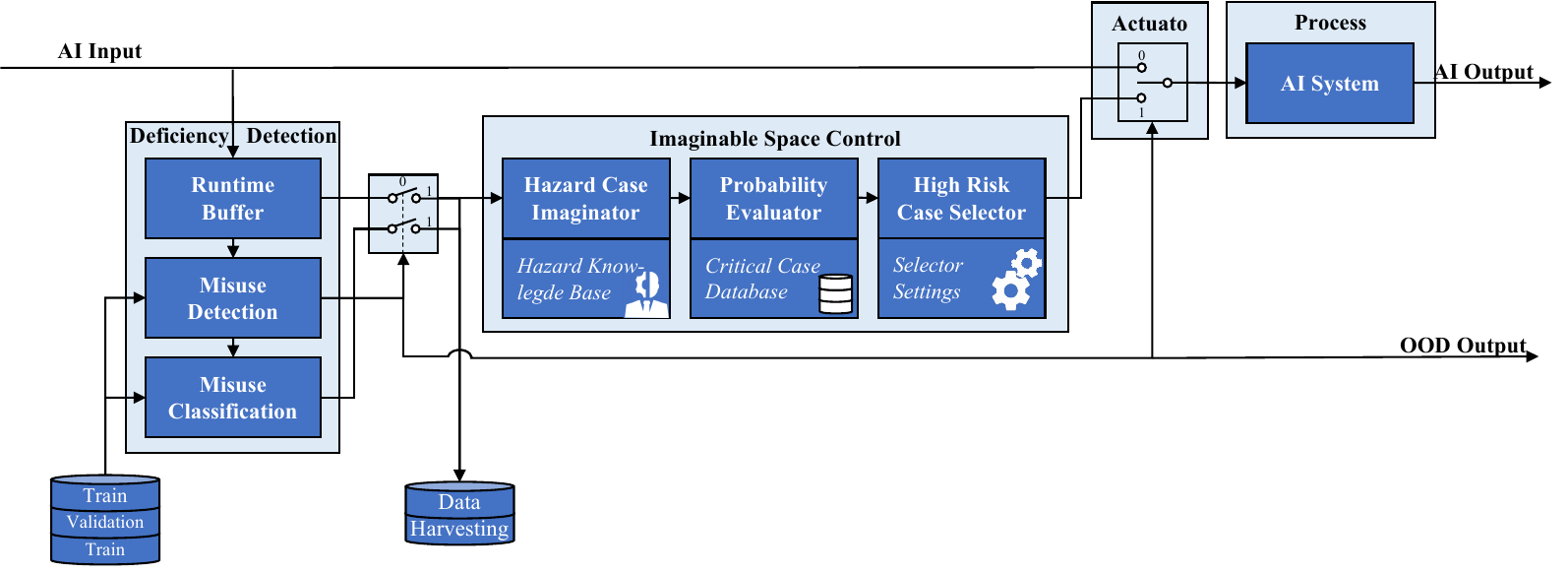}
	\caption{Conceptualization of an extended input inspection including technical measures as risk management for foreseeable hazards in an imaginable space.}
	\label{fig:fist_part3}
\end{figure*}

Accordingly, a methodology that goes beyond classical AI misuse detection by providing additional information for online fault prevention is desirable, and it also serves as a basis for continuous maintenance to improve AI safety in the long term. Therefore, the input data must be buffered to enable the general detection of deviations and shifts in a distribution-based manner. Inspired by Konrad Lorenz's vision of working in the imagined space \cite{lorenz1973ruckseite}, a general AI enhancement methodology could be formalized that incorporates additional knowledge and functionalities to provide a first stage of self-awareness. This is depicted schematically in Figure \ref{fig:fist_part3}. 

The approach illustrated in Figure \ref{fig:fist_part3} partially uses hard-coded reasoning and system and domain knowledge to achieve dedicated self-awareness and -governance to mitigate challenges like source degradation in a general manner. In essence, the concept resembles a human-inspired approach that has been dramatically simplified. This is briefly explained in more detail using a trajectory planning example in the context of automated driving. 

In the case of trajectory planning, it is not helpful if a perceptual AI indicates being used outside of specification. If, in contrast, a person loses perception for a short time, he will usually retain the last impressions, imagine conceivable developments in the current situation, and adapt his behavior accordingly, e.g., initiate an (emergency) posture. Here, the imaginary ability enables one to reason about evolution and its consequences. Therefore, this ability is essential for AI risk mitigation. Without this ability to imagine, danger-avoiding approaches usually assume worst-case scenarios, thus always conducting emergency stops. However, this could lead, especially in cases of internal degradation, to unpredictable behavior for externals such as other road users and thus poses potential hazards in the overall context \cite{rottmann2020detection}. Therefore, it is essential to classify the entered misuse, e.g., as zero or one, and provide further information to ensure the best possible safety. 

This is addressed in a simplified, knowledge-induced way in the concept shown in Figure \ref{fig:fist_part3}. The concept is based on established misuse detection and classification, considers data buffering of critical cases for continuous improvement of AI systems over the lifecycle, and considers a simplified, knowledge-induced imagination. The imagination comprises three elements: the hazard case imaginator, the probability evaluator, and a high-risk case selector. The hazard case imagination is driven by a data-based knowledge base that can depict foreseeable risks in conjunction with the runtime buffer. The generated imaginations can subsequently be assigned a probability of occurrence by the probability evaluator, using knowledge in the form of a database as well. Finally, the high-risk case selector represents a setting unit responsible for selecting and processing the generated and weighted scenarios. After all, these can be passed as input to the AI. Thus, in the event of a positive OOD detection, i.e., an impermissible input, the input signal is replaced by a foreseeable risk input signal fed to the AI system. Thus, the functionality of the AI system is used in one or more permissible foreseeable risk scenarios, whereby further information is provided for the safe use of the system. This is briefly illustrated below using the trajectory planning example once again.  

In the case of perception, for example, the worst-case scenario would be to suspect objects everywhere when perception is lost, which is unrealistic if there are no road users in the current context. Instead, by using expert knowledge, potentially risky situations can be anticipated. Experts may prioritize objects in front of a vehicle over those behind, considering various dangerous scenarios. For instance, in a highway scenario stored in the runtime buffer until the loss of perception (positive OOD detection), hazardous situations can be projected into the future based on the knowledge at hand. Accordingly, imagining oncoming traffic or junction scenarios at this stage is unnecessary. Instead, concentrating on new road users that appear in the vicinity, as usual in overtaking or merging maneuvers, is foreseeable. The hazardous situations generated can be compared with the training, validation, and test databases to assess the probability of their occurrence. The risk can be determined by multiplying the probability by the extent of the impact, and a ranking of the hazardous situations can be created. Depending on the setting, the AI can process one or more generated scenarios and thus create imaginary high-risk situations in addition to positive misuse detection. For instance, trajectory planning would now have input data upon which a more realistic emergency-stopping trajectory can be planned. If the situation is inherently safe, a safe stop that is predictable to others can be initiated appropriately. The methodology would account for a realistic hazard case consideration and thus improve safety. In addition, data harvesting allows for iterative revision of requirements and adaptation of the AI specification so that the system can be continuously improved and made more resilient. This human-like approach makes it possible to imagine dangerous situations and empirically estimate their probability, allowing dangerous but realistic situations to be considered in downstream tasks.

Accordingly, AI systems are integrated subsystems exposed to inputs, but with appropriate risk management, it is possible to incorporate system knowledge and provide further helpful information in addition to inadmissibility. This concept could be figuratively associated with a feed-forward controller in control engineering. Moreover, compared to Yann LeCun's visionary architecture \cite{lecun2022path}, the concept in Figure \ref{fig:fist_part3} provides an approach that utilizes partially hard-coded reasoning as well as system and domain knowledge to achieve dedicated self-governance to mitigate the challenges like source degradation of general AI systems. Thus, this represents a generic concept that could enhance any AI system.

\subsection{Enabled Online System Analysis}

While the previously outlined enhanced assumption validation primarily addresses the misuse, specification gaps, and associated uncertainties and risks \cite{burton2020mind, burton2023closing, burton2023addressing}, the concept of enabled online system analysis addresses the monitoring of functionality.
Thereby, the enabled online system analysis considers the input-output behavior through the assessment of AIC robustness, AIC sensitivity and AID stability properties introduced in Section \ref{sec:07_AI_System_Properties}. On this basis, the concept at hand serves to ensure safety and reliability.

\begin{figure}[b]
	\centering	
	\includegraphics[scale=0.54]{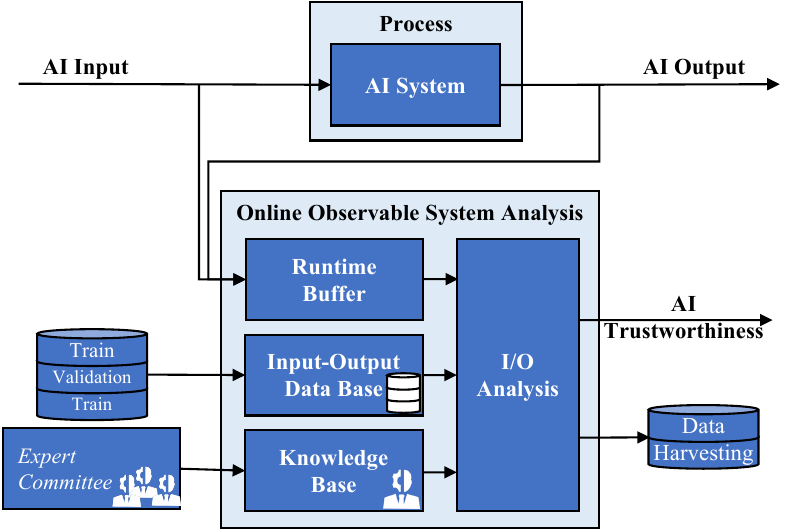}
	\caption{Concept of an online system analysis of observable AI inputs and outputs, whereby defined properties of robustness, sensitivity, and stability can be employed within the I/O analysis.}
	\label{fig:observer}
\end{figure}

Thereby, the statistical input-output relationship can be monitored online during operation based on dataset distributions, such as development-time and operation datasets. Similar to enhanced assumption validation, buffering is required to account for changes in the distributions regarding inputs and outputs. In addition, expert knowledge can also be considered, e.g., system-related limits. The corresponding concept is visualized in Figure \ref{fig:observer}.

Accordingly, the concept depicted in Figure \ref{fig:observer} offers the possibility of an online analysis of AIC robustness, sensitivity, and AID stability. Thus, the basic concept of input-output-driven safety monitors \cite{ferreira2021benchmarking} is taken up in a more formalized and theoretically founded manner. Through this, a better understanding and trustworthiness in the components can be developed. In addition, identifying trigger conditions for the AI system is possible. Trigger conditions according to SOTIF\footnote{\url{https://www.iso.org/standard/77490.html}} are of particular importance for functional safety \cite{lakkaraju2017identifying}. In this way, the safety of AI systems can also be improved in a human-like manner in the long term \cite{gopnik2004theory}. The concept of online system analysis thus offers a variety of options for monitoring and improving safety and can be used complementarily and independently of the enhanced assumption validation. In particular, data-based verification represents the surveillance of specified, verified, and validated functionality and is a fundamental basis for AI safety assurance. 

To build a bridge to the perspective of control engineering, the present concept can be related figuratively to the concept of observers in control engineering. Observers of control engineering offer the possibility to reconstruct or estimate system states based on input and output variables. This is especially important for states that cannot be measured. AI systems have tappable inputs and outputs and various unknown internal system states, as they are generally regarded as black box systems. While disentangled feature representations \cite{gowal2020achieving}, \cite{liu2023causal}, \cite{rombach2023controlled} counteract this, it is less concerned with safety. In addition, disentangled feature representation learning imposes requirements that cannot be accommodated for all AI systems and architectures. For instance, AI systems generally cannot be assumed to have their internal states tappable and interpretable. Therefore, an input-output-based concept is widely applicable. Accordingly, properties that depend on states can also be considered in addition to states. For example, safety and reliability, which in turn can be likewise interpreted as non-measurable states. These could be reconstructed in an observational manner from the inputs and outputs. This is taken up by the concept shown in Figure \ref{fig:observer}, which has similarities with control engineering concepts on an abstract level. 

\subsection{Responsible Self-aware AI Systems}

While the previous concepts can, in principle, be used as AI system augmentations to address the inherent challenges in safeguarding AI systems, the long-term goal for AI systems \cite{lorenz1973ruckseite, lecun2022path, kahneman2011thinking} is a less hard-coded human-like ability to reason in an imaginable way to achieve intrinsically safe outputs. However, these are concepts that are still far from realization. However, a closer look at the concepts reveals that approaches such as Kahneman's “System 1” and “System 2” \cite{kahneman2011thinking} and LeCun's “Mode-1” and “Mode-2” \cite{lecun2022path} take up concepts of control engineering, namely Model Predictive Control (MPC) \cite{rawlings2000tutorial, englert2019software}. This opens up the prospect of using control engineering concepts in a more direct form as the realization appears more predictable. In particular, advanced control engineering techniques such as MPC and Moving Horizon Estimation (MHE) \cite{allan2019moving} appear promising. 

MPC uses a dynamics system model to optimize control variables over a horizon into the future according to a cost function. In comparison, moving horizon estimation is backward-looking. More specifically, MHE \cite{allan2019moving} is an approach for state and parameter identification in dynamic systems. The basic principle is to estimate these quantities in a moving time window. While MPC is considered in far-reaching AI concepts, the methodology is also used more directly to increase safety, namely through predictive safety filters. In contrast, MHE has not yet been considered for enhancing AI safety. To underpin the benefits of control concepts, MHE is being selectively reconsidered. 

As with the observer, such as online safety monitoring, the validity and reliability of an AI component can be evaluated online via an MHE. The basis for this is information about the actual correct output of the AI system, often available in AI applications. For instance, as in control technology, measurements can be taken that provide the actual value of a previously predicted AI system output. Furthermore, other sources, such as human inputs or other external sources, are also conceivable in the field of AI systems. Based on the output of the system in the past as well as the knowledge about the correct actual production, the confidence of the AI output can be provided alongside by an safety increasing MHE. Subsequent components could take it into account and benefit from it. 

\begin{figure}
	\centering	
	\includegraphics[scale=0.54]{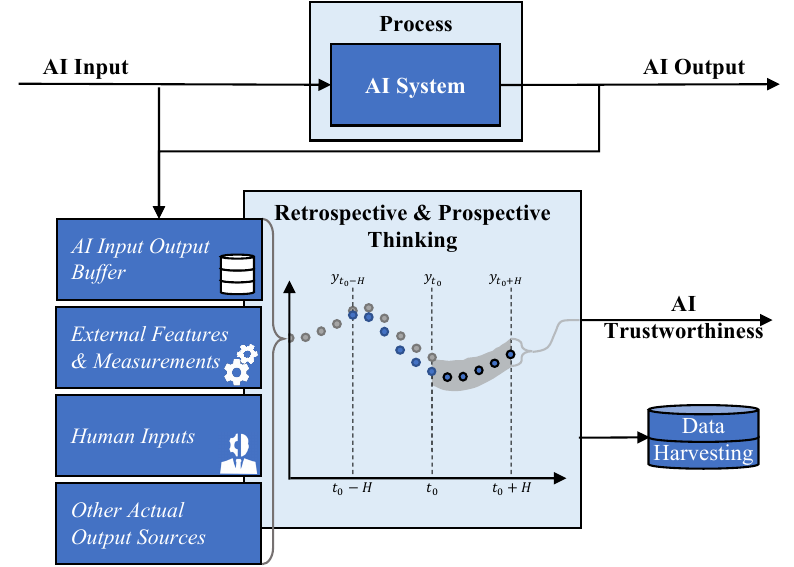}
	\caption{Visualization of a concept for the integration of retrospective and prospective thinking by figuratively employing advanced methods of control engineering.}
	\label{fig:MHE}
\end{figure}

If the AI system is a predictive AI, this MHE principle can be combined with model predictive control (MPC), similar to control engineering. The combination is shown in Figure \ref{fig:MHE} in the AI data-centric logic.

The applicability and relevance of this approach can be illustrated by trajectory planning in autonomous driving. One task that is nowadays mainly performed by AI due to the inherent complexity is the prediction of other road users  \cite{chang2019argoverse}, \cite{zhan2019interaction}, \cite{houston2021one}, \cite{caesar2021nuplan}. Even if a perfect AI model were present, the system would inherit assumptions that were assumed to be permissible. Still, it could be disregarded in the open world, e.g., by human participants who violate given rules unpredictably. In such cases, it is apparent that the AI prediction is wrong. However, it is not relevant for safety how the discrepancy between prediction and reality arises, but instead that there is a discrepancy. At this time, planning must be informed that the projections cannot be trusted, even though the AI component has been trained correctly and is working properly. While such corner cases could be considered after training in an iterative AI refinement, the MHE-based concept provides confidence during operation in the real world. For instance, if there is a high discrepancy, the prediction of others' behavior can be less trusted, and planning must provide more conservative and cautious trajectories. It does not matter who was wrong, the prediction, or the other participant. The critical aspect is to ensure safety. This can be achieved by recognizing and considering the deviation in planning via the data-based MHE and MPC approach.

While MPC figuratively represents an approach to prospective thinking, MHE presents an approach to retrospective thinking. Thus, both provide central components for responsible self-awareness of AI systems. Thus, it turns out that advanced control engineering methods could also provide advanced solutions for AI system enhancements toward increased safety. 
Initial efforts indicating such a methodological bridging, outlined in Table \ref{tab:AI_enhance}, underscore the need for further investigation in this direction.

% ToDo
% \begin{itemize}
%     \item Tabel with examples for: 1) prediction(prospective), 2) retrospective 
%     \item illustrate shift also towards latent 1) and 2)
% \end{itemize}

% Examples:
% \begin{itemize}
%     \item LeCuns MPC stuff, check citations, maybe some realizations
%     \item checkn self-supervised AI systems \cite{assran2023self}, in automated driving
%     \item latent predictions (being more efficient than real world redictions)
%     \item from \cite{guan2024world} "BEV labeling, underscores the practical necessity of innovtive solutions like world models [29], [31], [32]. By generating predictive scenarios from historical data, these models no"
%     \item JEPA: motion \cite{fei2023jepa}, listen \cite{fei2023jepa}
% \end{itemize}

% \begin{itemize}
%     \item retrospective:
%     \begin{itemize}
%         \item \cite{song2024motion}, retrospective, temporally interrelation consideration
%         \item TrustMHE 
%     \end{itemize}

%     \begin{itemize}
%         \item RL via V-value and Q-learning having expected etc... (but this is methodology based not really general AI that is forward-looking and evaluating expected return) Is there a paper in the prospective manner out there?
%     \end{itemize}
% \end{itemize}

\begin{table}[h!]
	\centering
	\caption{Overview of recent approaches in AD within the context of enhancing responsible, self-aware AI systems.}
	\resizebox{\linewidth}{!}{
    \begin{tabular}{p{1.2cm}p{1cm}p{4.5cm}}
			\toprule
			\textbf{Task} & \textbf{Method} & \textbf{Description} \\
            retrospective & TrustMHE \cite{ullrich2025iv}& Enhanced self-awareness in trajectory prediction systems through retrospective reliability estimation of predicted trajectories. \\ \midrule
            prospective & MC-JEPA \cite{bardes2023mc} & Joint-Embedding Predictive Architecture (JEPA) for self-supervised learning of motion and content features. \\ \midrule
            retro- \& prospective &  BridgeAD \cite{zhang2025bridging}& Ensures consistent retro- and prospective use by incorporating historical predictions into perception and integrating both historical prediction and planning for future frames into the motion planning module. \\
			\bottomrule
	\end{tabular}}
	\label{tab:AI_enhance}
\end{table}

% \begin{table}[h!]
% 	\centering
% 	\caption{ToDo.}
% 	\resizebox{\linewidth}{!}{
%     \begin{tabular}{p{1.2cm}p{1cm}p{4.5cm}}
% 			\toprule
% 			\textbf{Task} & \textbf{Method} & \textbf{Description} \\
% 			\midrule
%             retrospective  & (RealMotion)\cite{song2024motion}& . \\
%             \cmidrule{2-3}
%               & TrustMHE \cite{ullrich2025iv}& Retrospective reliability estimation of trajectory predictions. \\ \midrule
%             prospective & Dreamer \cite{hafner2019dream} &  . \\ \cmidrule{2-3}
%              & MC-JEPA \cite{bardes2023mc} & Joint-Embedding Predictive Architecture (JEPA) for self-supervised learning of motion and content features. \\ \midrule
%             retro- \& prospective &  BridgeAD \cite{zhang2025bridging}& . \\ \cmidrule{2-3}
%                 &  \cite{}& . \\ 
% 			\bottomrule
% 	\end{tabular}}
% 	\label{tab:AI_enhance}
% \end{table}

In general, there are now three high-level concepts that could increase the safety of AI systems. The first concept addresses the extension of existing input specification analysis and measures to address foreseeable risks. The second concept is dedicated to the online analysis of AI functionality, whereby the system class and system properties can be employed. Concept three illustrates how desired properties, more precisely, retrospective and prospective thinking, can create responsible, self-aware AI systems to increase safety. All concepts can be interpreted in terms of execution analysis, system monitoring, and technical measures to increase safety. In addition, all concepts recognize the need for iterative refinement of AI systems and consider the detection of specification gaps, trigger conditions and unknown unknowns \cite{burton2020mind, burton2023closing, burton2023addressing}. Furthermore, the present concepts integrate, similar to the concept of a quality vector of \cite{ecksteinautotech}, the expansion of payload data by quality and reliability data to offer downstream modules and higher-level monitoring and decision-making units the knowledge acquired. Overall, the concepts are figuratively related to control engineering concepts, which illustrates that using existing knowledge in an interdisciplinary manner could benefit AI safety in general.

% \textbf{ToDo: Optional}
% Approaches for Vertification and switching, Part 1
% \begin{itemize}
%     \item Paper HTML outcommented: %https://ieeexplore.ieee.org/stamp/stamp.jsp?tp=&arnumber=10107652
% \end{itemize}

% \cite{yang2023towards}

% Approaches for other class like sensitivity and robustness analysis
% \begin{itemize}
%     \item ?Cert-RNN Robust
% %https://nesa.zju.edu.cn/download/dty_pdf_cert_rnn.pdf
% \end{itemize}

\section{Discussion}\label{sec:09_Discussion} % and State of the Art

%- with existing higher-level systems and interaction with the environment

%First, a far-reaching reinterpretation and definition of key aspects such as robustness and stability is required. In the long run, these can be concretized in the individual domains. At this point, however, general understanding and bridging between the disciplines is first necessary. Methods of control engineering can and will be transferred in this context using selected examples.

In the field of data-based signals and AI systems, data is of central importance. Alterations to data can have significant consequences. The data usually originates from a process in which AI systems are integrated and can, therefore, only be controlled to a limited extent or even not at all. Moreover, assumptions are also made implicitly in the form of datasets during system development. Data-based implicit modeling poses dangers such as specification gaps, improper uses of AI systems, and remaining blind spots \cite{burton2020mind}, \cite{burton2023closing}, \cite{burton2023addressing}. 

Furthermore, in the real world, especially concerning cyber-physical systems, it can be postulated over the entire AI lifecycle that changes to the data can be expected. Consequently, it is of utmost importance for the safety of AI systems to analyze the behavior of the systems and their effects during development as well as observe the behavior across the AI lifecycle. Thus, safety relies on fault identification \& prevention \cite{rojas2018invariant, magliacane2018domain, subbaswamy2019preventing}, monitoring \cite{scholkopf2001estimating, hendrycks2016baseline, bendale2016towards} and continuous maintenance \cite{heckerman1995decision, lakkaraju2017identifying, nushi2018towards, hendrycks2019benchmarking}. Moreover, in terms of validation and approval, a generic AI architecture and application-agnostic approach is particularly desirable, which also offers the possibility of enabling long-term validity in the rapid development of AI systems. This motivates and justifies the discussion of a new paradigm called data control.

We are thinking of data control as a paradigm that focuses on AI systems, as data-based signals and AI systems, that can be represented by a model $\mathcal{M}$ formed via a data-based learning methodology, including at least some proportion of statistical learned components. Furthermore, the AI system could but is not required to include various levels of incorporated structural knowledge that could be gathered either data-based via statistical or causal learning or directly by expert and domain knowledge. Furthermore, in the context of the intended paradigm of data control, it could be assumed that the model perceives inputs $X$ and provides outputs $Y$ in the form of data, which are required to be observable. The input-output behavior is assumed to be described via the statistical input-output conditional probability $P(Y|X)$ that accounts via the probabilistic distributional manner, the characteristics of data-based signals and systems. Beyond that, the new paradigm of data control could assume that AI systems can be categorized at least as static, non-stationary, or dynamic AI systems. Moreover, even if such a paradigm does not require specific insights about the internal architecture and the general application, a detailed system specification describing the desired and intended functionality and the corresponding specification of the working environment $\mathcal{E}$ seems mandatory. For this setup, the paradigm of data control would be intended to provide a safety approach based on generic definitions of properties that can be used from synthesis to analysis, from development to deployment, and from evaluation across monitoring to diagnosis.

Despite the broad applicability of the intended paradigm, the establishment of data control is mainly motivated by the need to formalize and concretize system analysis, especially regarding system approval and safe deployment. In this way, safety monitoring can be improved without being limited to monitoring. The specifications demanded in this context would force developers to explicitly state the underlying assumptions. For instance, this enables improved release procedures for critical applications and targeted monitoring, diagnostics, release revocation, re-release, and much more. The methodology of the intended paradigm is generally applicable but targets explicitly the challenges during execution time, such as monitoring properties during deployment to increase safety.

The necessity of a new paradigm can be discussed and questioned. On the one hand, however, the development of AI systems is primarily performance-oriented, with system properties playing a subordinate role. On the other hand, efforts to increase the safety of AI systems are primarily very specific and not widely applicable. Instead, a stringent, generic, agnostic approach is desirable and, therefore, both wide-ranging and long-term relevant. To achieve this, a new perspective was adopted that combines approaches and ideas interdisciplinarily to improve the safety of AI systems. As part of this, the principle of system analysis of control engineering using system classes and properties is exemplarily transferred to AI systems through an interdisciplinary interpretation of the underlying process while considering the data engineering perspective. In doing so, some limitations of existing methods could be eliminated. This includes, for example, the applicability of existing system properties.

Indeed, while existing efforts such as distributional robustness \cite{esfahani2015data, sinha2017certifying, duchi2021learning, rahimian2019distributionally} increase robustness from a performance perspective, initial approaches exist that formalize robustification \cite{meinshausen2018causality} and the corresponding evaluation \cite{subbaswamy2021evaluating, gupta2024s, chen2021mandoline}. In particular, Adarsh Subbaswamy made some contributions in this area \cite{subbaswamy2019preventing, subbaswamy2020development, subbaswamy2021evaluating, subbaswamy2022unifying}. This way, stability and a stability analysis \cite{subbaswamy2021evaluating} based on stable distribution definitions \cite{subbaswamy2020development, subbaswamy2022unifying} and different stability levels were introduced. While some assumptions could be eliminated in this context, e.g., unstable edges are also legitimate as long as they are non-active \cite{subbaswamy2018counterfactual}, the methods do not apply to high-dimensional inputs \cite{subbaswamy2023causal} as they occur in a variety of AI applications. Besides that, from the control engineering perspective, stability corresponds to dynamics \cite{lunze2010regelungstechnik, follinger2011laplace, fadali2012digital} instead of invariances.

Given increasing complexity and autonomy and thus increasing internal memory states, we consider it imperative to reserve the concept of stability for corresponding dynamic systems. In addition, we take into account the robustness and sensitivity in relation to the circumstances and specifications. This results in a methodology with an application- and approval-oriented perspective. For example, robustness against signal noise and robustness against external interference such as crosswinds in drone applications. Robustness against recalibrations of sensors or robustness against a variation in the installation angle of a camera. At the same time, the property of sensitivity to changes, e.g., in the detection and adaptation of the prediction of the driving dynamic depending on the usage, e.g., the payload or driving behavior, as well as dependence on external factors, e.g., changes in friction due to different road surfaces and weather conditions, could be considered. Thus, these concepts allow the prevailing performance analysis to be supplemented by analyzing the desired system behavior. In addition, in the context of AI system enhancements, possibilities are outlined, showing how such properties can be considered beyond the development stage in the deployment stage and over the lifecycle in general. In the first instance, only high-level concepts are discussed. As the focus of this paper is on new perspectives, the realization and evaluation of these concepts is part of future research.

Overall, different perspectives are taken, new perspectives are considered, possible developments are identified, and a new paradigm called data control is discussed to develop such an interdisciplinary and formalized perspective to improve AI system analysis and safety assurance in a far-reaching scalable and transferable manner.

\section{CONCLUSIONS}\label{sec:10_Concluison}

In addition to physical, chemical, and technical systems, data-based systems are becoming increasingly important. These data-based systems have their own characteristics, but also many similarities to traditional engineering. For instance, concepts of signals and systems or control theory. The new perspective outlined in this paper reinterprets the concepts of control engineering in terms of considering data-based signals and AI systems while focusing, in particular, on system analysis based on data distributions. 

While the paradigm data-driven control applies the methods of AI in the context of control engineering to improve control engineering, a new paradigm called data control could represent the inverse, namely drawing on the knowledge of control engineering to improve AI systems engineering in terms of system analysis and safety assurance. By bringing together the different perspectives, the new perspective offers the possibility to cumulate the respective achievements and, moreover, establish an interdisciplinary and goal-oriented blueprint paradigm in the long term. Indeed, due to the AI lifecycle, system analysis and safety assurance of AI systems is a matter of permanent review and monitoring, which is also suggested by the term data control. Thus, the new perspective justifies the establishment of the paradigm of data control that promises to empower AI safety through control theory methodologies for safety-critical real-world applications and beyond general applications in an interdisciplinary manner.

\bibliographystyle{IEEEtran}
\bibliography{literature}

\begin{thebibliography}{100}
\providecommand{\url}[1]{#1}
\csname url@samestyle\endcsname
\providecommand{\newblock}{\relax}
\providecommand{\bibinfo}[2]{#2}
\providecommand{\BIBentrySTDinterwordspacing}{\spaceskip=0pt\relax}
\providecommand{\BIBentryALTinterwordstretchfactor}{4}
\providecommand{\BIBentryALTinterwordspacing}{\spaceskip=\fontdimen2\font plus
\BIBentryALTinterwordstretchfactor\fontdimen3\font minus
  \fontdimen4\font\relax}
\providecommand{\BIBforeignlanguage}[2]{{%
\expandafter\ifx\csname l@#1\endcsname\relax
\typeout{** WARNING: IEEEtran.bst: No hyphenation pattern has been}%
\typeout{** loaded for the language `#1'. Using the pattern for}%
\typeout{** the default language instead.}%
\else
\language=\csname l@#1\endcsname
\fi
#2}}
\providecommand{\BIBdecl}{\relax}
\BIBdecl

\bibitem{gao2018object}
H.~Gao, B.~Cheng, J.~Wang \emph{et~al.}, ``{Object Classification Using
  CNN-Based Fusion of Vision and LIDAR in Autonomous Vehicle Environment},''
  \emph{IEEE Trans. Ind. Inform.}, vol.~14, no.~9, pp. 4224--4231, 2018.

\bibitem{letzgus2022toward}
S.~Letzgus, P.~Wagner, J.~Lederer \emph{et~al.}, ``{Toward Explainable
  Artificial Intelligence for Regression Models: A methodological
  perspective},'' \emph{IEEE Signal Process. Mag.}, vol.~39, no.~4, pp. 40--58,
  2022.

\bibitem{ezugwu2022comprehensive}
A.~E. Ezugwu, A.~M. Ikotun, O.~O. Oyelade \emph{et~al.}, ``{A comprehensive
  survey of clustering algorithms: State-of-the-art machine learning
  applications, taxonomy, challenges, and future research prospects},''
  \emph{Eng. Appl. Artif. Intell.}, vol. 110, pp. 1--43, 2022, {Art. no.
  104743}.

\bibitem{lim2021time}
B.~Lim and S.~Zohren, ``Time-series forecasting with deep learning: a survey,''
  \emph{Phil. Trans. R. Soc.}, vol. 379, no. 2194, pp. 1--14, 2021, {Art. no.
  20200209}.

\bibitem{zhang2019application}
X.~Zhang and W.~Dahu, ``Application of artificial intelligence algorithms in
  image processing,'' \emph{J. Vis. Commun. Image Represent.}, vol.~61, pp.
  42--49, 2019.

\bibitem{lauriola2022introduction}
I.~Lauriola, A.~Lavelli, and F.~Aiolli, ``{An introduction to Deep Learning in
  Natural Language Processing: Models, techniques, and tools},''
  \emph{Neurocomputing}, vol. 470, pp. 443--456, 2022.

\bibitem{bareinboim2016causal}
E.~Bareinboim and J.~Pearl, ``Causal inference and the data-fusion problem,''
  \emph{Proc. Natl. Acad. Sci.}, vol. 113, no.~27, pp. 7345--7352, 2016.

\bibitem{zhuang2020comprehensive}
F.~Zhuang, Z.~Qi, K.~Duan \emph{et~al.}, ``{A Comprehensive Survey on Transfer
  Learning},'' \emph{Proc. IEEE}, vol. 109, no.~1, pp. 43--76, 2020.

\bibitem{cunningham2008supervised}
P.~Cunningham, M.~Cord, and S.~J. Delany, ``{Supervised Learning},'' in
  \emph{{Machine Learning Techniques for Multimedia: Case Studies on
  Organization and Retrieval}}.\hskip 1em plus 0.5em minus 0.4em\relax
  Springer, 2008, pp. 21--49.

\bibitem{hastie2009unsupervised}
T.~Hastie, R.~Tibshirani, J.~Friedman \emph{et~al.}, ``{Unsupervised
  Learning},'' \emph{{The Elements of Statistical Learning: Data Mining,
  Inference, and Prediction}}, pp. 485--585, 2009.

\bibitem{sutton2018reinforcement}
R.~S. Sutton and A.~G. Barto, \emph{{Reinforcement Learning: An Introduction}},
  2nd~ed.\hskip 1em plus 0.5em minus 0.4em\relax Cambridge, MA, US: MIT Press,
  2018.

\bibitem{jebara2012machine}
T.~Jebara, \emph{{Machine Learning: Discriminative and Generative}}.\hskip 1em
  plus 0.5em minus 0.4em\relax Springer Science \& Business Media, 2012, vol.
  755.

\bibitem{wang2025generative}
Y.~Wang, S.~Xing, C.~Can \emph{et~al.}, ``Generative ai for autonomous driving:
  Frontiers and opportunities,'' \emph{arXiv preprint arXiv:2505.08854}, 2025.

\bibitem{jiang2022quo}
Y.~Jiang, X.~Li, H.~Luo \emph{et~al.}, ``Quo vadis artificial intelligence?''
  \emph{Discover Artificial Intelligence}, vol.~2, no.~4, pp. 1--19, 2022.

\bibitem{oppenheim1997signals}
A.~V. Oppenheim, A.~S. Willsky, S.~H. Nawab \emph{et~al.}, \emph{{Signals and
  Systems}}.\hskip 1em plus 0.5em minus 0.4em\relax Prentice hall Upper Saddle
  River, NJ, 1997, vol.~2.

\bibitem{girod2013einfuhrung}
B.~Girod, R.~Rabenstein, and A.~K. Stenger, \emph{{Einf{\"u}hrung in die
  Systemtheorie: Signale und Systeme in der Elektrotechnik und
  Informationstechnik}}.\hskip 1em plus 0.5em minus 0.4em\relax
  Springer-Verlag, 2013.

\bibitem{dally1998digital}
W.~J. Dally and J.~W. Poulton, \emph{{Digital Systems Engineering}}.\hskip 1em
  plus 0.5em minus 0.4em\relax Cambridge university press, 1998.

\bibitem{phillips2007digital}
C.~L. Phillips and H.~T. Nagle, \emph{{Digital Control System Analysis and
  Design}}.\hskip 1em plus 0.5em minus 0.4em\relax Prentice Hall Press, 2007.

\bibitem{krittanawong2018rise}
C.~Krittanawong, ``The rise of artificial intelligence and the uncertain future
  for physicians,'' \emph{Eur. J. Intern. Med.}, vol.~48, pp. e13--e14, 2018.

\bibitem{jha2019comprehensive}
K.~Jha, A.~Doshi, P.~Patel \emph{et~al.}, ``A comprehensive review on
  automation in agriculture using artificial intelligence,'' \emph{Artif.
  Intell. Agric.}, vol.~2, pp. 1--12, 2019.

\bibitem{bohr2020rise}
A.~Bohr and K.~Memarzadeh, ``The rise of artificial intelligence in healthcare
  applications,'' in \emph{Artif. Intell. Healthcare}.\hskip 1em plus 0.5em
  minus 0.4em\relax Elsevier, 2020, pp. 25--60.

\bibitem{cioffi2020artificial}
R.~Cioffi, M.~Travaglioni, G.~Piscitelli \emph{et~al.}, ``{Artificial
  Intelligence and Machine Learning Applications in Smart Production: Progress,
  Trends, and Directions},'' \emph{Sustainability}, vol.~12, no.~2, p. 492,
  2020.

\bibitem{alhayani2021effectiveness}
B.~Alhayani, H.~J. Mohammed, I.~Z. Chaloob \emph{et~al.}, ``{Effectiveness of
  artificial intelligence techniques against cyber security risks apply of IT
  industry},'' \emph{Materials Today: Proceedings}, vol. 531, 2021.

\bibitem{ahmad2021artificial}
T.~Ahmad, D.~Zhang, C.~Huang \emph{et~al.}, ``{Artificial intelligence in
  sustainable energy industry: Status Quo, challenges and opportunities},''
  \emph{J. Clean. Prod.}, vol. 289, pp. 1--31, 2021, {Art. no. 125834}.

\bibitem{kurd2007developing}
Z.~Kurd, T.~Kelly, and J.~Austin, ``{Developing artificial neural networks for
  safety critical systems},'' \emph{Neural Comput. Appl.}, vol.~16, pp. 11--19,
  2007.

\bibitem{forsberg2020challenges}
H.~Forsberg, J.~Lind{\'e}n, J.~Hjorth \emph{et~al.}, ``{Challenges in Using
  Neural Networks in Safety-Critical Applications},'' in \emph{AIAA/IEEE Digit.
  Avion. Syst. Conf. - Proc. (DASC)}, 2020, pp. 1--7.

\bibitem{zawacki2019systematic}
O.~Zawacki-Richter, V.~I. Mar{\'\i}n, M.~Bond \emph{et~al.}, ``{Systematic
  review of research on artificial intelligence applications in higher
  education--where are the educators?}'' \emph{Int. J. Educ. Technol. High.
  Educ.}, vol.~16, no.~1, pp. 1--27, 2019.

\bibitem{sezer2020financial}
O.~B. Sezer, M.~U. Gudelek, and A.~M. Ozbayoglu, ``{Financial time series
  forecasting with deep learning : A systematic literature review:
  2005–2019},'' \emph{Appl. Soft Comput. J.}, vol.~90, pp. 1--32, 2020, {Art.
  no. 106181}.

\bibitem{chen2020artificial}
L.~Chen, P.~Chen, and Z.~Lin, ``{Artificial Intelligence in Education: A
  Review},'' \emph{IEEE Access}, vol.~8, pp. 75\,264--75\,278, 2020.

\bibitem{abdallah2020artificial}
M.~Abdallah, M.~A. Talib, S.~Feroz \emph{et~al.}, ``{Artificial intelligence
  applications in solid waste management: A systematic research review},''
  \emph{Waste Manag.}, vol. 109, pp. 231--246, 2020.

\bibitem{khanagar2021developments}
S.~B. Khanagar, A.~Al-Ehaideb, P.~C. Maganur \emph{et~al.}, ``{Developments,
  application, and performance of artificial intelligence in dentistry--A
  systematic review},'' \emph{J. Dent. Sci.}, vol.~16, no.~1, pp. 508--522,
  2021.

\bibitem{pmlr-v37-romera-paredes15}
B.~Romera-Paredes and P.~Torr, ``{An embarrassingly simple approach to
  zero-shot learning},'' in \emph{Proc. 32nd Int. Conf. Mach. Learn.
  (ICML)}.\hskip 1em plus 0.5em minus 0.4em\relax PMLR, 2015, pp. 2152--2161.

\bibitem{vinyals2016matching}
O.~Vinyals, C.~Blundell, T.~Lillicrap \emph{et~al.}, ``{Matching Networks for
  One Shot Learning},'' in \emph{Proc. 29th Int. Conf. Neural Inf. Process.
  Syst. (NeurIPS)}, 2016.

\bibitem{kadam2020review}
S.~Kadam and V.~Vaidya, ``{Review and Analysis of Zero, One and Few Shot
  Learning Approaches},'' in \emph{Proc. 18th Int. Conf. Intell. Syst. Design
  Appl. (ISDA)}.\hskip 1em plus 0.5em minus 0.4em\relax Springer, 2018, pp.
  100--112.

\bibitem{vanschoren2019meta}
J.~Vanschoren, ``Meta-learning,'' in \emph{Automated Machine Learning: Methods,
  Systems, Challenges}.\hskip 1em plus 0.5em minus 0.4em\relax Springer, 2019,
  pp. 35--61.

\bibitem{neto2022safety}
A.~V.~S. Neto, J.~B. Camargo, J.~R. Almeida \emph{et~al.}, ``{Safety Assurance
  of Artificial Intelligence-Based Systems: A Systematic Literature Review on
  the State of the Art and Guidelines for Future Work},'' \emph{IEEE Access},
  vol.~10, pp. 130\,733--130\,770, 2022.

\bibitem{gillula2012guaranteed}
J.~H. Gillula and C.~J. Tomlin, ``{Guaranteed Safe Online Learning via
  Reachability: tracking a ground target using a quadrotor},'' in \emph{Proc.
  IEEE Int. Conf. Robot. Autom. (ICRA)}, 2012, pp. 2723--2730.

\bibitem{corso2021survey}
A.~Corso, R.~Moss, M.~Koren \emph{et~al.}, ``{A Survey of Algorithms for
  Black-Box Safety Validation of Cyber-Physical Systems},'' \emph{J. Artif.
  Intell. Res.}, vol.~72, pp. 377--428, 2021.

\bibitem{pei2017deepxplore}
K.~Pei, Y.~Cao, J.~Yang \emph{et~al.}, ``{DeepXplore: Automated Whitebox
  Testing of Deep Learning Systems},'' in \emph{Proc. 26th ACM Symp. Oper.
  Syst. Princ.}, 2017, pp. 1--18.

\bibitem{chen2020tensorfi}
Z.~Chen, N.~Narayanan, B.~Fang \emph{et~al.}, ``{TensorFI: A Flexible Fault
  Injection Framework for TensorFlow Applications},'' in \emph{Proc. IEEE 31st
  Int. Symp. on Software Reliability Engineering (ISSRE)}, 2020, pp. 426--435.

\bibitem{alemany2021jespipe}
S.~Alemany, J.~Nucciarone, and N.~Pissinou, ``{Jespipe: A Plugin-Based, Open
  MPI Framework for Adversarial Machine Learning Analysis},'' in \emph{Proc.
  IEEE Int. Conf. Big Data (Big Data)}.\hskip 1em plus 0.5em minus 0.4em\relax
  IEEE, 2021, pp. 3663--3670.

\bibitem{trusted2022github}
\BIBentryALTinterwordspacing
G.~Trusted-AI, ``{GitHub Trusted-AI/Adversarial-Robustness-Toolbox: Adversarial
  Robustness Toolbox (ART) Python Library for Machine Learning Security
  Evasion, Poisoning, Extraction, Inference Red and Blue Teams},'' 2022.
  [Online]. Available:
  \url{{https://github.com/TrustedAI/adversarial-robustness-toolbox}}
\BIBentrySTDinterwordspacing

\bibitem{machin2016smof}
M.~Machin, J.~Guiochet, H.~Waeselynck \emph{et~al.}, ``{SMOF: A safety
  monitoring framework for autonomous systems},'' \emph{IEEE Trans. Syst. Man
  Cybern.: Syst.}, vol.~48, no.~5, pp. 702--715, 2016.

\bibitem{schirmer2018considerations}
S.~Schirmer, C.~Torens, F.~Nikodem \emph{et~al.}, ``{Considerations of
  Artificial Intelligence Safety Engineering for Unmanned Aircraft},'' in
  \emph{Proc. 37th Int. Conf. on Computer Safety, Reliability, and Security
  (SAFECOMP)}.\hskip 1em plus 0.5em minus 0.4em\relax Springer, 2018, pp.
  465--472.

\bibitem{jia2022role}
Y.~Jia, J.~McDermid, T.~Lawton \emph{et~al.}, ``{The Role of Explainability in
  Assuring Safety of Machine Learning in Healthcare},'' \emph{IEEE Trans.
  Emerg. Top. Comput.}, vol.~10, no.~4, pp. 1746--1760, 2022.

\bibitem{liu2021dloam}
W.~Liu, W.~Sun, and Y.~Liu, ``{DLOAM: Real-time and Robust LiDAR SLAM System
  Based on CNN in Dynamic Urban Environments},'' \emph{IEEE Open J. Intell.
  Transp. Syst.}, 2021.

\bibitem{goodwin2001control}
G.~C. Goodwin, S.~F. Graebe, M.~E. Salgado \emph{et~al.}, \emph{{Control System
  Design}}.\hskip 1em plus 0.5em minus 0.4em\relax NJ, USA: Prentice-Hall,
  2001.

\bibitem{lunze2010regelungstechnik}
J.~Lunze, \emph{Regelungstechnik 1: Systemtheoretische Grundlagen, Analyse und
  Entwurf einschleifiger Regelungen}.\hskip 1em plus 0.5em minus 0.4em\relax
  Springer, 2016, vol.~10.

\bibitem{follinger2011laplace}
O.~F{\"o}llinger and K.~Mathias, ``{Laplace-, Fourier-und z-Transformation,
  10., {\"u}berarb},'' \emph{10. Aufl. VDE-Verlag, Berlin}, 2011.

\bibitem{fadali2012digital}
M.~S. Fadali and A.~Visioli, \emph{{Digital Control Engineering: Analysis and
  Design}}.\hskip 1em plus 0.5em minus 0.4em\relax Amsterdam, Netherlands:
  Elsevier Science, 2012.

\bibitem{nise2020control}
N.~S. Nise, \emph{{Control Systems Engineering}}.\hskip 1em plus 0.5em minus
  0.4em\relax John Wiley \& Sons, 2020.

\bibitem{zhang2018overview}
Y.~Zhang and Q.~Yang, ``An overview of multi-task learning,'' \emph{National
  Science Review}, vol.~5, no.~1, pp. 30--43, 2018.

\bibitem{chang2019neural}
Y.-C. Chang, N.~Roohi, and S.~Gao, ``{Neural Lyapunov Control},'' in
  \emph{Proc. 32nd Int. Conf. Neural Inf. Process. Syst. (NeurIPS)}, 2019, pp.
  1--10.

\bibitem{berberich2020data}
J.~Berberich, J.~K{\"o}hler, M.~A. M{\"u}ller \emph{et~al.}, ``{Data-Driven
  Model Predictive Control With Stability and Robustness Guarantees},''
  \emph{IEEE Trans. Autom. Control}, vol.~66, no.~4, pp. 1702--1717, 2020.

\bibitem{pauli2021training}
P.~Pauli, A.~Koch, J.~Berberich \emph{et~al.}, ``{Training Robust Neural
  Networks Using Lipschitz Bounds},'' \emph{IEEE Control Syst. Lett.}, vol.~6,
  pp. 121--126, 2021.

\bibitem{wabersich2021predictive}
K.~P. Wabersich and M.~N. Zeilinger, ``A predictive safety filter for
  learning-based control of constrained nonlinear dynamical systems,''
  \emph{Automatica}, vol. 129, p. 109597, 2021.

\bibitem{nagabandi2018deep}
A.~Nagabandi, C.~Finn, and S.~Levine, ``{Deep Online Learning Via
  Meta-Learning: Continual Adaptation for Model-Based RL},'' in \emph{Int.
  Conf. Learn. Represent. (ICLR)}, 2018, pp. 1--15.

\bibitem{wei2017online}
C.-Y. Wei, Y.-T. Hong, and C.-J. Lu, ``{Online Reinforcement Learning in
  Stochastic Games},'' in \emph{Proc. 30th Int. Conf. Neural Inf. Process.
  Syst. (NeurIPS)}, 2017, pp. 1--11.

\bibitem{dogru2021online}
O.~Dogru, N.~Wieczorek, K.~Velswamy \emph{et~al.}, ``{Online reinforcement
  learning for a continuous space system with experimental validation},''
  \emph{J. Process Control}, vol. 104, pp. 86--100, 2021.

\bibitem{berkenkamp2017safe}
F.~Berkenkamp, M.~Turchetta, A.~Schoellig \emph{et~al.}, ``{Safe Model-based
  Reinforcement Learning with Stability Guarantees},'' in \emph{Proc. 31st Int.
  Conf. Neural Inf. Process. Syst. (NeurIPS)}, 2017, pp. 1--11.

\bibitem{mao2019towards}
H.~Mao, M.~Schwarzkopf, H.~He \emph{et~al.}, ``{Towards Safe Online
  Reinforcement Learning in Computer Systems},'' in \emph{Proc. 33rd Int. Conf.
  Neural Inf. Process. Syst. (NeurIPS), Machine Learning for Systems Workshop},
  2019, pp. 1--9.

\bibitem{valiente2022robustness}
R.~Valiente, B.~Toghi, R.~Pedarsani \emph{et~al.}, ``{Robustness and
  Adaptability of Reinforcement Learning-Based Cooperative Autonomous Driving
  in Mixed-Autonomy Traffic},'' \emph{IEEE Open J. Intell. Transp. Syst.},
  vol.~3, pp. 397--410, 2022.

\bibitem{zheng2023learning}
H.~Zheng, C.~Chen, S.~Li \emph{et~al.}, ``{Learning-Based Safe Control for
  Robot and Autonomous Vehicle Using Efficient Safety Certificate},''
  \emph{IEEE Open J. Intell. Transp. Syst.}, vol.~4, pp. 419--430, 2023.

\bibitem{wabersich2023data}
K.~P. Wabersich, A.~J. Taylor, J.~J. Choi \emph{et~al.}, ``{Data-Driven Safety
  Filters: Hamilton-Jacobi Reachability, Control Barrier Functions, and
  Predictive Methods for Uncertain Systems},'' \emph{IEEE Control Syst. Mag.},
  vol.~43, no.~5, pp. 137--177, 2023.

\bibitem{leeman2023predictive}
A.~Leeman, J.~K{\"o}hler, S.~Bennani \emph{et~al.}, ``{Predictive safety filter
  using system level synthesis},'' in \emph{Proc. 5th Annual Learning for
  Dynamics and Control Conference (L4DC)}.\hskip 1em plus 0.5em minus
  0.4em\relax PMLR, 2023, pp. 1180--1192.

\bibitem{kouvaritakis2016model}
B.~Kouvaritakis and M.~Cannon, ``{Model Predictive Control},''
  \emph{Switzerland: Springer International Publishing}, vol.~38, pp. 13--56,
  2016.

\bibitem{hou2013model}
Z.-S. Hou and Z.~Wang, ``{From model-based control to data-driven control:
  Survey, classification and perspective},'' \emph{Inf. Sci.}, vol. 235, pp.
  3--35, 2013.

\bibitem{hou2016overview}
Z.~Hou, R.~Chi, and H.~Gao, ``{An Overview of Dynamic-Linearization-Based
  Data-Driven Control and Applications},'' \emph{IEEE Trans. Ind. Electron.},
  vol.~64, no.~5, pp. 4076--4090, 2016.

\bibitem{maupong2017data}
T.~M. Maupong and P.~Rapisarda, ``{Data-driven control: A behavioral
  approach},'' \emph{Syst. Control Lett.}, vol. 101, pp. 37--43, 2017.

\bibitem{de2019formulas}
C.~De~Persis and P.~Tesi, ``{Formulas for Data-Driven Control: Stabilization,
  Optimality, and Robustness},'' \emph{IEEE Trans. Autom. Control}, vol.~65,
  no.~3, pp. 909--924, 2019.

\bibitem{torrente2021data}
G.~Torrente, E.~Kaufmann, P.~F{\"o}hn \emph{et~al.}, ``Data-driven mpc for
  quadrotors,'' \emph{IEEE Robot. Autom. Lett.}, vol.~6, no.~2, pp. 3769--3776,
  2021.

\bibitem{markovsky2021behavioral}
I.~Markovsky and F.~D{\"o}rfler, ``{Behavioral systems theory in data-driven
  analysis, signal processing, and control},'' \emph{Annual Reviews in
  Control}, vol.~52, pp. 42--64, 2021.

\bibitem{rosolia2017learning}
U.~Rosolia and F.~Borrelli, ``{Learning model predictive control for iterative
  tasks. a data-driven control framework},'' \emph{IEEE Trans. Autom. Control},
  vol.~63, no.~7, pp. 1883--1896, 2017.

\bibitem{dorfler2022bridging}
F.~D{\"o}rfler, J.~Coulson, and I.~Markovsky, ``{Bridging Direct and Indirect
  Data-Driven Control Formulations via Regularizations and Relaxations},''
  \emph{IEEE Trans. Autom. Control}, vol.~68, no.~2, pp. 883--897, 2022.

\bibitem{xu2013adaptive}
D.~Xu, B.~Jiang, and P.~Shi, ``{Adaptive Observer Based Data-Driven Control for
  Nonlinear Discrete-Time Processes},'' \emph{IEEE Trans. Autom. Sci. Eng.},
  vol.~11, no.~4, pp. 1037--1045, 2013.

\bibitem{van2023behavioral}
H.~J. van Waarde, J.~Eising, M.~K. Camlibel \emph{et~al.}, ``{A Behavioral
  Approach to Data-Driven Control With Noisy Input–Output Data},'' \emph{IEEE
  Trans. Autom. Control}, 2023.

\bibitem{shukla2020flight}
D.~Shukla, R.~Lal, D.~Hauptman \emph{et~al.}, ``{Flight Test Validation of a
  Safety-Critical Neural Network Based Longitudinal Controller for a Fixed-Wing
  UAS},'' in \emph{AIAA Aviation Forum}, 2020, p. 3093.

\bibitem{scholkopf2012causal}
B.~Sch{\"o}lkopf, D.~Janzing, J.~Peters \emph{et~al.}, ``{On Causal and
  Anticausal Learning},'' \emph{arXiv preprint arXiv:1206.6471}, 2012.

\bibitem{sugiyama2007covariate}
M.~Sugiyama, M.~Krauledat, and K.-R. M{\"u}ller, ``{Covariate Shift Adaptation
  by Importance Weighted Cross Validation},'' \emph{J. Mach. Learn. Res.},
  vol.~8, no.~5, 2007.

\bibitem{zhang2013domain}
K.~Zhang, B.~Sch{\"o}lkopf, K.~Muandet \emph{et~al.}, ``{Domain Adaptation
  under Target and Conditional Shift},'' in \emph{Proc. 32th Int. Conf. Mach.
  Learn. (ICML)}.\hskip 1em plus 0.5em minus 0.4em\relax PMLR, 2013, pp.
  819--827.

\bibitem{schulam2017reliable}
P.~Schulam and S.~Saria, ``{Reliable Decision Support using Counterfactual
  Models},'' in \emph{Proc. 30th Int. Conf. Neural Inf. Process. Syst.
  (NeurIPS)}, 2017, pp. 1--12.

\bibitem{pearl2009causality}
J.~Pearl, \emph{Causality}.\hskip 1em plus 0.5em minus 0.4em\relax Cambridge
  University Press, 2009.

\bibitem{shriram2025towards}
S.~Shriram, S.~Perisetla, A.~Keskar \emph{et~al.}, ``Towards a multi-agent
  vision-language system for zero-shot novel hazardous object detection for
  autonomous driving safety,'' \emph{IEEE RAS Conference on Automation Science
  and Engineering}, 2025.

\bibitem{greer2024perception}
R.~Greer and M.~Trivedi, ``Perception without vision for trajectory prediction:
  Ego vehicle dynamics as scene representation for efficient active learning in
  autonomous driving,'' \emph{IEEE ITSS Intelligent Transportation Systems
  Conference}, 2025.

\bibitem{wang2020generalizing}
Y.~Wang, Q.~Yao, J.~T. Kwok \emph{et~al.}, ``{Generalizing from a Few Examples:
  A Survey on Few-shot Learning},'' \emph{ACM Comput. Surv.}, vol.~53, no.~3,
  pp. 1--34, 2020.

\bibitem{greer2025language}
R.~Greer, B.~Antoniussen, A.~M{\o}gelmose \emph{et~al.}, ``Language-driven
  active learning for diverse open-set 3d object detection,'' in
  \emph{Proceedings of the Winter Conference on Applications of Computer
  Vision}, 2025, pp. 980--988.

\bibitem{greer2024towards}
R.~Greer and M.~Trivedi, ``Towards explainable, safe autonomous driving with
  language embeddings for novelty identification and active learning: Framework
  and experimental analysis with real-world data sets,'' \emph{arXiv preprint
  arXiv:2402.07320}, 2024.

\bibitem{hochreiter2001learning}
S.~Hochreiter, A.~S. Younger, and P.~R. Conwell, ``{Learning to Learn Using
  Gradient Descent},'' in \emph{Proc. 11st Int. Conf. Art. Neural Netw.
  (ICANN)}.\hskip 1em plus 0.5em minus 0.4em\relax Springer, 2001, pp. 87--94.

\bibitem{finn2017model}
C.~Finn, P.~Abbeel, and S.~Levine, ``{Model-Agnostic Meta-Learning for Fast
  Adaptation of Deep Networks},'' in \emph{Proc. 34th Int. Conf. Mach. Learn.
  (ICML)}.\hskip 1em plus 0.5em minus 0.4em\relax PMLR, 2017, pp. 1126--1135.

\bibitem{esfahani2015data}
P.~Mohajerin~Esfahani and D.~Kuhn, ``{Data-driven distributionally robust
  optimization using the Wasserstein metric: performance guarantees and
  tractable reformulations},'' \emph{Math. Program.}, vol. 171, no.~1, pp.
  115--166, 2018.

\bibitem{sinha2017certifying}
A.~Sinha, H.~Namkoong, R.~Volpi \emph{et~al.}, ``{Certifiable Distributional
  Robustness with Principled Adversarial Training},'' \emph{arXiv preprint
  arXiv:1710.10571}, 2017.

\bibitem{rahimian2019distributionally}
H.~Rahimian and S.~Mehrotra, ``{Distributionally Robust Optimization: A
  Review},'' \emph{arXiv preprint arXiv:1908.05659}, 2019.

\bibitem{subbaswamy2020development}
A.~Subbaswamy and S.~Saria, ``{From development to deployment: dataset shift,
  causality, and shift-stable models in health AI},'' \emph{Biostatistics},
  vol.~21, no.~2, pp. 345--352, 2020.

\bibitem{subbaswamy2021evaluating}
A.~Subbaswamy, R.~Adams, and S.~Saria, ``Evaluating model robustness and
  stability to dataset shift,'' in \emph{International conference on artificial
  intelligence and statistics}.\hskip 1em plus 0.5em minus 0.4em\relax PMLR,
  2021, pp. 2611--2619.

\bibitem{amodei2016concrete}
D.~Amodei, C.~Olah, J.~Steinhardt \emph{et~al.}, ``Concrete problems in ai
  safety,'' \emph{arXiv preprint arXiv:1606.06565}, 2016.

\bibitem{afxentiou2025evaluation}
V.~Afxentiou and T.~Vladimirova, ``{Evaluation of CNN-Based Approaches to
  Adverse Weather Image Classification for Autonomous Driving Systems},''
  \emph{IEEE Open J. Intell. Transp. Syst.}, 2025.

\bibitem{subbaswamy2023causal}
A.~Subbaswamy \emph{et~al.}, ``{Causal Modeling for Training and Evaluating
  Dataset Shift-Stable Machine Learning Models in Healthcare},'' Ph.D.
  dissertation, Johns Hopkins University, 2023.

\bibitem{rumelhart1986learning}
D.~E. Rumelhart, G.~E. Hinton, and R.~J. Williams, ``Learning representations
  by back-propagating errors,'' \emph{Nature}, vol. 323, no. 6088, pp.
  533--536, 1986.

\bibitem{hochreiter1997long}
S.~Hochreiter and J.~Schmidhuber, ``{Long Short-Term Memory},'' \emph{Neural
  Comput.}, vol.~9, no.~8, pp. 1735--1780, 1997.

\bibitem{lecun2022path}
Y.~LeCun, ``{A Path Towards Autonomous Machine Intelligence Version 0.9. 2,
  2022-06-27},'' \emph{Open Review}, vol.~62, 2022.

\bibitem{OECDpub}
\BIBentryALTinterwordspacing
OECD, ``{Scoping the OECD AI principles},'' \emph{OECDpublishing}, no. 291,
  2019. [Online]. Available:
  \url{https://www.oecd-ilibrary.org/content/paper/d62f618a-en}
\BIBentrySTDinterwordspacing

\bibitem{OECD_AI_safe}
\BIBentryALTinterwordspacing
------, ``{Robustness, security and safety (Principle 1.4)},'' 2025. [Online].
  Available: \url{https://oecd.ai/en/dashboards/ai-principles/P8}
\BIBentrySTDinterwordspacing

\bibitem{bengio2024international}
Y.~Bengio, S.~Mindermann, D.~Privitera \emph{et~al.}, ``{International
  Scientific Report on the Safety of Advanced AI (Interim Report)},''
  \emph{arXiv preprint arXiv:2412.05282}, 2024.

\bibitem{ruess2022safe}
H.~Rue{\ss} and S.~Burton, ``Safe ai--how is this possible?'' \emph{arXiv
  preprint arXiv:2201.10436}, 2022.

\bibitem{NIST_AI_safe}
\BIBentryALTinterwordspacing
N.~I. of~Standards and T.~(NIST), ``{The United States Artificial Intelligence
  Safety Institute: Vision, Mission, and Strategic Goals},'' 2024. [Online].
  Available:
  \url{https://www.nist.gov/system/files/documents/2024/05/21/AISI-vision-21May2024.pdf}
\BIBentrySTDinterwordspacing

\bibitem{herrera2025responsible}
A.~Herrera-Poyatos, J.~Del~Ser, M.~L. de~Prado \emph{et~al.}, ``{Responsible
  Artificial Intelligence Systems: A Roadmap to Society's Trust through
  Trustworthy AI, Auditability, Accountability, and Governance},'' \emph{arXiv
  preprint arXiv:2503.04739}, 2025.

\bibitem{hendrycks2025introduction}
D.~Hendrycks, \emph{{Introduction to AI Safety, Ethics, and Society}}.\hskip
  1em plus 0.5em minus 0.4em\relax Taylor \& Francis, 2025.

\bibitem{li2023probabilities}
A.~Li and J.~Pearl, ``{Probabilities of Causation: Role of Observational
  Data},'' in \emph{Proc. 26th Int. Conf. Artif. Intell. Statist.
  (AISTATS)}.\hskip 1em plus 0.5em minus 0.4em\relax PMLR, 2023, pp.
  10\,012--10\,027.

\bibitem{bissoto2024even}
A.~Bissoto, C.~Barata, E.~Valle \emph{et~al.}, ``Even small correlation and
  diversity shifts pose dataset-bias issues,'' \emph{Pattern Recognit. Lett.},
  2024.

\bibitem{holzinger2021next}
A.~Holzinger, ``{The Next Frontier: AI We Can Really Trust},'' in \emph{Machine
  Learning and Principles and Practice of Knowledge Discovery in
  Databases}.\hskip 1em plus 0.5em minus 0.4em\relax Cham: Springer
  International Publishing, 2021, pp. 427--440.

\bibitem{storkey2008training}
A.~Storkey, ``{When Training and Test Sets Are Different: Characterizing
  Learning Transfer},'' in \emph{Dataset Shift in Machine Learning},
  J.~Quiñonero-Candela, M.~Sugiyama, A.~Schwaighofer \emph{et~al.}, Eds.\hskip
  1em plus 0.5em minus 0.4em\relax Cambridge, MA, USA: MIT Press, 2009, pp.
  3--28.

\bibitem{reichenbach1991direction}
H.~Reichenbach, \emph{{The Direction of Time}}.\hskip 1em plus 0.5em minus
  0.4em\relax Univ. of California Press, 1991, vol.~65.

\bibitem{scholkopf2021toward}
B.~Sch{\"o}lkopf, F.~Locatello, S.~Bauer \emph{et~al.}, ``{Toward Causal
  Representation Learning},'' \emph{Proc. IEEE}, vol. 109, no.~5, pp. 612--634,
  2021.

\bibitem{takahashi2020review}
C.~C. Takahashi and A.~P. Braga, ``{A Review of Off-Line Mode Dataset
  Shifts},'' \emph{IEEE Comput. Intell. Mag.}, vol.~15, no.~3, pp. 16--27,
  2020.

\bibitem{zhou2022domain}
K.~Zhou, Z.~Liu, Y.~Qiao \emph{et~al.}, ``{Domain Generalization: A Survey},''
  \emph{IEEE Trans. Pattern Anal. Mach. Intell.}, vol.~45, no.~4, pp.
  4396--4415, 2022.

\bibitem{shimodaira2000improving}
H.~Shimodaira, ``{Improving predictive inference under covariate shift by
  weighting the log-likelihood function},'' \emph{J. Stat. Plan. Inference},
  vol.~90, no.~2, pp. 227--244, 2000.

\bibitem{sugiyama2007direct}
M.~Sugiyama, S.~Nakajima, H.~Kashima \emph{et~al.}, ``{Direct Importance
  Estimation with Model Selection and Its Application to Covariate Shift
  Adaptation},'' in \emph{Proc. 20th Int. Conf. Neural Inf. Process. Syst.
  (NeurIPS)}, 2007, pp. 1--8.

\bibitem{raza2013dataset}
H.~Raza, G.~Prasad, and Y.~Li, ``{Dataset Shift Detection in Non-stationary
  Environments Using EWMA Charts},'' in \emph{Proc. IEEE Int. Conf. Syst. Man
  Cybern.: Syst.}, 2013, pp. 3151--3156.

\bibitem{otte1985probabilistic}
R.~Otte, ``{Probabilistic Causality and Simpson's Paradox},'' \emph{Philosophy
  of Science}, vol.~52, no.~1, pp. 110--125, 1985.

\bibitem{gopnik2004theory}
A.~Gopnik, C.~Glymour, D.~M. Sobel \emph{et~al.}, ``{A Theory of Causal
  Learning in Children: Causal Maps and Bayes Nets.}'' \emph{Psychol. Rev.},
  vol. 111, no.~1, p.~3, 2004.

\bibitem{gopnik2007causal}
A.~Gopnik, L.~Schulz, and L.~E. Schulz, \emph{{Causal Learning: Psychology,
  Philosophy, and Computation}}.\hskip 1em plus 0.5em minus 0.4em\relax Oxford
  University Press, 2007.

\bibitem{holyoak2011causal}
K.~J. Holyoak and P.~W. Cheng, ``{Causal Learning and Inference as a Rational
  Process: The New Synthesis},'' \emph{Annu. Rev. Psychol.}, vol.~62, pp.
  135--163, 2011.

\bibitem{locatello2019challenging}
F.~Locatello, S.~Bauer, M.~Lucic \emph{et~al.}, ``{Challenging Common
  Assumptions in the Unsupervised Learning of Disentangled Representations},''
  in \emph{Proc. 36th Int. Conf. Mach. Learn. (ICML)}.\hskip 1em plus 0.5em
  minus 0.4em\relax PMLR, 2019, pp. 4114--4124.

\bibitem{tran2017disentangled}
L.~Tran, X.~Yin, and X.~Liu, ``{Disentangled Representation Learning GAN for
  Pose-Invariant Face Recognition},'' in \emph{Proc. IEEE/CVF Comput. Soc.
  Conf. Comput. Vis. Pattern Recognit. (CVPR)}, 2017, pp. 1415--1424.

\bibitem{cheng2023disentangled}
H.~Cheng, Y.~Wang, H.~Li \emph{et~al.}, ``{Disentangled Feature Representation
  for Few-Shot Image Classification},'' \emph{IEEE Trans. Neural Netw. Learn.
  Syst.}, 2023.

\bibitem{higgins2017beta}
I.~Higgins, L.~Matthey, A.~Pal \emph{et~al.}, ``{beta-VAE: Learning Basic
  Visual Concepts with a Constrained Variational Framework},'' in \emph{Int.
  Conf. Learn. Represent. (ICLR)}, 2017, pp. 1--22.

\bibitem{kim2018disentangling}
H.~Kim and A.~Mnih, ``{Disentangling by Factorising},'' in \emph{Proc. 35th
  Int. Conf. Mach. Learn. (ICML)}.\hskip 1em plus 0.5em minus 0.4em\relax PMLR,
  2018, pp. 2649--2658.

\bibitem{vapnik1998statistical}
V.~N. Vapnik, V.~Vapnik \emph{et~al.}, \emph{{Statistical Learning
  Theory}}.\hskip 1em plus 0.5em minus 0.4em\relax New York, NY, USA: Wiley,
  1998.

\bibitem{caruana1997multitask}
R.~Caruana, ``{Multitask Learning},'' \emph{Machine Learning}, vol.~28, pp.
  41--75, 1997.

\bibitem{10.1145/3293318}
W.~Wang, V.~W. Zheng, H.~Yu \emph{et~al.}, ``{A Survey of Zero-Shot Learning:
  Settings, Methods, and Applications},'' \emph{ACM Trans. Intell. Syst.
  Technol.}, vol.~10, no.~2, 2019.

\bibitem{ganin2015unsupervised}
Y.~Ganin and V.~Lempitsky, ``{Unsupervised Domain Adaptation by
  Backpropagation},'' in \emph{Proc. 32nd Int. Conf. Mach. Learn.
  (ICML)}.\hskip 1em plus 0.5em minus 0.4em\relax PMLR, 2015, pp. 1180--1189.

\bibitem{saito2018maximum}
K.~Saito, K.~Watanabe, Y.~Ushiku \emph{et~al.}, ``{Maximum Classifier
  Discrepancy for Unsupervised Domain Adaptation},'' in \emph{Proc. IEEE/CVF
  Comput. Soc. Conf. Comput. Vis. Pattern Recognit. (CVPR)}, 2018, pp.
  3723--3732.

\bibitem{saenko2010adapting}
K.~Saenko, B.~Kulis, M.~Fritz \emph{et~al.}, ``{Adapting Visual Category Models
  to New Domains},'' in \emph{Proc. 11th IEEE Europ. Conf. Comp. Vision
  (ECCV)}.\hskip 1em plus 0.5em minus 0.4em\relax Springer, 2010, pp. 213--226.

\bibitem{pan2009survey}
S.~J. Pan and Q.~Yang, ``{A Survey on Transfer Learning},'' \emph{IEEE Trans.
  Knowl. Data Eng.}, vol.~22, no.~10, pp. 1345--1359, 2009.

\bibitem{blanchard2011generalizing}
G.~Blanchard, G.~Lee, and C.~Scott, ``{Generalizing from Several Related
  Classification Tasks to a New Unlabeled Sample},'' in \emph{Proc. 24th Int.
  Conf. Neural Inf. Process. Syst. (NeurIPS)}, 2011, pp. 1--9.

\bibitem{roese1994functional}
N.~J. Roese, ``{The Functional Basis Of Counterfactual Thinking},'' \emph{J.
  Pers. Soc. Psychol.}, vol.~66, no.~5, p. 805, 1994.

\bibitem{lu2020sample}
C.~Lu, B.~Huang, K.~Wang \emph{et~al.}, ``{Sample-Efficient Reinforcement
  Learning via Counterfactual-Based Data Augmentation},'' \emph{arXiv preprint
  arXiv:2012.09092}, 2020.

\bibitem{buesing2018woulda}
L.~Buesing, T.~Weber, Y.~Zwols \emph{et~al.}, ``{Woulda, Coulda, Shoulda:
  Counterfactually-Guided Policy Search},'' \emph{arXiv preprint
  arXiv:1811.06272}, 2018.

\bibitem{sharma2017activation}
S.~Sharma, S.~Sharma, and A.~Athaiya, ``{Activation Functions in Neural
  Networks},'' \emph{Int. J. Eng. Sci. Technol.}, vol.~6, no.~12, pp. 310--316,
  2017.

\bibitem{zhang2018new}
Z.~Zhang, Y.~Lu, L.~Zheng \emph{et~al.}, ``{A New Varying-Parameter
  Convergent-Differential Neural-Network for Solving Time-Varying Convex QP
  Problem Constrained by Linear-Equality},'' \emph{IEEE Trans. Autom. Control},
  vol.~63, no.~12, pp. 4110--4125, 2018.

\bibitem{an2020novel}
Z.~An, S.~Li, J.~Wang \emph{et~al.}, ``{A novel bearing intelligent fault
  diagnosis framework under time-varying working conditions using recurrent
  neural network},'' \emph{ISA Transactions}, vol. 100, pp. 155--170, 2020.

\bibitem{hua2023dynamic}
C.~Hua, X.~Cao, Q.~Xu \emph{et~al.}, ``{Dynamic Neural Network Models for
  Time-Varying Problem Solving: A Survey on Model Structures}s,'' \emph{IEEE
  Access}, 2023.

\bibitem{minsky1969introduction}
M.~Minsky and S.~Papert, ``{An Introduction to Computational Geometry},''
  \emph{Cambridge tiass., HIT}, vol. 479, no. 480, p. 104, 1969.

\bibitem{fukushima1980neocognitron}
K.~Fukushima, ``{Neocognitron: A self-organizing neural network model for a
  mechanism of pattern recognition unaffected by shift in position},''
  \emph{Biol. Cybern.}, vol.~36, no.~4, pp. 193--202, 1980.

\bibitem{lecun1998gradient}
Y.~LeCun, L.~Bottou, Y.~Bengio \emph{et~al.}, ``{Gradient-based learning
  applied to document recognition},'' \emph{Proc. IEEE}, vol.~86, no.~11, pp.
  2278--2324, 1998.

\bibitem{scarselli2008graph}
F.~Scarselli, M.~Gori, A.~C. Tsoi \emph{et~al.}, ``{The Graph Neural Network
  Model},'' \emph{IEEE Trans. Neural Netw. Learn. Syst.}, vol.~20, no.~1, pp.
  61--80, 2008.

\bibitem{baldi2012autoencoders}
P.~Baldi, ``{Autoencoders, Unsupervised Learning, and Deep Architectures},'' in
  \emph{Proc. 35th Int. Conf. Mach. Learn. (ICML), Workshop on Unsupervised and
  Transfer Learning}.\hskip 1em plus 0.5em minus 0.4em\relax JMLR Workshop and
  Conference Proceedings, 2012, pp. 37--49.

\bibitem{kingma2013auto}
D.~P. Kingma and M.~Welling, ``Auto-encoding variational bayes,'' in
  \emph{Proc. 2nd Int. Conf. Learn. Represent. (ICLR)}, 2013.

\bibitem{jaeger2004harnessing}
H.~Jaeger and H.~Haas, ``{Harnessing Nonlinearity: Predicting Chaotic Systems
  and Saving Energy in Wireless Communication},'' \emph{Science}, vol. 304, no.
  5667, pp. 78--80, 2004.

\bibitem{cho2014learning}
K.~Cho, B.~Van~Merri{\"e}nboer, C.~Gulcehre \emph{et~al.}, ``{Learning phrase
  representations using RNN encoder-decoder for statistical machine
  translation},'' \emph{arXiv preprint arXiv:1406.1078}, 2014.

\bibitem{jaeger2007echo}
H.~Jaeger, ``Echo state network,'' \emph{scholarpedia}, vol.~2, no.~9, p. 2330,
  2007.

\bibitem{qiao2014online}
J.~Qiao, Z.~Zhang, and Y.~Bo, ``{An online self-adaptive modular neural network
  for time-varying systems},'' \emph{Neurocomputing}, vol. 125, pp. 7--16,
  2014.

\bibitem{liu2019adaptive}
Y.-J. Liu, L.~Ma, L.~Liu \emph{et~al.}, ``{Adaptive Neural Network Learning
  Controller Design for a Class of Nonlinear Systems With Time-Varying State
  Constraints},'' \emph{IEEE Trans. Neural Netw. Learn. Syst.}, vol.~31, no.~1,
  pp. 66--75, 2019.

\bibitem{guo2012novel}
D.~Guo and Y.~Zhang, ``{Novel Recurrent Neural Network for Time-Varying
  Problems Solving [Research Frontier]},'' \emph{IEEE Comput. Intell. Mag.},
  vol.~7, no.~4, pp. 61--65, 2012.

\bibitem{ullrich2023cnp}
L.~Ullrich, A.~Völz, and K.~Graichen, ``{Robust Meta-Learning of Vehicle Yaw
  Rate Dynamics via Conditional Neural Processes},'' in \emph{Proc. 62nd IEEE
  Conf. Decis. Control (CDC)}, 2023, pp. 2611--2619.

\bibitem{foellinger1983statespace}
O.~F{\"o}llinger and D.~Franke, \emph{{Einf{\"u}hrung in die
  Zustandsbeschreibung dynamischer Systeme}}.\hskip 1em plus 0.5em minus
  0.4em\relax München: Oldenbourg, 1982.

\bibitem{liao2002lmi}
X.~Liao, G.~Chen, and E.~N. Sanchez, ``{LMI-based approach for asymptotically
  stability analysis of delayed neural networks},'' \emph{IEEE Trans. Circuits
  Syst. I: Fundamental Theory and Applications}, vol.~49, no.~7, pp.
  1033--1039, 2002.

\bibitem{yang2005stability}
Z.~Yang and D.~Xu, ``{Stability Analysis of Delay Neural Networks With
  Impulsive Effects},'' \emph{IEEE Trans. Circuits Syst. II: Express Briefs},
  vol.~52, no.~8, pp. 517--521, 2005.

\bibitem{wang2006stability}
Z.~Wang, Y.~Liu, M.~Li \emph{et~al.}, ``{Stability analysis for stochastic
  Cohen-Grossberg neural networks with mixed time delays},'' \emph{IEEE Trans.
  Neural Netw.}, vol.~17, no.~3, pp. 814--820, 2006.

\bibitem{wan2010exponential}
L.~Wan and Q.~Zhou, ``{Exponential stability of impulsive Cohen-Grossberg-type
  BAM neural networks with delays and diffusion terms},'' in \emph{Proc. 6th
  IEEE Int. Conf. Nat. Comput. (ICNC))}, vol.~1.\hskip 1em plus 0.5em minus
  0.4em\relax IEEE, 2010, pp. 282--286.

\bibitem{zhang2014comprehensive}
H.~Zhang, Z.~Wang, and D.~Liu, ``{A Comprehensive Review of Stability Analysis
  of Continuous-Time Recurrent Neural Networks},'' \emph{IEEE Trans. Neural
  Netw. Learn. Syst.}, vol.~25, no.~7, pp. 1229--1262, 2014.

\bibitem{kim2018standard}
K.-K.~K. Kim, E.~R. Patr{\'o}n, and R.~D. Braatz, ``{Standard representation
  and unified stability analysis for dynamic artificial neural network
  models},'' \emph{Neural Netw.}, vol.~98, pp. 251--262, 2018.

\bibitem{fazlyab2020safety}
M.~Fazlyab, M.~Morari, and G.~J. Pappas, ``{Safety Verification and Robustness
  Analysis of Neural Networks via Quadratic Constraints and Semidefinite
  Programming},'' \emph{IEEE Trans. Autom. Control}, vol.~67, no.~1, pp. 1--15,
  2020.

\bibitem{shi2020artificial}
Z.~Shi, W.~Yao, Z.~Li \emph{et~al.}, ``Artificial intelligence techniques for
  stability analysis and control in smart grids: Methodologies, applications,
  challenges and future directions,'' \emph{Appl. Energy}, vol. 278, p. 115733,
  2020.

\bibitem{jin2020stability}
M.~Jin and J.~Lavaei, ``{Stability-Certified Reinforcement Learning: A
  Control-Theoretic Perspective},'' \emph{IEEE Access}, vol.~8, pp.
  229\,086--229\,100, 2020.

\bibitem{hu2020reach}
H.~Hu, M.~Fazlyab, M.~Morari \emph{et~al.}, ``{Reach-SDP: Reachability Analysis
  of Closed-Loop Systems with Neural Network Controllers via Semidefinite
  Programming},'' in \emph{Proc. 59th IEEE Conf. Decis. Control (CDC)}.\hskip
  1em plus 0.5em minus 0.4em\relax IEEE, 2020, pp. 5929--5934.

\bibitem{yin2021stability}
H.~Yin, P.~Seiler, and M.~Arcak, ``{Stability Analysis Using Quadratic
  Constraints for Systems With Neural Network Controllers},'' \emph{IEEE Trans.
  Autom. Control}, vol.~67, no.~4, pp. 1980--1987, 2021.

\bibitem{kaur2022trustworthy}
D.~Kaur, S.~Uslu, K.~J. Rittichier \emph{et~al.}, ``{Trustworthy Artificial
  Intelligence: A Review},'' \emph{ACM Comput. Surv. (CSUR)}, vol.~55, no.~2,
  pp. 1--38, 2022.

\bibitem{chao2022fusing}
M.~A. Chao, C.~Kulkarni, K.~Goebel \emph{et~al.}, ``{Fusing physics-based and
  deep learning models for prognostics},'' \emph{Reliab. Eng. Syst. Saf.}, vol.
  217, p. 107961, 2022.

\bibitem{seff2023motionlm}
A.~Seff, B.~Cera, D.~Chen \emph{et~al.}, ``{MotionLM: Multi-Agent Motion
  Forecasting as Language Modeling},'' in \emph{Proc. IEEE/CVF Int. Conf.
  Comput. Vis. (ICCV)}, 2023, pp. 8579--8590.

\bibitem{gunning2019xai}
D.~Gunning, M.~Stefik, J.~Choi \emph{et~al.}, ``{XAI-Explainable artificial
  intelligence},'' \emph{Sci. Robot.}, vol.~4, no.~37, p. eaay7120, 2019.

\bibitem{grushin2019decoding}
A.~Grushin, J.~Nanda, A.~Tyagi \emph{et~al.}, ``{Decoding the Black Box:
  Extracting Explainable Decision Boundary Approximations from Machine Learning
  Models for Real Time Safety Assurance of the National Airspace},'' in
  \emph{AIAA Scitech Forum}, 2019, p. 136.

\bibitem{confalonieri2021historical}
R.~Confalonieri, L.~Coba, B.~Wagner \emph{et~al.}, ``{A historical perspective
  of explainable Artificial Intelligence},'' \emph{Wiley Interdiscip. Rev.:
  Data Min. Knowl. Discov.}, vol.~11, no.~1, p. e1391, 2021.

\bibitem{pearl2019seven}
J.~Pearl, ``{The seven tools of causal inference, with reflections on machine
  learning},'' \emph{Commun. ACM}, vol.~62, no.~3, pp. 54--60, 2019.

\bibitem{scholkopf2022causality}
B.~Sch{\"o}lkopf, ``{Causality for Machine Learning},'' in \emph{Probabilistic
  and Causal Inference: The Works of Judea Pearl}, 2022, pp. 765--804.

\bibitem{subbaswamy2018counterfactual}
A.~Subbaswamy and S.~Saria, ``{Counterfactual Normalization: Proactively
  Addressing Dataset Shift and Improving Reliability Using Causal
  Mechanisms},'' \emph{arXiv preprint arXiv:1808.03253}, 2018.

\bibitem{subbaswamy2019preventing}
A.~Subbaswamy, P.~Schulam, and S.~Saria, ``{Preventing Failures Due to Dataset
  Shift: Learning Predictive Models That Transport},'' in \emph{Proc. 22th Int.
  Conf. Artif. Intell. Statist. (AISTATS)}.\hskip 1em plus 0.5em minus
  0.4em\relax PMLR, 2019, pp. 3118--3127.

\bibitem{liu2020input}
J.~Liu, L.~Chen, A.~Mine \emph{et~al.}, ``{Input Validation for Neural Networks
  via Runtime Local Robustness Verification},'' \emph{arXiv preprint
  arXiv:2002.03339}, 2020.

\bibitem{sabokrou2018adversarially}
M.~Sabokrou, M.~Khalooei, M.~Fathy \emph{et~al.}, ``{Adversarially Learned
  One-Class Classifier for Novelty Detection},'' in \emph{Proc. IEEE/CVF
  Comput. Soc. Conf. Comput. Vis. Pattern Recognit. (CVPR)}, 2018, pp.
  3379--3388.

\bibitem{cheng2019runtime}
C.-H. Cheng, G.~N{\"u}hrenberg, and H.~Yasuoka, ``{Runtime Monitoring Neuron
  Activation Patterns},'' in \emph{Proc. IEEE Des. Autom. Test Eur. (DATE)},
  2019, pp. 300--303.

\bibitem{henzinger2019outside}
T.~A. Henzinger, A.~Lukina, and C.~Schilling, ``{Outside the Box:
  Abstraction-Based Monitoring of Neural Networks},'' in \emph{ECAI
  2020}.\hskip 1em plus 0.5em minus 0.4em\relax IOS Press, 2020, pp.
  2433--2440.

\bibitem{hendrycks2016baseline}
D.~Hendrycks and K.~Gimpel, ``{A Baseline for Detecting Misclassified and
  Out-of-Distribution Examples in Neural Networks},'' in \emph{Proc. Int. Conf.
  Learn. Represent. (ICLR)}, 2017, pp. 1--9.

\bibitem{liang2017enhancing}
S.~Liang, Y.~Li, and R.~Srikant, ``{Enhancing The Reliability of
  Out-of-distribution Image Detection in Neural Networks},'' in \emph{Proc.
  Int. Conf. Learn. Represent. (ICLR)}, 2017, pp. 1--27.

\bibitem{mooij2013ordinary}
J.~M. Mooij, D.~Janzing, and B.~Sch{\"o}lkopf, ``{From Ordinary Differential
  Equations to Structural Causal Models: the deterministic case},'' in
  \emph{Proc. 29th Conf. Uncert. Artif. Intell. (UAI)}.\hskip 1em plus 0.5em
  minus 0.4em\relax AUAI Press, 2013, pp. 440--448.

\bibitem{murray2002models}
J.~Murray, ``{Models for Interacting Populations},'' \emph{Mathematical
  Biology: I. An Introduction}, pp. 79--118, 2002.

\bibitem{lu2018beyond}
Y.~Lu, A.~Zhong, Q.~Li \emph{et~al.}, ``{Beyond Finite Layer Neural Networks:
  Bridging Deep Architectures and Numerical Differential Equations},'' in
  \emph{Proc. 35th Int. Conf. Mach. Learn. (ICML)}.\hskip 1em plus 0.5em minus
  0.4em\relax PMLR, 2018, pp. 3276--3285.

\bibitem{haber2017stable}
E.~Haber and L.~Ruthotto, ``{Stable architectures for deep neural networks},''
  \emph{Inverse Probl.}, vol.~34, no.~1, p. 014004, 2017.

\bibitem{schramm2014vehicle}
D.~Schramm, M.~Hiller, and R.~Bardini, ``{Vehicle Dynamics},'' \emph{Modeling
  and Simulation. Berlin, Heidelberg}, vol. 151, 2014.

\bibitem{huang2018neural}
C.-W. Huang, D.~Krueger, A.~Lacoste \emph{et~al.}, ``{Neural Autoregressive
  Flows},'' in \emph{Proc. 35th Int. Conf. Mach. Learn. (ICML)}.\hskip 1em plus
  0.5em minus 0.4em\relax PMLR, 2018, pp. 2078--2087.

\bibitem{kaiser2018fast}
L.~Kaiser, S.~Bengio, A.~Roy \emph{et~al.}, ``{Fast Decoding in Sequence Models
  Using Discrete Latent Variables},'' in \emph{Proc. 35th Int. Conf. Mach.
  Learn. (ICML)}.\hskip 1em plus 0.5em minus 0.4em\relax PMLR, 2018, pp.
  2390--2399.

\bibitem{choe2017probabilistic}
Y.~J. Choe, J.~Shin, and N.~Spencer, ``{Probabilistic Interpretations of
  Recurrent Neural Networks},'' \emph{Probabilistic Graphical Models}, 2017.

\bibitem{bitzer2012recognizing}
S.~Bitzer and S.~J. Kiebel, ``{Recognizing recurrent neural networks (rRNN):
  Bayesian inference for recurrent neural networks},'' \emph{Biol. Cybern.},
  vol. 106, pp. 201--217, 2012.

\bibitem{eddy1996hidden}
S.~R. Eddy, ``{Hidden Markov models},'' \emph{Curr. Opin. Struct. Biol.},
  vol.~6, no.~3, pp. 361--365, 1996.

\bibitem{baum1966statistical}
L.~E. Baum and T.~Petrie, ``{Statistical Inference for Probabilistic Functions
  of Finite State Markov Chains},'' \emph{Ann. Math. Stat.}, vol.~37, no.~6,
  pp. 1554--1563, 1966.

\bibitem{jazwinski2007stochastic}
A.~H. Jazwinski, \emph{{Stochastic Processes and Filtering Theory}}.\hskip 1em
  plus 0.5em minus 0.4em\relax Courier Corporation, 2007.

\bibitem{doucet2003parameter}
A.~Doucet and V.~B. Tadi{\'c}, ``{Parameter estimation in general state-space
  models using particle methods},'' \emph{Ann. Inst. Stat. Math.}, vol.~55, pp.
  409--422, 2003.

\bibitem{djuric2003particle}
P.~M. Djuric, J.~H. Kotecha, J.~Zhang \emph{et~al.}, ``Particle filtering,''
  \emph{IEEE Signal Process. Mag.}, vol.~20, no.~5, pp. 19--38, 2003.

\bibitem{graves2014neural}
A.~Graves, G.~Wayne, and I.~Danihelka, ``{Neural Turing Machines},''
  \emph{arXiv preprint arXiv:1410.5401}, 2014.

\bibitem{graves2016hybrid}
A.~Graves, G.~Wayne, M.~Reynolds \emph{et~al.}, ``{Hybrid computing using a
  neural network with dynamic external memory},'' \emph{Nature}, vol. 538, no.
  7626, pp. 471--476, 2016.

\bibitem{nam2023neural}
H.~Nam and S.~B. Seo, ``{Neural Attention Memory},'' \emph{arXiv preprint
  arXiv:2302.09422}, 2023.

\bibitem{weston2014memory}
J.~Weston, S.~Chopra, and A.~Bordes, ``{Memory Networks},'' \emph{arXiv
  preprint arXiv:1410.3916}, 2014.

\bibitem{hopfield1984neurons}
J.~J. Hopfield, ``{Neurons with graded response have collective computational
  properties like those of two-state neurons.}'' \emph{Proc. Natl. Acad. Sci.
  U. S. A.}, vol.~81, no.~10, pp. 3088--3092, 1984.

\bibitem{krotov2016dense}
D.~Krotov and J.~J. Hopfield, ``{Dense Associative Memory for Pattern
  Recognition},'' in \emph{Proc. 29th Int. Conf. Neural Inf. Process. Syst.
  (NeurIPS)}, 2016, pp. 1--9.

\bibitem{vaswani2017attention}
A.~Vaswani, N.~Shazeer, N.~Parmar \emph{et~al.}, ``{Attention is All you
  Need},'' in \emph{Proc. 30th Int. Conf. Neural Inf. Process. Syst.
  (NeurIPS)}, 2017.

\bibitem{hopfield1982neural}
J.~J. Hopfield, ``{Neural networks and physical systems with emergent
  collective computational abilities.}'' \emph{Proc. Natl. Acad. Sci. U. S.
  A.}, vol.~79, no.~8, pp. 2554--2558, 1982.

\bibitem{alahi2016social}
A.~Alahi, K.~Goel, V.~Ramanathan \emph{et~al.}, ``{Social LSTM: Human
  Trajectory Prediction in Crowded Spaces},'' in \emph{Proc. IEEE/CVF Comput.
  Soc. Conf. Comput. Vis. Pattern Recognit. (CVPR)}, 2016, pp. 961--971.

\bibitem{Rhinehart2020Deep}
N.~Rhinehart, R.~McAllister, and S.~Levine, ``{Deep Imitative Models for
  Flexible Inference, Planning, and Control},'' in \emph{Int. Conf. Learn.
  Represent. (ICLR)}, 2020, pp. 1--20.

\bibitem{hu2022model}
A.~Hu, G.~Corrado, N.~Griffiths \emph{et~al.}, ``{Model-Based Imitation
  Learning for Urban Driving},'' in \emph{Proc. 35th Int. Conf. Neural Inf.
  Process. Syst. (NeurIPS)}, 2022, pp. 20\,703--20\,716.

\bibitem{garnelo2018conditional}
M.~Garnelo, D.~Rosenbaum, C.~Maddison \emph{et~al.}, ``{Conditional Neural
  Processes},'' in \emph{Proc. 35th Int. Conf. Mach. Learn. (ICML)}.\hskip 1em
  plus 0.5em minus 0.4em\relax PMLR, 2018, pp. 1704--1713.

\bibitem{ferreira2021benchmarking}
R.~S. Ferreira, J.~Arlat, J.~Guiochet \emph{et~al.}, ``{Benchmarking Safety
  Monitors for Image Classifiers with Machine Learning},'' in \emph{Proc. IEEE
  Pac. Rim Int. Symp. Dependable Comput. (PRDC)}, 2021, pp. 7--16.

\bibitem{carion2020end}
N.~Carion, F.~Massa, G.~Synnaeve \emph{et~al.}, ``{End-to-End Object Detection
  with Transformers},'' in \emph{Proc. Eur. Conf. Comput. Vis. (ECCV)}.\hskip
  1em plus 0.5em minus 0.4em\relax Springer, 2020, pp. 213--229.

\bibitem{shi2022motion}
S.~Shi, L.~Jiang, D.~Dai \emph{et~al.}, ``{Motion Transformer with Global
  Intention Localization and Local Movement Refinement},'' in \emph{Proc. 36th
  Int. Conf. Neural Inf. Process. Syst. (NeurIPS)}, 2022, pp. 6531--6543.

\bibitem{chen2024vadv2}
S.~Chen, B.~Jiang, H.~Gao \emph{et~al.}, ``{VADv2: End-to-End Vectorized
  Autonomous Driving via Probabilistic Planning},'' \emph{arXiv preprint
  arXiv:2402.13243}, 2024.

\bibitem{rosenblatt1958perceptron}
F.~Rosenblatt, ``{The perceptron: A probabilistic model for information storage
  and organization in the brain.}'' \emph{Psychol. Rev.}, vol.~65, no.~6, p.
  386, 1958.

\bibitem{redmon2016you}
J.~Redmon, S.~Divvala, R.~Girshick \emph{et~al.}, ``{You Only Look Once:
  Unified, Real-Time Object Detection},'' in \emph{Proc. IEEE/CVF Comput. Soc.
  Conf. Comput. Vis. Pattern Recognit. (CVPR)}, 2016, pp. 779--788.

\bibitem{girshick2015fast}
R.~Girshick, ``{Fast R-CNN},'' in \emph{Proc. IEEE/CVF Int. Conf. Comput. Vis.
  (ICCV)}, 2015, pp. 1440--1448.

\bibitem{dhanachandra2015image}
N.~Dhanachandra, K.~Manglem, and Y.~J. Chanu, ``{Image Segmentation Using
  K-means Clustering Algorithm and Subtractive Clustering Algorithm},''
  \emph{Procedia Computer Science}, vol.~54, pp. 764--771, 2015.

\bibitem{lorenz1973ruckseite}
K.~Lorenz, \emph{{Die R{\"u}ckseite des Spiegels. Versuch einer Naturgeschichte
  menschlichen Erkennens}}.\hskip 1em plus 0.5em minus 0.4em\relax M{\"u}nchen
  and Z{\"u}rich: Piper \& Co Verlag, 1973.

\bibitem{zhu2020deformable}
X.~Zhu, W.~Su, L.~Lu \emph{et~al.}, ``{Deformable DETR: Deformable Transformers
  for End-to-End Object Detection},'' \emph{arXiv preprint arXiv:2010.04159},
  2020.

\bibitem{li2021universal}
D.~Li, J.~Zhang, and K.~Huang, ``{Universal adversarial perturbations against
  object detection},'' \emph{Pattern Recognit.}, vol. 110, p. 107584, 2021.

\bibitem{li2018humanlike}
L.~Li, K.~Ota, and M.~Dong, ``{Humanlike Driving: Empirical Decision-Making
  System for Autonomous Vehicles},'' \emph{IEEE Trans. Veh. Technol.}, vol.~67,
  no.~8, pp. 6814--6823, 2018.

\bibitem{narayanan2021divide}
S.~Narayanan, R.~Moslemi, F.~Pittaluga \emph{et~al.}, ``{Divide-and-Conquer for
  Lane-Aware Diverse Trajectory Prediction},'' in \emph{Proc. IEEE/CVF Comput.
  Soc. Conf. Comput. Vis. Pattern Recognit. (CVPR)}, 2021, pp.
  15\,799--15\,808.

\bibitem{chen2020learning}
D.~Chen, B.~Zhou, V.~Koltun \emph{et~al.}, ``{Learning by Cheating},'' in
  \emph{Proc. Conf. Robo. Learn. (CoRL)}.\hskip 1em plus 0.5em minus
  0.4em\relax PMLR, 2020, pp. 66--75.

\bibitem{shevitz1994lyapunov}
D.~Shevitz and B.~Paden, ``{Lyapunov Stability Theory of Nonsmooth Systems},''
  \emph{IEEE Trans. Autom. Control}, vol.~39, no.~9, pp. 1910--1914, 1994.

\bibitem{sastry1999lyapunov}
S.~Sastry and S.~Sastry, ``{Lyapunov Stability Theory},'' \emph{Nonlinear
  Systems: Analysis, Stability, and Control}, pp. 182--234, 1999.

\bibitem{drgovna2022dissipative}
J.~Drgo{\v{n}}a, A.~Tuor, S.~Vasisht \emph{et~al.}, ``Dissipative deep neural
  dynamical systems,'' \emph{IEEE Open J. of Control Syst.}, vol.~1, pp.
  100--112, 2022.

\bibitem{kullback1951information}
S.~Kullback and R.~A. Leibler, ``{On Information and Sufficiency},'' \emph{Ann.
  Math. Stat.}, vol.~22, no.~1, pp. 79--86, 1951.

\bibitem{milan2017online}
A.~Milan, S.~H. Rezatofighi, A.~Dick \emph{et~al.}, ``{Online Multi-Target
  Tracking Using Recurrent Neural Networks},'' in \emph{Proc. AAAI Conf. Artif.
  Intell. (AAAI)}, vol.~31, no.~1, 2017.

\bibitem{song2024motion}
N.~Song, B.~Zhang, X.~Zhu \emph{et~al.}, ``{Motion Forecasting in Continuous
  Driving},'' \emph{arXiv preprint arXiv:2410.06007}, 2024.

\bibitem{renz2022plant}
K.~Renz, K.~Chitta, O.-B. Mercea \emph{et~al.}, ``{PlanT: Explainable Planning
  Transformers via Object-Level Representations},'' in \emph{Proc. Conf. Robo.
  Learn. (CoRL)}.\hskip 1em plus 0.5em minus 0.4em\relax PMLR, 2022, pp.
  459--470.

\bibitem{eu_parliament_2024corr}
\BIBentryALTinterwordspacing
{European Parliament}, ``{Corrigendum to the position of the European
  Parliament adopted at first reading on 13 March 2024 with a view to the
  adoption of Regulation (EU) 2024/... of the European Parliament and of the
  Council laying down harmonised rules on artificial intelligence and amending
  Regulations (EC) No 300/2008, (EU) No 167/2013, (EU) No 168/2013, (EU)
  2018/858, (EU) 2018/1139 and (EU) 2019/2144 and Directives 2014/90/EU, (EU)
  2016/797 and (EU) 2020/1828 (Artificial Intelligence Act)
  P9\_TA(2024)0138"},'' (COM(2021)0206 – C9-0146/2021 – 2021/0106(COD)),
  Brussels, April 19 2024. [Online]. Available:
  \url{https://www.europarl.europa.eu/doceo/document/TA-9-2024-0138-FNL-COR01_EN.pdf}
\BIBentrySTDinterwordspacing

\bibitem{metiaigovernanceguidelines}
\BIBentryALTinterwordspacing
{Ministry of Economy, Trade and Industry (METI)}. (2021, July 9) {\textit{Call
  for Public Comments on "AI Governance Guidelines for Implementation of AI
  Principles Ver. 1.0" Opens}}. [Online]. Available:
  \url{https://www.meti.go.jp/english/press/2021/0709_004.html}
\BIBentrySTDinterwordspacing

\bibitem{dsiaiframework}
F.~Thouvenin, M.~Christen, A.~Bernstein \emph{et~al.}, ``{Positionspapier: Ein
  Rechtsrahmen f{\"u}r K{\"u}nstliche Intelligenz},'' \emph{Digital Society
  Initiative}, 2021.

\bibitem{billc27}
\BIBentryALTinterwordspacing
{House of Commons of Canada, First Session}. (2022, June) {\textit{Bill C-27,
  Consumer Privacy Protection Act, PART 3 Artificial Intelligence and Data
  Act}}. [Online]. Available:
  \url{https://www.parl.ca/DocumentViewer/en/44-1/bill/C-27/first-reading}
\BIBentrySTDinterwordspacing

\bibitem{aigovernanceprinciples}
\BIBentryALTinterwordspacing
{National Committee on the Governance of the New Generation of Artificial
  Intelligence}. (2019, June) {\textit{Developing Responsible Artificial
  Intelligence: Release of the New Generation of Artificial Intelligence
  Governance Principles}}. [Online]. Available:
  \url{https://www.most.gov.cn/kjbgz/201906/t20190617_147107.html}
\BIBentrySTDinterwordspacing

\bibitem{ai2023artificial}
E.~Tabassi, ``{Artificial Intelligence Risk Management Framework (AI RMF
  1.0)},'' \emph{{National Institute of Standards and Technology (NIST),
  Gaithersburg, MD}}, 2023, {DOI: 10.6028/NIST.AI.100-1}.

\bibitem{burton2020mind}
S.~Burton, I.~Habli, T.~Lawton \emph{et~al.}, ``{Mind the gaps: Assuring the
  safety of autonomous systems from an engineering, ethical, and legal
  perspective},'' \emph{Artif. Intell.}, vol. 279, pp. 1--16, 2020, {Art. no.
  103201}.

\bibitem{burton2023closing}
\BIBentryALTinterwordspacing
S.~Burton and J.~A. McDermid, ``{Closing the gaps: Complexity and uncertainty
  in the safety assurance and regulation of automated driving},'' 2023.
  [Online]. Available:
  \url{https://publica-rest.fraunhofer.de/server/api/core/bitstreams/c0198205-8061-4fcf-bfa3-02e37bc2780c/content}
\BIBentrySTDinterwordspacing

\bibitem{liu2020energy}
W.~Liu, X.~Wang, J.~Owens \emph{et~al.}, ``{Energy-based Out-of-distribution
  Detection},'' in \emph{Proc. 33th Int. Conf. Neural Inf. Process. Syst.
  (NeurIPS)}, 2020, pp. 21\,464--21\,475.

\bibitem{gangal2020likelihood}
V.~Gangal, A.~Arora, A.~Einolghozati \emph{et~al.}, ``{Likelihood Ratios and
  Generative Classifiers for Unsupervised Out-of-Domain Detection in Task
  Oriented Dialog},'' in \emph{Proc. AAAI Conf. on Artif. Intell.}, vol.~34,
  no.~05, 2020, pp. 7764--7771.

\bibitem{xu2020deep}
H.~Xu, K.~He, Y.~Yan \emph{et~al.}, ``{A Deep Generative Distance-Based
  Classifier for Out-of-Domain Detection with Mahalanobis Space},'' in
  \emph{Proc. 28th Int. Conf. Comput. Linguist.}, 2020, pp. 1452--1460.

\bibitem{hacker2023insufficiency}
L.~Hacker and J.~Seewig, ``{Insufficiency-driven DNN error detection in the
  context of SOTIF on traffic sign recognition use case},'' \emph{IEEE Open J.
  Intell. Transp. Syst.}, vol.~4, pp. 58--70, 2023.

\bibitem{rottmann2020detection}
M.~Rottmann, K.~Maag, R.~Chan \emph{et~al.}, ``{Detection of False Positive and
  False Negative Samples in Semantic Segmentation},'' in \emph{Proc. IEEE Des.
  Autom. Test Eur. (DATE)}, 2020, pp. 1351--1356.

\bibitem{burton2023addressing}
S.~Burton and B.~Herd, ``Addressing uncertainty in the safety assurance of
  machine-learning,'' \emph{Front. Comput. Sci.}, vol.~5, pp. 1--17, 2023,
  {Art. no. 1132580}.

\bibitem{lakkaraju2017identifying}
H.~Lakkaraju, E.~Kamar, R.~Caruana \emph{et~al.}, ``{Identifying Unknown
  Unknowns in the Open World: Representations and Policies for Guided
  Exploration},'' in \emph{Proc. AAAI Conf. on Artif. Intell.}, vol.~31, no.~1,
  2017.

\bibitem{gowal2020achieving}
S.~Gowal, C.~Qin, P.-S. Huang \emph{et~al.}, ``{Achieving Robustness in the
  Wild via Adversarial Mixing With Disentangled Representations},'' in
  \emph{Proc. IEEE/CVF Comput. Soc. Conf. Comput. Vis. Pattern Recognit.
  (CVPR)}, 2020, pp. 1211--1220.

\bibitem{liu2023causal}
J.~Liu, S.~Zheng, and C.~Wang, ``{Causal Graph Attention Network with
  Disentangled Representations for Complex Systems Fault Detection},''
  \emph{Reliab. Eng. Syst. Saf.}, vol. 235, p. 109232, 2023.

\bibitem{rombach2023controlled}
K.~Rombach, G.~Michau, and O.~Fink, ``{Controlled generation of unseen faults
  for Partial and Open-Partial domain adaptation},'' \emph{Reliab. Eng. Syst.
  Saf.}, vol. 230, p. 108857, 2023.

\bibitem{kahneman2011thinking}
D.~Kahneman, \emph{Thinking, Fast and Slow}.\hskip 1em plus 0.5em minus
  0.4em\relax New York, NY, USA: Farrar, Straus and Giroux, 2011.

\bibitem{rawlings2000tutorial}
J.~B. Rawlings, ``Tutorial overview of model predictive control,'' \emph{IEEE
  control systems magazine}, vol.~20, no.~3, pp. 38--52, 2000.

\bibitem{englert2019software}
T.~Englert, A.~V{\"o}lz, F.~Mesmer \emph{et~al.}, ``{A software framework for
  embedded nonlinear model predictive control using a gradient-based augmented
  Lagrangian approach (GRAMPC)},'' \emph{Optim. Eng.}, vol.~20, pp. 769--809,
  2019.

\bibitem{allan2019moving}
D.~A. Allan and J.~B. Rawlings, ``{Moving Horizon Estimation},'' \emph{Handbook
  of Model Predictive Control}, pp. 99--124, 2019.

\bibitem{chang2019argoverse}
M.-F. Chang, J.~Lambert, P.~Sangkloy \emph{et~al.}, ``{Argoverse: 3D Tracking
  and Forecasting With Rich Maps},'' in \emph{Proc. IEEE/CVF Comput. Soc. Conf.
  Comput. Vis. Pattern Recognit. (CVPR)}, 2019, pp. 8748--8757.

\bibitem{zhan2019interaction}
W.~Zhan, L.~Sun, D.~Wang \emph{et~al.}, ``{INTERACTION Dataset: An
  INTERnational, Adversarial and Cooperative moTION Dataset in Interactive
  Driving Scenarios with Semantic Maps},'' \emph{arXiv preprint
  arXiv:1910.03088}, 2019.

\bibitem{houston2021one}
J.~Houston, G.~Zuidhof, L.~Bergamini \emph{et~al.}, ``{One Thousand and One
  Hours: Self-driving Motion Prediction Dataset},'' in \emph{Int. Conf. Learn.
  Represent. (ICLR)}.\hskip 1em plus 0.5em minus 0.4em\relax PMLR, 2021, pp.
  409--418.

\bibitem{caesar2021nuplan}
H.~Caesar, J.~Kabzan, K.~S. Tan \emph{et~al.}, ``{NuPlan: A closed-loop
  ML-based planning benchmark for autonomous vehicles},'' \emph{arXiv preprint
  arXiv:2106.11810}, 2021.

\bibitem{ullrich2025iv}
L.~Ullrich, Z.~Mujirishvili, and K.~Graichen, ``{Enhancing System
  Self-Awareness and Trust of AI: A Case Study in Trajectory Prediction and
  Planning},'' in \emph{Proc. IEEE Intell. Veh. Symp. (IV)}, 2025, pp. 1--8,
  {accepted}.

\bibitem{bardes2023mc}
A.~Bardes, J.~Ponce, and Y.~LeCun, ``{MC-JEPA: A Joint-Embedding Predictive
  Architecture for Self-Supervised Learning of Motion and Content Features},''
  \emph{arXiv preprint arXiv:2307.12698}, 2023.

\bibitem{zhang2025bridging}
B.~Zhang, N.~Song, X.~Jin \emph{et~al.}, ``{Bridging Past and Future:
  End-to-End Autonomous Driving with Historical Prediction and Planning},''
  \emph{arXiv preprint arXiv:2503.14182}, 2025.

\bibitem{ecksteinautotech}
R.~van Kempen, B.~Lampe, M.~Leuffen \emph{et~al.}, ``{AUTOtech.agil:
  Architecture and Technologies for Orchestrating Automotive Agility},'' in
  \emph{Proc. of 32nd Aachen Colloquium Sustainable Mobility}, 2023.

\bibitem{rojas2018invariant}
M.~Rojas-Carulla, B.~Sch{\"o}lkopf, R.~Turner \emph{et~al.}, ``{Invariant
  Models for Causal Transfer Learning},'' \emph{J. Mach. Learn. Res.}, vol.~19,
  no.~36, pp. 1--34, 2018.

\bibitem{magliacane2018domain}
S.~Magliacane, T.~Van~Ommen, T.~Claassen \emph{et~al.}, ``{Domain Adaptation by
  Using Causal Inference to Predict Invariant Conditional Distributions},'' in
  \emph{Proc. 31st Int. Conf. Neural Inf. Process. Syst. (NeurIPS)}, 2018, pp.
  1--11.

\bibitem{scholkopf2001estimating}
B.~Sch{\"o}lkopf, J.~C. Platt, J.~Shawe-Taylor \emph{et~al.}, ``{Estimating the
  Support of a High-Dimensional Distribution},'' \emph{Neural Comput.},
  vol.~13, no.~7, pp. 1443--1471, 2001.

\bibitem{bendale2016towards}
A.~Bendale and T.~E. Boult, ``{Towards Open Set Deep Networks},'' in
  \emph{Proc. IEEE/CVF Comput. Soc. Conf. Comput. Vis. Pattern Recognit.
  (CVPR)}, 2016, pp. 1563--1572.

\bibitem{heckerman1995decision}
D.~Heckerman, J.~S. Breese, and K.~Rommelse, ``Decision-theoretic
  troubleshooting,'' \emph{Commun. ACM}, vol.~38, no.~3, pp. 49--57, 1995.

\bibitem{nushi2018towards}
B.~Nushi, E.~Kamar, and E.~Horvitz, ``Towards accountable ai: Hybrid
  human-machine analyses for characterizing system failure,'' in \emph{Proc.
  AAAI Conf. Human Comput. Crowdsourc. (HCOMP)}, vol.~6, 2018, pp. 126--135.

\bibitem{hendrycks2019benchmarking}
D.~Hendrycks and T.~Dietterich, ``{Benchmarking Neural Network Robustness to
  Common Corruptions and Perturbations},'' in \emph{Int. Conf. Learn.
  Represent. (ICLR)}, 2019, pp. 1--16.

\bibitem{duchi2021learning}
J.~C. Duchi and H.~Namkoong, ``{Learning models with uniform performance via
  distributionally robust optimization},'' \emph{Ann. Stat.}, vol.~49, no.~3,
  pp. 1378--1406, 2021.

\bibitem{meinshausen2018causality}
N.~Meinshausen, ``{Causality from a Distributional Robustness Point of View},''
  in \emph{IEEE Data Science Workshop (DSW)}, 2018, pp. 6--10.

\bibitem{gupta2024s}
S.~Gupta and D.~Rothenh{\"a}usler, ``{The s-value: evaluating stability with
  respect to distributional shifts},'' in \emph{Proc. 36th Int. Conf. Neural
  Inf. Process. Syst. (NeurIPS)}, 2024, pp. 1--13.

\bibitem{chen2021mandoline}
M.~Chen, K.~Goel, N.~S. Sohoni \emph{et~al.}, ``{Mandoline: Model Evaluation
  under Distribution Shift},'' in \emph{Proc. 38th Int. Conf. Mach. Learn.
  (ICML)}.\hskip 1em plus 0.5em minus 0.4em\relax PMLR, 2021, pp. 1617--1629.

\bibitem{subbaswamy2022unifying}
A.~Subbaswamy, B.~Chen, and S.~Saria, ``{A unifying causal framework for
  analyzing dataset shift-stable learning algorithms},'' \emph{J. Causal
  Inference}, vol.~10, no.~1, pp. 64--89, 2022.

\end{thebibliography}

\vspace{-15 mm}
\begin{IEEEbiography}[{\includegraphics[width=1in,height=1.25in,clip,keepaspectratio]{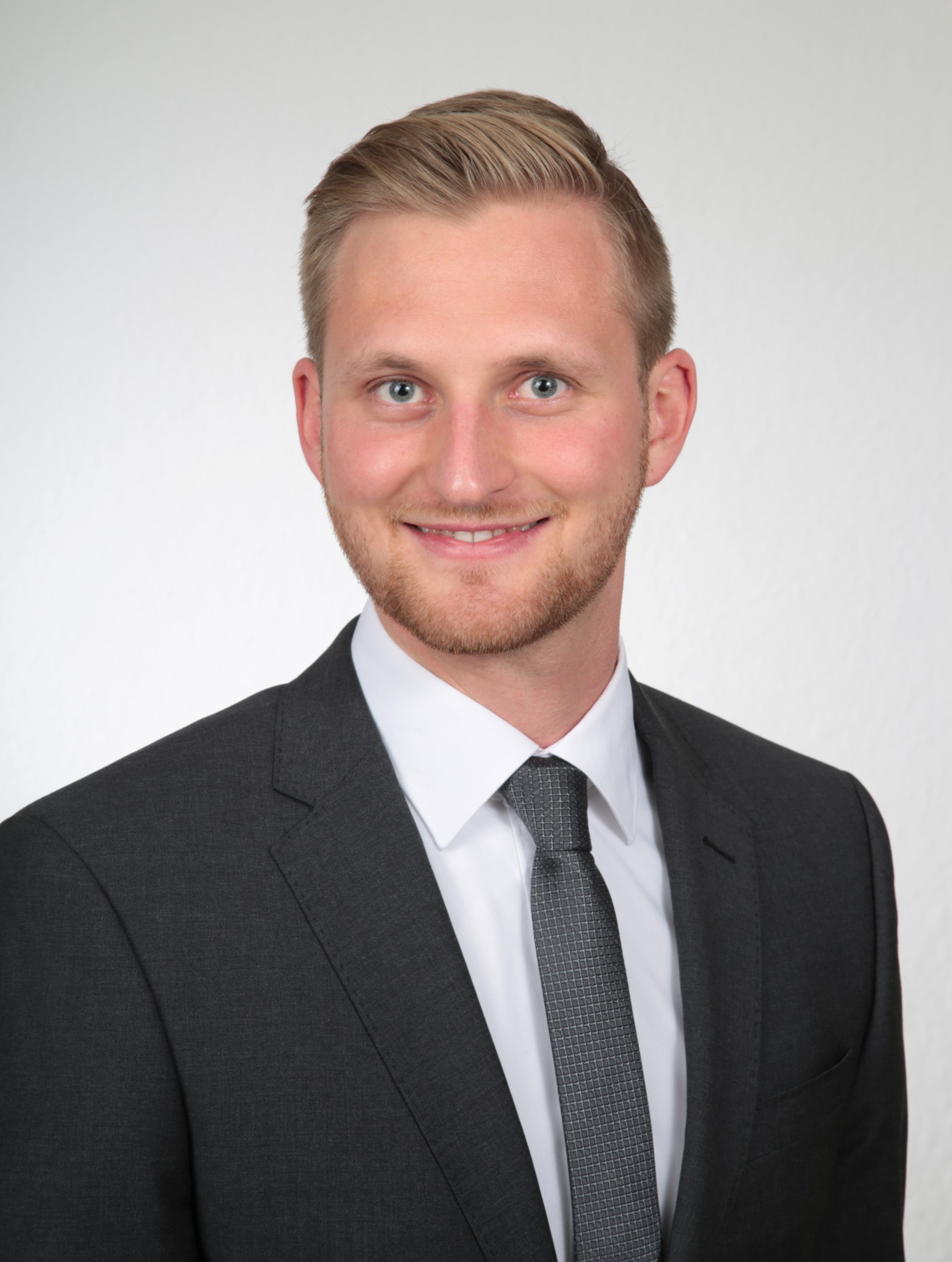}}]{Lars Ullrich}
	received the M.Sc. degree in mechatronics from Friedrich–Alexander–Universita\"at Erlangen–N\"urnberg, Germany, in 2022, where he is currently pursuing the Ph.D. (Dr.Ing.) degree with the Chair of Automatic Control. His research interests include probabilistic trajectory planning for safe and reliable autonomous driving in uncertain dynamic environments with a focus on addressing challenges arising from the use of AI systems in automated driving. Since early 2025, he has been elected Vice-Chair of the IEEE Intelligent Transportation Systems Society (ITSS) German Chapter.
\end{IEEEbiography}
\vspace*{-15.0mm}

\begin{IEEEbiography}[{\includegraphics[width=1in,height=1.25in,clip,keepaspectratio]{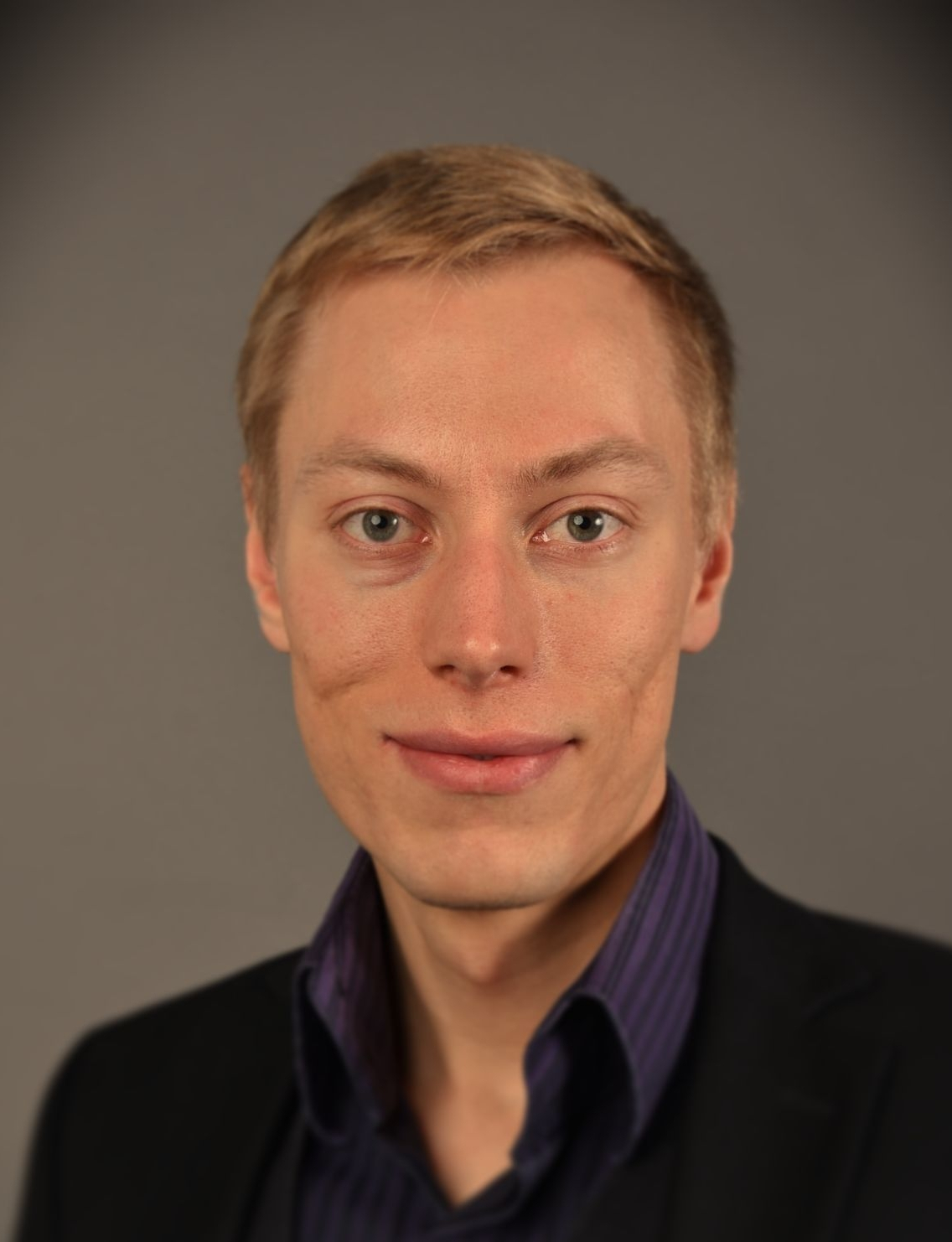}}]{Walter Zimmer} received his M.Sc. and PhD in Computer Science at the Technical University of Munich (TUM) in 2025. During his studies, he stayed abroad at the Technical University of Delft (TUD) in the Netherlands and the University of California, San Diego (UCSD) in the United States where he was involved in developing perception and autonomous driving algorithms. During his PhD at TUM, he also worked for one year as an Autonomous Systems Engineer at STTech GmbH. His current research interests are mainly 3D object detection, multi-object tracking, sensor data fusion, vision language models, and 3D scene understanding for automated driving systems and intelligent transportation systems.
\end{IEEEbiography}
\vspace*{-15.0mm}
\begin{IEEEbiography}[{\includegraphics[width=1in,height=1.25in,clip,keepaspectratio]{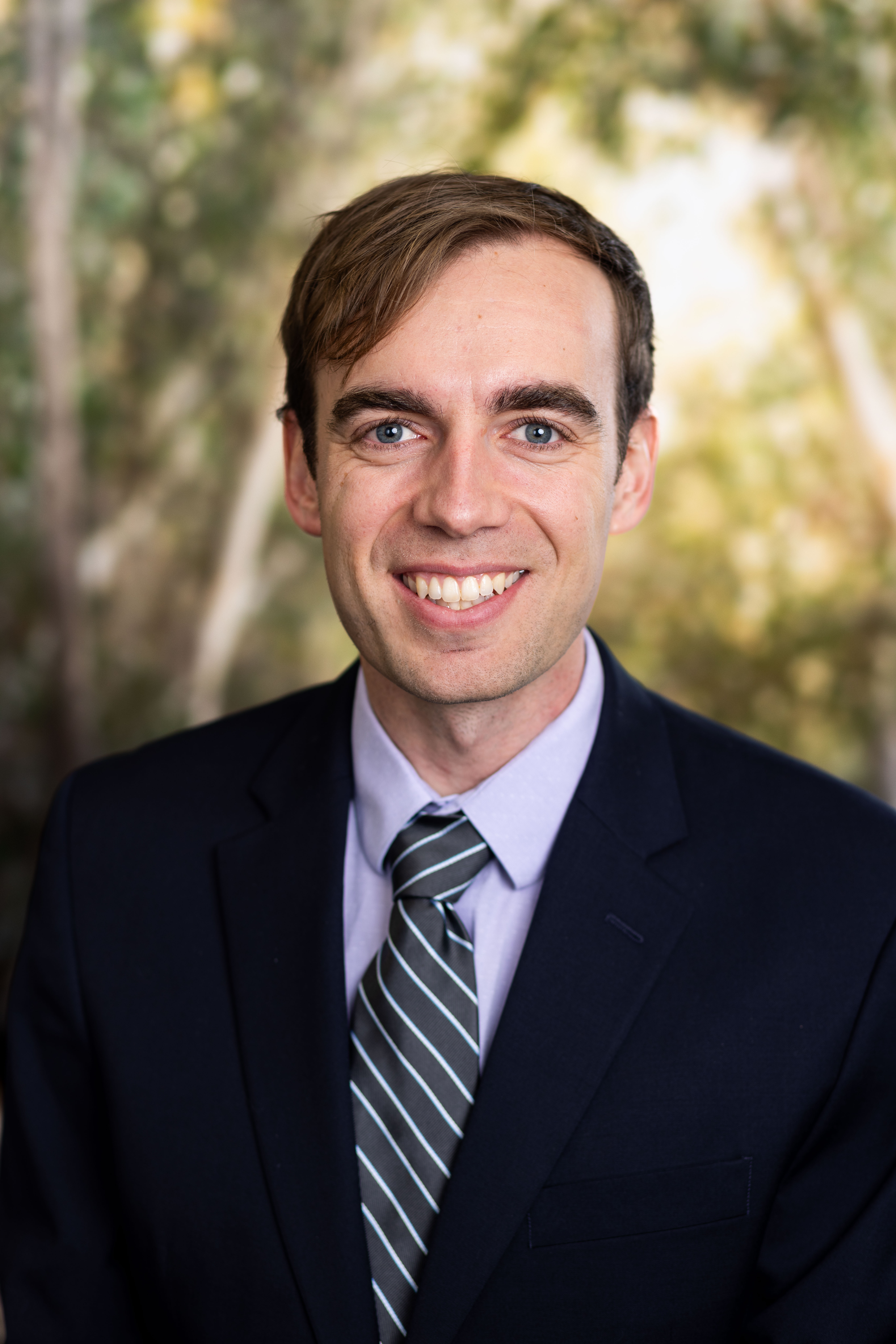}}]{Ross Greer} is an assistant professor in the Department of Computer Science and Engineering at the University of California Merced. He received his PhD in Electrical and Computer Engineering from the University of California San Diego, as a member of Mohan Trivedi's Laboratory for Intelligent and Safe Automobiles (LISA), where Ross's research has received fellowships and awards from Qualcomm, NHTSA, and DAAD AI-Net. His current research centers on active learning and safe planning for autonomous vehicles, especially during interactions with humans inside and outside the vehicle. 
\end{IEEEbiography}
\vspace*{-15.0mm}
\begin{IEEEbiography}[{\includegraphics[width=1in,height=1.25in,clip,keepaspectratio]{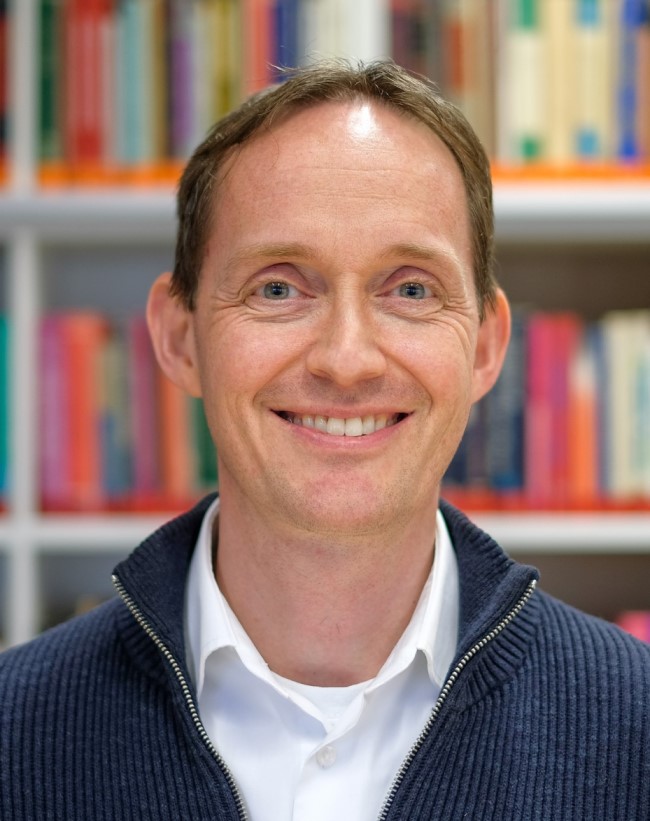}}]{Knut Graichen}
	(Senior Member, IEEE) received the Diploma-Ing. degree in engineering cybernetics and the Ph.D. (Dr.-Ing.) degree from the University of Stuttgart, Stuttgart, Germany, in 2002 and 2006, respectively. In 2007, he was a Post-Doctoral Researcher with the Center Automatique et Syst\`emes, MINES ParisTech, France. In 2008, he joined the Automation and Control Institute, Vienna University of Technology, Vienna, Austria, as a Senior Researcher. In 2010, he became a Professor with the Institute of Measurement, Control and Microtechnology, Ulm University, Ulm, Germany. Since 2019, he has been the Head of the Chair of Automatic Control, Friedrich–Alexander–Universita\"at Erlangen–N\"urnberg, Germany. His current research interests include distributed and learning control and model predictive control of dynamical systems for automotive, mechatronic, and robotic applications. Dr. Graichen is the Editor-in-Chief of Control Engineering Practice.
\end{IEEEbiography}
\newpage
\vspace*{-129.0mm}
\begin{IEEEbiography}[{\includegraphics[width=1in,height=1.25in,clip,keepaspectratio]{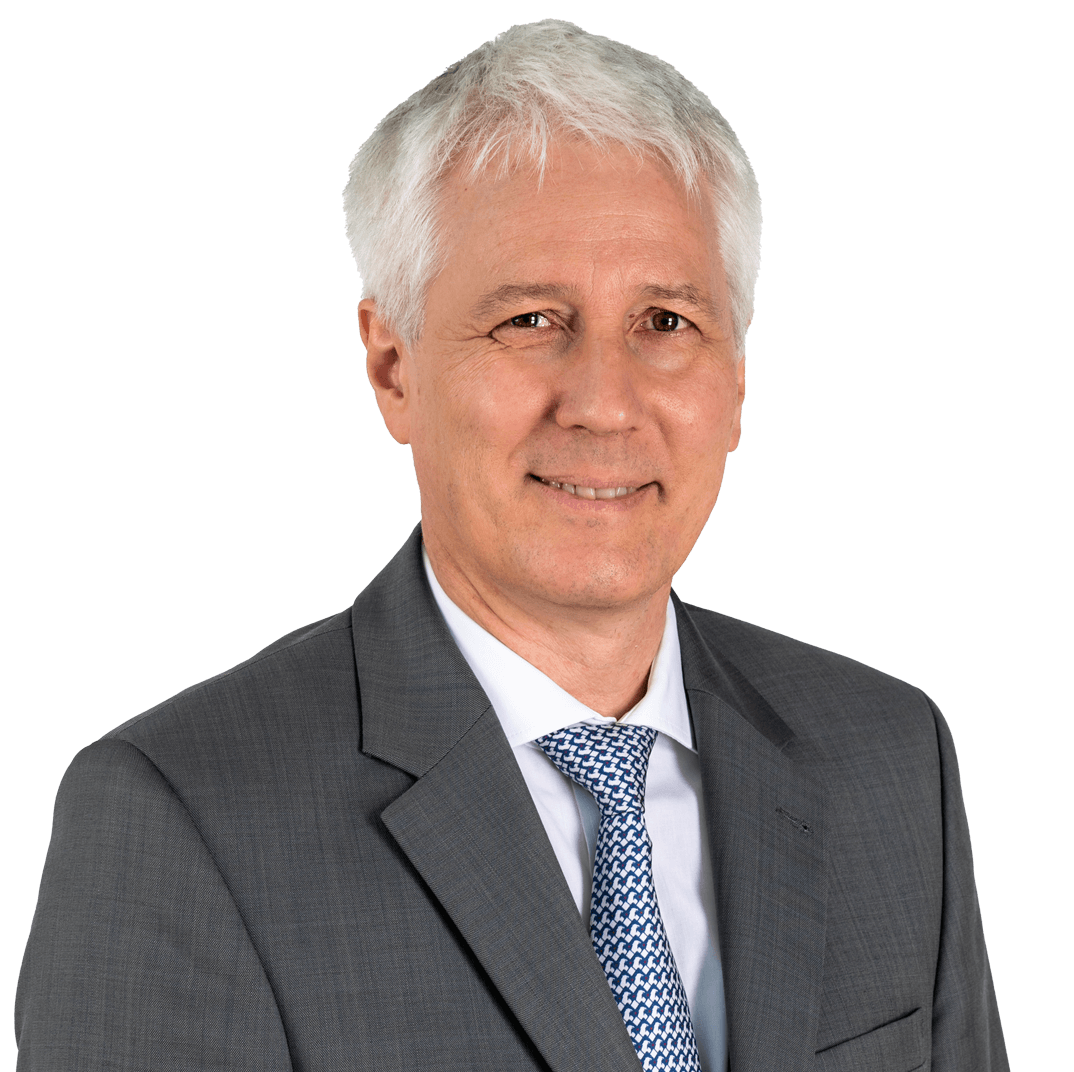}}]{Alois C. Knoll}
(Fellow, IEEE) received the M.Sc. degree in electrical / communications engineering from the Uni. of Stuttgart, Stuttgart, Germany, in 1985, and the Ph.D. degree (summa cum laude) in computer science from the Tech. Uni. of Berlin (TU Berlin), Berlin, Germany, in 1988. He was with the Faculty of the Computer Science Department, TU Berlin, until 1993. He joined Bielefeld University, Bielefeld, Germany, as a Full Professor, where he has served as the Director for the Tech. Informatics Research Group, until 2001. Since 2001, he has been a Professor at the Department of Informatics, Technical Uni. of Munich (TUM), Munich. He was also on the Board of Directors of the Central Institute of Medical Technology, TUM (IMETUM). From 2004 to 2006, he was an Executive Director of the Institute of Computer Science, TUM. His research interests include cognitive, medical robotics, multi-agent systems, data fusion, adaptive systems, multimedia information retrieval, model-driven development of embedded systems with applications to automotive software and electric transportation, and simulation systems for robotics and trafﬁc. He was a member of the EU’s Highest Advisory Board on Information Technology, Information Society Technology Advisory Group (ISTAG), and its subgroup on Future and Emerging Technologies (FET), from 2007 to 2009.
\end{IEEEbiography}
\vspace*{-110.0mm}
\begin{IEEEbiography}[{\includegraphics[width=1in,height=1.25in,clip,keepaspectratio]{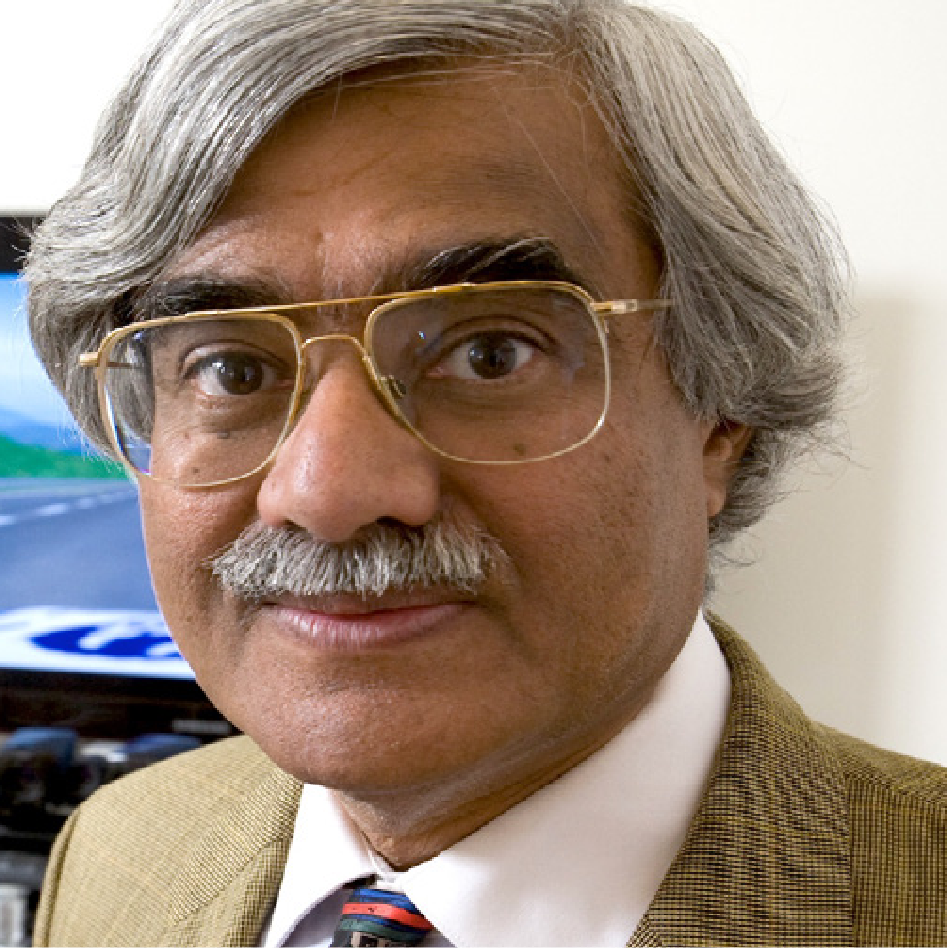}}]{Mohan Manubhai Trivedi}
(Life Fellow IEEE, SPIE, and IAPR) is a Distinguished Professor of Electrical and Computer Engineering at the University of California, San Diego, and the founding director of the Computer Vision and Robotics Research (CVRR, est. 1986) and the Laboratory for Intelligent and Safe Automobiles (LISA, est. 2001). His research includes intelligent vehicles, intelligent transportation systems (ITS), autonomous driving, driver assistance systems, active safety, human-robot interactivity, and machine vision areas. UCSD LISA was awarded the IEEE ITSS Lead Institution Award. Trivedi has served as the editor-in-chief of Machine Vision and Applications, Senior Editor of the IEEE Transactions on IV and ITSC, and Chairman of the Robotics Technical Committee of the IEEE Computer Society and Board of Governors of the IEEE ITSS and IEEE SMC societies.
\end{IEEEbiography}
%\newpage

\end{document}